%% file: main.tex
\documentclass{article} %
\usepackage{iclr2026_conference,times}

\input{math_commands.tex}

\usepackage{amsmath,amssymb}

\usepackage{hyperref}
\usepackage{url}
\usepackage[capitalize,noabbrev]{cleveref} %
\usepackage{graphicx}
\usepackage{xcolor}
\usepackage{colortbl} %
\usepackage{enumitem} %
\usepackage{pgfplots} %
\pgfplotsset{compat=1.18}
\usepgfplotslibrary{groupplots}
\usepackage{xspace}
\usepackage{booktabs}
\usepackage{longtable} %
\usepackage{array}     %
\usepackage{needspace}
\usepackage{float} %
\usepackage{wrapfig} %
\usepackage{multirow}
\usepackage[most]{tcolorbox} %
\usepackage{pmboxdraw} %
\usepackage{forest} %
\usepackage{tikz} %
\usetikzlibrary{arrows.meta,positioning,fit}
\usepackage{fontawesome5} %

\input{macros}

\title{Filesystem-Based Memory for LLM Agents:\\
Organization, Evolution, and Sustainability}

\author{Sizhe Zhou\textsuperscript{*,1,\textdagger}, Sheldon Yu\textsuperscript{*,2}, Hui Wei\textsuperscript{*,3}, Junda Wu\textsuperscript{4}, Siru Ouyang\textsuperscript{1},\\
\bf Yizhu Jiao\textsuperscript{1}, Shijia Pan\textsuperscript{3}, Julian McAuley\textsuperscript{2}, Yu Zhang\textsuperscript{5}, Tong Yu\textsuperscript{4}, Jiawei Han\textsuperscript{1}\\[3pt]
\normalfont\small \textsuperscript{1}University of Illinois Urbana-Champaign \quad \textsuperscript{2}University of California, San Diego\\
\normalfont\small \textsuperscript{3}University of California, Merced \quad \textsuperscript{4}Adobe Research \quad \textsuperscript{5}Texas A\&M University\\[2pt]
\normalfont\small \texttt{\{sizhez,siruo2,yizhuj2,hanj\}@illinois.edu} \quad \texttt{\{ziy040,jmcauley\}@ucsd.edu}\\
\normalfont\small \texttt{\{huiwei2,span24\}@ucmerced.edu} \quad \texttt{\{jundaw,tyu\}@adobe.com} \quad \texttt{yuzhang@tamu.edu}}

\iclrfinalcopy %

\begin{document}

\maketitle
\lhead{}
\renewcommand{\headrulewidth}{0pt}
{\let\thefootnote\relax\begin{NoHyper}\footnotetext{\textsuperscript{*}Equal
contribution. \textsuperscript{\textdagger}Corresponding author:
\texttt{sizhez@illinois.edu}.}\end{NoHyper}}

\begin{abstract}
\input{sections/00_abstract}
\end{abstract}

\input{sections/01_introduction}

\input{sections/03_setting}

\input{sections/04_experimental_setup}

\input{sections/05_results}   %

\input{sections/06_conclusion}

\subsubsection*{Acknowledgements}
Research was supported in part by the Institute for Geospatial
Understanding through an Integrative Discovery Environment (I-GUIDE) by
NSF under Award No. 2118329. The research has used the Delta/DeltaAI
advanced computing and data resource, supported in part by the University
of Illinois Urbana-Champaign and through allocation \#250851 from the
Advanced Cyberinfrastructure Coordination Ecosystem: Services \& Support
(ACCESS) program, which is supported by National Science Foundation grants
OAC 2320345, \#2138259, \#2138286, \#2138307, \#2137603, and \#2138296. Any
opinions, findings and conclusions or recommendations expressed in this
material are those of the author(s) and do not necessarily reflect those
of the National Science Foundation.

\bibliography{related_work}
\bibliographystyle{iclr2026_conference}

\input{sections/99_appendix}

\end{document}

%% file: math_commands.tex
\usepackage{amsmath,amsfonts,bm}

\def\eqref#1{equation~\ref{#1}}

\def\1{\bm{1}}

\DeclareMathAlphabet{\mathsfit}{\encodingdefault}{\sfdefault}{m}{sl}
\SetMathAlphabet{\mathsfit}{bold}{\encodingdefault}{\sfdefault}{bx}{n}

\def\gH{{\mathcal{H}}}

\def\gM{{\mathcal{M}}}



%% file: macros.tex
\newcommand{\mgmtagent}{management agent\xspace}
\newcommand{\Mgmtagent}{Management agent\xspace}
\newcommand{\searchagent}{search agent\xspace}
\newcommand{\Searchagent}{Search agent\xspace}
\newcommand{\execagent}{execution agent\xspace}
\newcommand{\Execagent}{Execution agent\xspace}

\newcommand{\locomo}{LoCoMo\xspace}
\newcommand{\personamem}{PersonaMem\xspace}
\newcommand{\realtalk}{REALTALK\xspace}
\newcommand{\alfworld}{ALFWorld\xspace}
\newcommand{\skillnone}{No store\xspace}
\newcommand{\skillflat}{Episode log\xspace}
\newcommand{\skillcur}{Curated skills\xspace}
\newcommand{\skillgated}{Curated skills{+}mem\xspace}
\newcommand{\skillsynth}{Curated skills{+}mem (GS)\xspace}

\definecolor{PropReview}{RGB}{0,90,181}

\definecolor{PromptAccent}{RGB}{70,95,155} 
\DeclareTextSymbol{\textquotedbl}{OT1}{34}
\lstdefinestyle{promptstyle}{%
  basicstyle=\scriptsize\ttfamily\color{black!85},
  columns=fullflexible, keepspaces=true,
  breaklines=true, breakindent=0pt,
  upquote=true,
  escapeinside={(*@}{@*)},
  literate={├}{{\textSFviii}}1 {─}{{\textSFx}}1 {│}{{\textSFxi}}1 {└}{{\textSFii}}1,
}
\newtcblisting[auto counter,
               crefname={prompt}{prompts}]{promptbox}[2][]{%
  enhanced jigsaw, breakable,
  listing only, listing options={style=promptstyle},
  colback=PromptAccent!4!white,
  colframe=PromptAccent!35!white,
  boxrule=0.6pt, arc=2mm,
  attach boxed title to top left={yshift=-2.4mm, xshift=4mm},
  boxed title style={
    colback=PromptAccent!10!white, colframe=PromptAccent!35!white,
    boxrule=0.6pt, arc=1.2mm,
    left=1.6mm, right=1.6mm, top=0.7mm, bottom=0.7mm},
  coltitle=PromptAccent!50!black,
  fonttitle=\small\bfseries\sffamily,
  title={Prompt~\thetcbcounter: #2},
  left=3.2mm, right=3.2mm, top=4mm, bottom=2.8mm,
  #1}

\newcommand{\promptvar}[1]{\textcolor{PromptAccent!60!black}{\texttt{#1}}}

\definecolor{FsFolder}{RGB}{31,78,121}
\forestset{
  memfs/.style={
    for tree={
      font=\footnotesize\ttfamily, text=black!80,
      grow'=0, child anchor=west, parent anchor=south, anchor=west,
      calign=first,
      edge path={\noexpand\path [draw, black!30, line width=0.5pt]
        (!u.south west) +(7.5pt,0) |- (.child anchor)\forestoption{edge label};},
      before typesetting nodes={if n=1 {insert before={[,phantom]}}{}},
      fit=band, before computing xy={l=15pt}, s sep=2.5pt,
    },
  },
  dir/.style={font=\footnotesize\ttfamily\bfseries, text=FsFolder},
  fsfile/.style={},
}
\newcommand{\fsanno}[1]{{\normalfont\scriptsize\itshape\color{black!50}~~#1}}
\newcommand{\fsmore}[1]{{\normalfont\scriptsize\itshape\color{black!50}\ldots~#1}}

\newcommand{\pcell}{\textcolor{black!45}{$\cdot$}}

\newcommand{\subclosed}{Closed-book\xspace}
\newcommand{\subrag}{Chunk retrieval\xspace}
\newcommand{\subdump}{Verbatim dump\xspace}
\newcommand{\subfoldered}{Foldered sessions\xspace}
\newcommand{\subreorg}{Reorganized store\xspace}
\newcommand{\subreorgpres}{Reorg.\ (preserve)\xspace}
\newcommand{\subreorgcond}{Reorg.\ (condense)\xspace}
\newcommand{\subcurated}{Agent-curated store\xspace}

%% file: sections/00_abstract.tex
Deployed LLM agents increasingly keep their long-term memory as a filesystem:
a directory tree of markdown files that the agent itself reads, writes, and
reorganizes through generic file tools. Yet research has largely passed over this medium: prior systems design
bespoke memory representations and study retrieval over them, leaving the
default's two working assumptions untested: that an agent can keep a
growing store organized as memories accumulate, conflict, and go stale,
and that this organization pays. We present the first systematic exploration of
filesystem-based memory for LLM agents. We formalize the setting as three
roles around one memory filesystem: a management agent integrates and
organizes incoming content, a search agent answers queries with cited
sources, and an execution agent supplies task trajectories that are distilled
into skills, unifying declarative memory and skills in a single store. Across
long-conversation benchmarks and embodied tasks, we vary memory shape
(agent-organized hierarchy, verbatim dump, chunk retrieval), stream scale,
tool harness (sandboxed shell, memory-tool-style functions, varied search
tooling), and the strengths of the management and search agents, tracking answer quality,
cost, and store health as memory grows.
What organization reliably buys is search economy: organized stores
roughly halve retrieval cost where material is large. Today's agents,
however, fall short of the default's promise: in our growth study, organization erodes for all but the strongest
management agent, and no agent we
measure converts organization itself into better answers. And the model is
not the only lever over a store's shape: changing the tool set alone
reshapes the store as strongly as swapping the model. The study turns the filesystem
default from an assumption into a design space for agent memory.

%% file: sections/01_introduction.tex
\section{Introduction}
\label{sec:introduction}

Large language model (LLM) agents increasingly work over horizons no single
context window can span, maintaining codebases across sessions and assisting
the same user for months. The field's ambitions reach further still, toward
agents that learn continually from their own experience and, ultimately,
intelligence of a human kind. Both demand a memory that works less like a
transcript and more like a brain: persisting across episodes and evolving,
absorbing new information, reconciling it with what is stored, and staying
organized enough to remain efficiently searchable and trustworthy as it grows.
Today's models offer no such memory: the context window is ephemeral and
degrades long before it is full \citep{liu2024lost}, so persistent external
memory has become a first-order component of agent design
\citep{zhang2025memorysurvey}. Throughout, we use \emph{memory} in the
inclusive, classical sense \citep{sumers2023cognitive}: it spans
\emph{declarative} content (facts, events, preferences, rules) and
\emph{procedural} content, the reusable \emph{skills} an agent distills from
experience \citep{wang2023voyager,anthropic2025skills}, with one store serving
both.

Research has explored many forms for this memory: OS-style paged context
\citep{packer2023memgpt}, extracted
fact stores \citep{chhikara2025mem0}, temporal knowledge graphs
\citep{rasmussen2025zep}, self-linking note networks \citep{xu2025amem},
summary banks \citep{zhong2024memorybank}, discourse-unit stores
\citep{pan2025secom,zhou2025edu}, and embedding-organized trees
\citep{rezazadeh2025memtree}, each behind its own purpose-built interface.
Deployed practice has converged on something plainer: the \emph{filesystem}.
Coding agents already live on files, so files became the natural interface for
extending their context: Anthropic's memory tool exposes memory as a directory
of files behind six generic file operations \citep{anthropic2025memorytool},
Claude Code maintains agent-written notes in an indexed memory folder
\citep{anthropic2026claudecodememory}, and agent context files and ``skills''
ship as repository markdown at ecosystem scale
\citep{openai2025agentsmd,anthropic2025skills}. The filesystem earns its
place: it is inspectable, portable, and operated with the file tools agents
already master. It is also natively hierarchical: folders form a taxonomy
whose names are its labels. Yet the medium that ships by default is the one
research has largely passed over. Prior work builds agent systems \emph{on}
filesystem memory and studies retrieval \emph{over} files
\citep{zhang2025memorysurvey}; the memory form
itself has received little systematic study: how an agent-curated file store
should be built, shaped, and kept healthy.

This default practice rests on an untested assumption: that the store stays
manageable as it grows. Writing memories once and reading them back is the
easy case; over long horizons, memories accumulate, and with them duplicates,
contradictions, stale facts, and content on one subject scattered across many
files. The store must be \emph{evolved}: updated, reconciled, reorganized.
Getting this wrong is costly: continuously rewriting a memory bank with an LLM
can degrade it below the no-memory baseline \citep{zhang2026faulty}. Existing
mechanisms do not close this gap. Academic memory operations act per item
(add, update, delete) \citep{chhikara2025mem0}, never on the shape of the
store. Industry consolidation (``dreaming'') runs outside the agent: OpenAI's
dreaming synthesizes a flat memory summary in the background
\citep{openai2026dreaming}, and Anthropic's \emph{Dreams} rebuilds a store
wholesale from past sessions \citep{anthropic2026dreamsdocs}, introduced
precisely because the working agent's own writes remain ``local and
incremental'' and the store degrades between rebuilds. External cleanup treats
the symptom; whether the agent itself can keep a growing store organized, and
whether organization repays its cost, remains assumed rather than tested. The
filesystem's native answer is the human one,
\emph{organizing}: grouping related material into folders, naming it so it can
be found again, splitting and merging as content demands. Whether LLM agents
can do the same, and whether it pays, is open in both directions. Organization
might be exactly what keeps a growing memory \emph{sustainable}, efficiently
searchable and trustworthy in content as it scales; or a flat store swept by
strong search tools might serve just as well, making curation an expensive
detour.

This paper presents, to our knowledge, the first systematic exploration of
filesystem-based memory for LLM agents. \Cref{sec:setting} formalizes the
setting as three roles around a single store (Figure~\ref{fig:overview}): a
\emph{\mgmtagent} that integrates each incoming chunk and keeps the store
organized, a \emph{\searchagent} that answers queries over it with cited
sources, and, in the skill setting, a fixed \emph{\execagent} whose task
attempts supply the chunks and consume the retrieved skills. The contracts are
minimal by design, so the roles map onto deployed harnesses; in a coding
agent, all three may be one model. We instantiate the setting for
conversational memory (question answering with source attributions over long
dialogues: \locomo, \personamem, \realtalk;
\citealp{maharana2024locomo,jiang2025personamem,lee2025realtalk}) and for
procedural memory (task success of a fixed \execagent{}: \alfworld;
\citealp{shridhar2021alfworld}), and we vary four components: how the store is
built (an organizing agent, a verbatim dump, or the raw stream chunked and
indexed for retrieval \citep{lewis2020rag}), the scale of the stream, the harness
through which agents touch the store, and the strengths of the models that
build and search. Throughout, we measure answer quality together with cost
(rounds, tokens, content read) and store health over time (whether early
memories survive later evolution and whether updates land correctly): growth
curves, not endpoints.

Our study yields an empirical characterization of filesystem memory along
five questions, each answered in both settings.
\begin{itemize}
\item \textbf{RQ1 (organization).} Left to organize, management agents grow
subject-based trees, but the shape is a signature of the model more than a
response to scale: given more material the store consolidates rather than shards, hierarchy relocating between folders, files, and in-file headings; the clearest
degenerate behavior is a reorganizing pass that silently condenses content
unless one preservation rule is added.
\item \textbf{RQ2 (value of shape).} No shape wins correctness everywhere;
organization's unambiguous payoff is search cost, and it grows with the
material. On skills the winner flips with the execution agent: a verbatim episode log serves a strong execution agent best, distilled
guidance a weak one.
\item \textbf{RQ3 (backbone model capability).} On conversation the
management agent's strength buys organizational style, not answer quality,
while the search agent's strength pays directly. Capability does couple where writing
itself fails: on one benchmark the management agent inconsistently
records changed preferences as dated updates, leaving stale facts
standing as live traits; a stronger backbone executing the same
instruction recovers about half the loss, while the same upgrade moves
nothing where the store already serves its search agent. When memory must be
distilled into procedures, capability acts as a threshold: once crossed,
what the store contains matters more than which model executes.
\item \textbf{RQ4 (sustainability under scaling).} Within our horizons,
stores only become more useful as they grow, and accumulated experience
substitutes for execution-agent capability. Store health holds in both
settings: conversational stores create their few files early and afterwards
only edit them, never deleting, while on skills early memories survive and
a stronger management agent maintains files in place rather than replacing
them. Organization is the weaker half: adherence to the taxonomy contract erodes as most stores grow, and only
the strongest management agent we track holds it. The costs that scale are effort and volume: curation never gets
cheaper per episode, kept stores stay compact relative to what the task
chains generate, and the one liability that grows with the store is the verbatim episode
log's serve-everything retrieval.
\item \textbf{RQ5 (harness).} Adding a tool changes agent behavior but not
outcomes; replacing the tool set reshapes the store itself, with direction
and payoff set by the setting, sharding and tying on long dialogue,
consolidating and winning on skills: the harness is a lever over memory
organization, not a neutral wrapper.
\end{itemize}
Read against the default's two assumptions, the answer splits: an agent can
keep a growing store useful and healthy within every horizon we measured,
though how well it stays organized tracks the management agent's
capability; whether organizing pays is conditional, on the material, on
the agent that consumes memory, and on the tools in hand; no agent we
measure converts organization itself into better answers.

\Needspace{8\baselineskip}
\paragraph{Contributions.}
\begin{itemize}
  \item \textbf{Formalization and unification.} A management/search/execution
  role decomposition of filesystem-based agent memory with minimal contracts,
  unifying declarative memory and skills in one store that mirrors deployed
  harnesses.
  \item \textbf{A systematic study framework.} Benchmarks spanning long
  conversations and embodied tasks; controlled memory shapes; harness, scale,
  and model-strength axes; and a sustainability protocol tracking store
  health, cost, and quality as memory grows.
  \item \textbf{Findings.} An empirical characterization that turns the
  filesystem default from an assumption into a design space: evidence that a
  growing store stays useful and healthy within the horizons we measure,
  while the quality of its organization remains bound to the management
  agent's capability; guidance on when curation repays its bill and when a
  dump or chunk index suffices, on what to serve weak and strong consumers
  of memory, and on where model strength pays, the search agent on
  conversation and the management agent, past a threshold, on skills; and
  the tool set established as a control knob over store organization, with
  the open problems isolated: quality benchmarks largely blind to shape,
  and horizons beyond one conversation.
\end{itemize}

%% file: sections/03_setting.tex
\section{Formalizing Filesystem-Based Agent Memory}
\label{sec:setting}

The formalization below is deliberately minimal: it abstracts the memory systems that deployed harnesses already implement into the components our study
varies.
\input{figures/fig1_overview}

\subsection{The memory store}
\label{sec:setting-store}

A \emph{memory store} $\gM$ is a finite set of files organized in a rooted
path hierarchy. Each file $f \in \gM$ is a triple $f = (p_f, d_f, c_f)$: a
path $p_f$ (for example \texttt{/memories/people/alice.md}), a one-line
description $d_f$, and text content $c_f$. Folders are the shared prefixes of
paths and carry no content of their own. The folder structure, the file
names, and the markdown headings inside files together form the store's
\emph{taxonomy}: one labeled tree that continues below the file level into
nested sections, navigated top-down, and whose names and
one-line descriptions are its only signage.
\input{figures/fig2_filezoom}

Descriptions matter because
directory listings and ranked search expose them: with names, they are what an
agent sees before opening a file. We write
$\varnothing$ for the empty store. In our instantiation each file is a
markdown document that is required to open with structured frontmatter
carrying its name and the one-line description $d_f$, and that may carry
additional optional frontmatter fields (free-form metadata), mirroring the
industry-default memory tool \citep{anthropic2025memorytool}.
\Cref{fig:filezoom} dissects two files of one such store. Nothing below
depends on this choice, and any hierarchical file store with per-file
descriptions satisfies the definitions.\footnote{For simplicity we take each
memory to be a single markdown file. Richer packagings exist: in Anthropic's
skill definition, a skill is a folder containing a \texttt{SKILL.md} together
with optional subfolders of auxiliary context or scripts
\citep{anthropic2025skills}. Our definitions extend naturally to such
folder-valued memories, but we do not study that variant.}

\paragraph{The taxonomy contract.}
What should this tree look like? Both settings' management instructions
state the same five properties (\Cref{app:prompts-builder,app:skill_prompts}),
and we adopt them as the paper's working definition of a well-organized
store, stated as five principles the tree must satisfy:
\begin{enumerate}[label=\textbf{P\arabic*.}, leftmargin=2.4em, itemsep=1.5pt, topsep=2pt]
\item \textbf{Sibling distinction.} \emph{Siblings are distinguishable by
  labels alone}: items under one parent can be told apart by name, at worst
  name plus description, without opening bodies.
\item \textbf{Sibling relatedness.} \emph{Siblings belong together}: what
  shares a parent is related enough that sharing it is natural.
\item \textbf{Parent-child coverage.} \emph{A parent covers its children}:
  everything under a parent falls within what its name declares and, as far
  as practical, everything in its scope lives under it, so descending
  narrows the search without losing the sought fact, an exhausted subtree
  is conclusive, and each child is strictly more specific than its parent.
\item \textbf{Tree-wide proximity.} \emph{Distance mirrors relatedness}:
  the more related two pieces of content, the nearer they sit in the tree.
\item \textbf{Structural economy.} \emph{Structure serves the search, not
  itself}: depth is added only where it improves routing to a fact; a level
  that does not help routing is overhead.
\end{enumerate}
These principles bind at every level of the tree, headings included; the
hierarchy metrics of \Cref{app:hierarchy_metrics} operationalize them, and
\Cref{sec:results-rq1} reads the stores agents actually build against this
contract.

\subsection{Tool harness and agent roles}
\label{sec:setting-roles}

\paragraph{Tool harness.}
Agents never manipulate $\gM$ directly; every access passes through a
\emph{tool harness} $\gH$, a finite set of operations
$o : (\gM, \mathrm{args}) \mapsto (\gM', \omega)$ that return a text
observation $\omega$ and may mutate the store (read operations leave
$\gM' = \gM$). Harnesses differ in granularity and power. We consider a
sandboxed shell over the store's directory; a function set
mirroring the industry-default memory tool's six operations (view, create,
string-replacement and insertion edits, delete, rename); and
search-augmented variants that add line-level regex search or ranked
keyword search. The harness is a first-class experimental axis:
the same store under a different $\gH$ affords different behavior.

\paragraph{Agents.}
An agent binds an LLM policy $\pi$ to the harness: given its input, it runs a
tool loop, a sequence of operation calls and observations through $\gH$, and
terminates by emitting its output. Three roles share this form and differ only
in their contracts.

\paragraph{Management agent.}
The management agent integrates new content into the store. Given an
instruction $\iota$, an incoming chunk $x_t$, and the current store, it
produces the next store:
\begin{equation}
  \label{eq:update}
  \gM_t \;=\; \mathrm{m}^{\gH}_{\pi}\!\left(\iota,\, x_t,\, \gM_{t-1}\right),
  \qquad \gM_0 = \varnothing ,
\end{equation}
so a stream $x_1, \dots, x_T$ induces a store trajectory
$\gM_1, \dots, \gM_T$. The mandate is integration \emph{and} maintenance: the
agent may create, rewrite, merge, split, move, or delete anything in the
store, so organization is part of its job rather than a side effect. How well
this mandate is exercised as $T$ grows is a central object of study.

\paragraph{Search agent.}
The search agent answers queries over a fixed store. Given an instruction, a
query $q$, and a store, it returns an answer with citations:
\begin{equation}
  \label{eq:search}
  (a, \Gamma) \;=\; \mathrm{s}^{\gH_r}_{\pi'}\!\left(\iota,\, q,\, \gM\right),
\end{equation}
where $\Gamma$ is a set of references into $\gM$ (file paths, optionally
sections or lines) supporting $a$, and $\gH_r$ is the harness available to
search. Search is read-only in intent: the store an answer is graded against
must be the store that was searched. Depending on the harness this is enforced
structurally (an operation set without writes) or only by instruction (a
write-capable shell told not to modify the store).

\paragraph{Execution agent.}
In the skill setting a third role appears: an execution agent
$\mathrm{e}(\tau, \Gamma) = (\xi, z)$ attempts a task $\tau$ given a set
$\Gamma$ of retrieved skill files, whose contents are placed in its context,
and returns its trajectory $\xi$ and a success signal $z \in \{0,1\}$. It is
the probe that converts store quality into task outcomes.

\paragraph{How the roles compose.}
When an execution agent is present, the other two serve it: the search agent
acts as its retrieval subroutine, and its trajectories become the chunks the
management agent integrates, whether the \execagent{} invokes management directly
or its logged traces are handed over after the fact. When no execution agent
exists, as in the conversational instantiation, management and search operate
on their own: the stream comes directly from the environment and queries are
posed externally. The contracts are deliberately minimal so the roles map onto
deployed harnesses: in a coding agent, one model may play all three, and the
instruction $\iota$ can be a fixed constant. The roles isolate the two
capabilities this paper studies, writing memory well and reading it well,
without prescribing how either is implemented.

\subsection{Two instantiations, one store}
\label{sec:setting-instantiations}

\paragraph{Conversational memory.}
The stream $x_1, \dots, x_T$ consists of contiguous slices of a long
multi-session dialogue, each turn tagged with an inline source locator, and
the management agent builds $(\gM_t)$ by \Cref{eq:update}. Evaluation asks
questions about the conversation, posed against the final store or against
intermediate $\gM_t$ at checkpoints along the stream: an answer is graded for
correctness against gold, and its citations $\Gamma$ for whether the cited
memory supports it. We instantiate this setting on \locomo, \personamem, and
\realtalk{} \citep{maharana2024locomo,jiang2025personamem,lee2025realtalk}.

\paragraph{Procedural memory (skills).}
A group of $K$ tasks, not necessarily related, is attempted in sequence, and
the store holds \emph{skills}, files describing reusable procedures, and may
also hold \emph{notes}: lessons, observations, and warnings distilled from
attempts; each file's description states what it contains and when to apply
it. For task $\tau_i$ the retrieval,
attempt, and curation steps are
\begin{equation}
  \label{eq:skill}
  \Gamma_i = \mathrm{s}^{\gH_r}_{\pi'}(\iota_r, \tau_i, \gM_{i-1}), \qquad
  (\xi_i, z_i) = \mathrm{e}(\tau_i, \Gamma_i), \qquad
  \gM_i = \mathrm{m}^{\gH}_{\pi}(\iota_c, \mathrm{render}(\xi_i), \gM_{i-1}),
\end{equation}
where the search agent acts as the retriever, its citation set being its
output (we write $\mathrm{s}$ for that component), $\iota_r$ and $\iota_c$ are
fixed role instructions, and $\mathrm{render}(\xi_i)$ serializes the
trajectory into a chunk. The protocol is leak-free by construction: $\tau_i$
is attempted with a store built only from earlier tasks, so performance on
later tasks measures what the store transfers. We instantiate this setting on
\alfworld{} \citep{shridhar2021alfworld}.

Both instantiations share the store class, the contracts of
\Cref{eq:update,eq:search}, and the harnesses; they differ only in what a
chunk is (a dialogue slice or a rendered trajectory) and what a query is (a
question or a task). One system thus serves declarative and procedural memory.

\subsection{What is held fixed and what varies}
\label{sec:setting-axes}

In all experiments the management and search agents are the objects of study,
while the execution agent is held fixed. The contracts above likewise stay
fixed; the study varies four components:
the \emph{builder} that produces the store (the management agent of
\Cref{eq:update}; a deterministic verbatim dump of the stream into session
files; or a mechanical split of the stream into indexed raw chunks, the store that retrieval-augmented generation reads; only a closed-book
control keeps no store at all); the
\emph{scale} $T$ of the stream; the \emph{harness} $\gH$ available to each
role; and the \emph{strengths} of the policies $\pi$ and $\pi'$ that build and
search. All measurement attaches to the store trajectory $(\gM_t)$ and its
use: downstream answer quality, the cost of building and searching, and the
health of the store itself over time. Concrete configurations and metrics are
specified in the experimental setup.

%% file: figures/fig1_overview.tex
\begin{figure}[t]
  \vspace{-10pt}%
  \centering
  \setlength{\abovecaptionskip}{4pt}%
  \includegraphics[width=\textwidth]{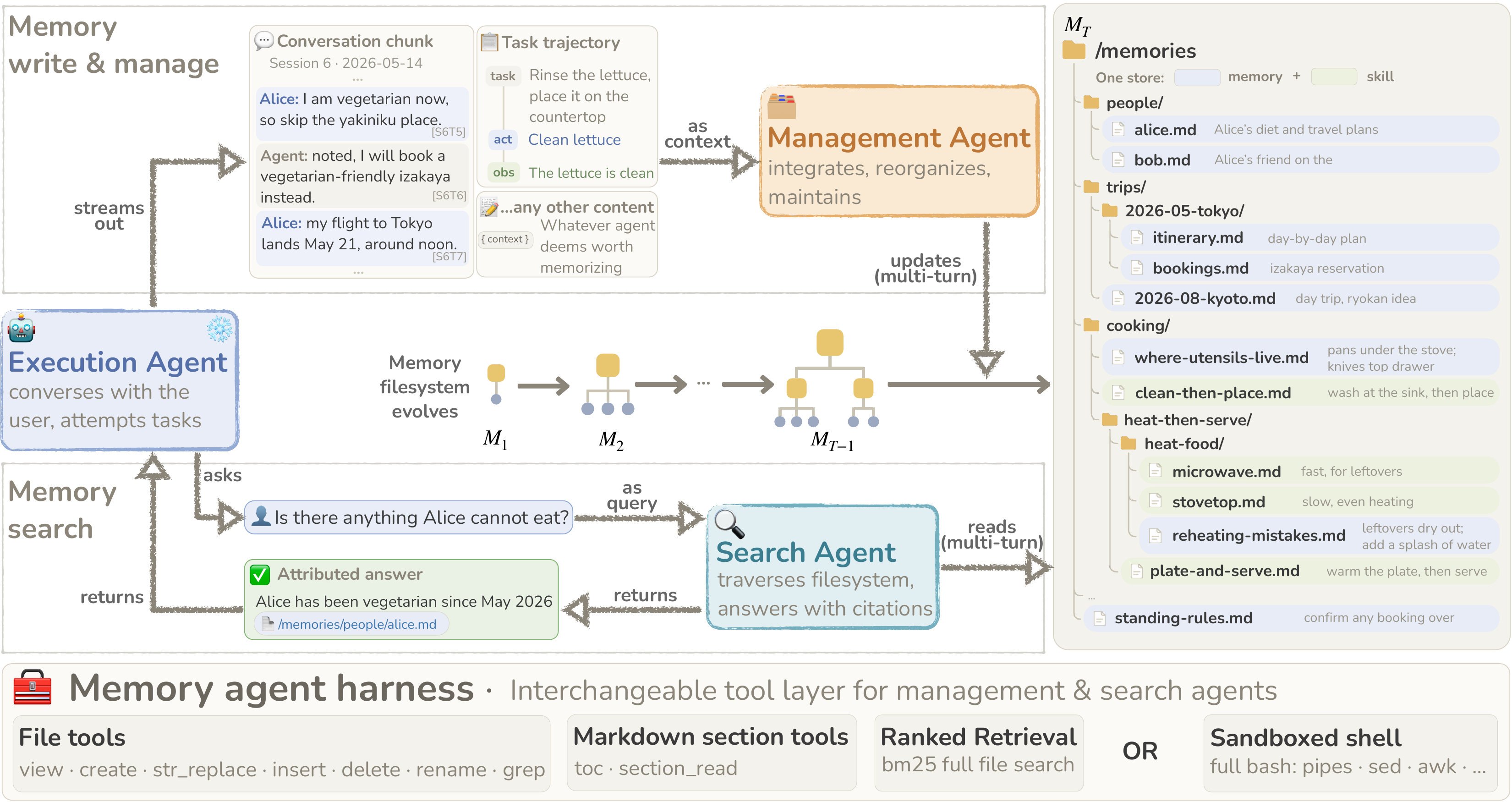}
  \caption{Overview of filesystem-based agent memory. An execution agent does
  the work; what it experiences or chooses to save (conversation slices, task
  trajectories, or any other content) streams out as chunks into a
  \emph{\mgmtagent} that integrates each chunk into one memory filesystem,
  serving declarative memory and skills alike, and keeps it organized; when
  the agent needs to remember, it asks a \emph{\searchagent}, which traverses
  the store and returns attributed answers, cited to the store, or retrieved skills. The management and
  search agents act on the store only through an interchangeable tool
  harness, and the store's evolution over the stream is tracked. The
  execution agent is optional and need not invoke the management agent
  directly (its logged content can be handed over).}
  \label{fig:overview}
\end{figure}

%% file: figures/fig2_filezoom.tex
\begin{figure}[t]
  \begin{minipage}[c]{0.42\textwidth}
    \caption{Two files of one memory filesystem. A declarative memory file
    (top) and a skill file (bottom) share a single anatomy: YAML frontmatter
    (\texttt{name}, \texttt{description}) is what listings and
    search expose first;
    markdown headings nest, continuing the taxonomy within the file; facts
    carry inline source locators (\texttt{[S6T5]} reads session~6, turn~5);
    repeated content is cross-referenced rather than copied; and the file
    kind sits in an optional \texttt{metadata} field.}
    \label{fig:filezoom}
  \end{minipage}\hfill
  \begin{minipage}[c]{0.55\textwidth}
    \centering
    \includegraphics[width=\linewidth]{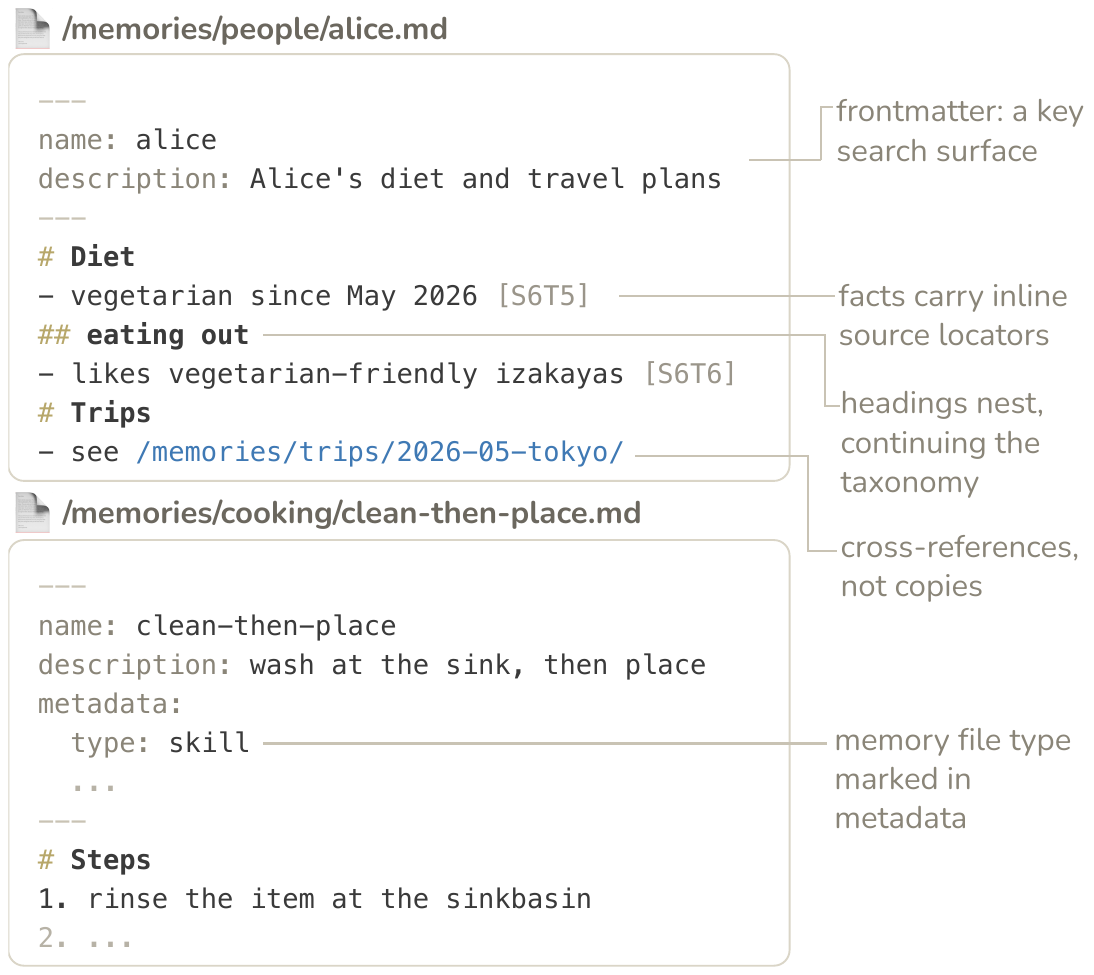}
  \end{minipage}
\end{figure}

%% file: sections/04_experimental_setup.tex
\section{Experimental Setup}
\label{sec:setup}

\paragraph{Benchmarks.}
\locomo{} \citep{maharana2024locomo} pairs long multi-session dialogues with
questions about them and is the field's default long-memory testbed; we evaluate on a
held-out test conversation, 158 questions spanning four categories: multi-hop, temporal reasoning, open-domain, and single-hop (the adversarial
category excluded). Four questions whose gold answers contradict the
transcript are cataloged in \Cref{app:locomo-defects}; excluding them
changes no conclusion. \personamem{} \citep{jiang2025personamem} poses
multiple-choice questions at checkpoints along a persona-rich stream in two
context tiers, 32k and 128k tokens; answers are graded by exact match, with
no judge model involved. Comparing across the tiers is only a rough scale test, since the
tiers differ in which conversations they contain, not only in length. Both tiers are
evaluated at each conversation's terminal checkpoint (32k: three
conversations, 32 questions; 128k: one conversation, 42 questions). \realtalk{}
\citep{lee2025realtalk} contributes 21 days of authentic human-to-human
messaging (one test conversation, 85 questions), testing whether conclusions
survive naturalistic dialogue.

\paragraph{Memory variants.}
Six memory variants span the space from no store to an agent-curated one.
Only \subclosed{} keeps no store; every other variant builds one from the
same conversation, and the four filesystem variants are searched under the
contract of \Cref{eq:search}.
\textbf{\subclosed} answers each question directly, with no store and no
retrieval, measuring the floor the model reaches on its own knowledge. \textbf{\subrag} stores the stream itself: the conversation is split into
chunks (chunking rule in \Cref{app:campaign_settings}), one file per
chunk, indexed for ranked keyword search (BM25), and
its search agent has a single tool that returns the top 3 chunks in full
for each query it issues, the standard retrieval-augmented control
\citep{lewis2020rag}. \textbf{\subdump} copies the stream into the store wholesale: one file per
session, content verbatim, each file's description stating only the
session's number, date, and speakers; a deterministic flat store built at
zero model cost, the zero-curation baseline.
\textbf{\subfoldered} has an LLM design a folder taxonomy into which the
\subdump's session files are moved intact (move only, zero byte edits), so
the taxonomy is the only signal added over the \subdump. \textbf{\subreorg}
has an LLM restructure the \subdump, splitting and merging content across
sessions
into files whose grouping the model chooses, so the layout changes. We run
it as two versions that differ by a single instruction, which lets us separate
compression from layout. Its \emph{default} behavior turns out to be lossy: on
\locomo{} we observed that without an explicit rule the store shrinks markedly
across the reorganization pass, the model dropping detail as it rewrites. We
label this default the \subreorgcond{} version; adding one ``keep every fact''
rule gives the \subreorgpres{} version, which holds content roughly fixed. The condensing version therefore captures not an instruction to compress but the backbone model's own tendency when it reorganizes freely, and the preserving version is the
intervention that counteracts it. \textbf{\subcurated} is built from empty over the stream by
the \mgmtagent{} of \Cref{eq:update}, which decides both what to write and
how to organize it; unlike the constrained variants above, it varies
content and organization together.

\paragraph{Models and judge.}
The \mgmtagent{} and \searchagent{} roles run \texttt{gpt-5.4-mini} at high
reasoning effort, the center of a \texttt{gpt-5.4-nano} to \texttt{gpt-5.4}
strength ladder; the other two models appear in the strength studies below. The
foldering and reorganization passes also run \texttt{gpt-5.4-mini} at high
effort. Grading uses one judge model with one fixed prompt throughout:
\texttt{gpt-5.4-mini} at low reasoning effort and temperature 0. Before any experiment ran, we calibrated the judge on held-out
validation answers and froze its configuration for every judged cell.

\paragraph{\Searchagent{} prompts.}
Pilot runs showed the search prompt alone can reorder memory variants, so
all variants run one prompt family built from the same shared blocks,
adapted only where the store form requires it (a raw-session reading note
for flat stores, for example). Every prompt requires the answer to be drawn from what the
store contains and cited to it, never from the \searchagent's own
knowledge, and is fixed across all experiments; the full texts appear in
\Cref{app:prompts}. Filesystem variants cite file paths with line ranges;
\subrag{} cites the turn locators its chunks carry.

\paragraph{Tool sets.}
The two roles see deliberately different harnesses. The \mgmtagent{} writes
through the six generic file operations of the industry memory tool
\citep{anthropic2025memorytool} (view, create, string-replace edit, insert,
delete, rename) plus line-level regex search: a write tool set faithful to
deployed practice. The \searchagent{} reads through
a filesystem-native read-only set (view, regex search, table of contents,
section read): tools that answer only what the store's own names, layout,
and text expose. A ranked search engine would find content however the
store is organized, hiding exactly the differences we study; file tools
make the \searchagent{} navigate the organization, so those differences can
show. This asymmetry is a design choice, not a finding; the harness study
(\Cref{sec:results-rq5}) varies both tool sets. Every tool's parameters and
description, exactly as sent, appear in \Cref{app:tool_schemas}.

\paragraph{Strength, harness, and scale studies.}
Each study changes one thing relative to the setup above and holds
everything else fixed. The \textbf{builder ladder} adds two rebuilds of the \personamem{} 128k
store, with \texttt{gpt-5.4-nano} and \texttt{gpt-5.4} as the \mgmtagent{}
beside the main \texttt{gpt-5.4-mini} build, while the
stream, the instruction, and the \searchagent{} stay the same. The
\textbf{searcher ladder} then has each of those three models read each of the
three stores: a three-by-three grid whose center cell is the main setup.
The \textbf{harness study} keeps the agent-curated store setup and the model,
and changes only the tools through which memory is written and read; the
three tool sets carry the same names in both settings: \textbf{Center}, the
default tools above; \textbf{Center{+}BM25}, which adds ranked keyword
search; and \textbf{Shell}, which replaces the tools with a bash shell over
the same store. \textbf{Scale} is compared through \personamem's 32k and 128k
tiers. The skill setting repeats the strength axes: its curator ladder swaps
the \mgmtagent's model while the \execagent{} and \searchagent{} stay the same, and its
two \execagent{} tiers vary execution strength; and its two protocols give the
scale contrast, the same tasks run with stores that accumulate over one
140-task chain against three shorter chains
(\Cref{sec:results-rq4}). Every study records its configuration
like the main runs; the remaining details and the numbers appear with the
results (\Cref{sec:results-rq3,sec:results-rq5}).

\paragraph{Metrics.}
Outcome quality is measured per setting. On the conversational benchmarks
each answer is graded for \emph{correctness} against gold and for
\emph{attribution}, whether the cited memory actually supports it, both by
the judge; on \personamem{} correctness is exact match over the options. On
the skill setting the environment itself reports task \emph{success},
read as overall and per-family rates. Cost and effort are counted per model
call in token categories (uncached input, provider-cached input, output
with reasoning as a sub-count) together with tool rounds and calls,
aggregated per query, per chunk, or per task; every input token is tallied whether or not the provider served it from cache, and
\Cref{tab:cost,tab:cost_build} price the totals at stated rates. From the
same per-call records we compute each cell's intrinsic compute bounds, the
perfect-cache floor and the no-cache ceiling that bracket its measured
spend (\Cref{app:cost_accounting}). Stores are measured directly: file,
folder, and section counts, sizes, and cross-references; the tree-shape and
taxonomy-adherence metrics of \Cref{tab:shape-panel}; and, under the full-chain
protocol, the per-task store trajectory. Definitions and equations for the
compute bounds and the shape metrics are in
\Cref{app:cost_accounting,app:hierarchy_metrics}. Where two variants are
compared on the same items we report paired differences with sign tests.

\paragraph{Skill setting.}
The skill setting of \Cref{sec:setting-instantiations} runs on
\alfworld{} \citep{shridhar2021alfworld}: 140 held-out household tasks
spanning six goal families, attempted in sequence under the leak-free
protocol of \Cref{eq:skill}, each task retrieving from a store built only
from earlier attempts. Two evaluation protocols are used: a
\emph{three-chain} protocol runs three independent chains, each interleaving
tasks from all six families, with stores reset between chains, and a
\emph{full-chain} protocol runs all 140 tasks as one
family-interleaved chain in a fixed order, so the store accumulates
end-to-end and every store trajectory is snapshotted per task. Five memory
variants span the design space. \textbf{\skillnone} attempts every task
without memory, the \execagent's floor. \textbf{\skillflat} appends each
task's rendered trajectory to the store verbatim at zero model cost, the
zero-curation baseline; retrieval serves whole episode files.
\textbf{\skillcur} has the \mgmtagent{} of \Cref{eq:update} distill
attempts into procedure files, and retrieval cites the files whose contents
are placed in the \execagent's context. \textbf{\skillgated} adds a second
entry kind to the curated store: alongside skill files it holds \emph{memory
notes} (lessons, warnings, observations), with an outcome gate (only a
successful attempt may create or extend a positive procedure) and a
stricter rule for what retrieval may serve: entries are cited only when they clearly apply to the task, rather than liberally. \textbf{\skillsynth} keeps that store and changes what
retrieval returns: instead of serving files, the \searchagent{} reads the
store and writes task-specific guidance (\emph{guidance synthesis}, GS), so
the \execagent{} sees synthesized text rather than raw entries. The \mgmtagent{} and \searchagent{} run
\texttt{gpt-5.4-mini} at high effort throughout, matching the conversational
roles; the \execagent{}, held fixed within each cell, runs at temperature 0
with a 50-step cap and the environment's own success signal as $z$, and we
report two \execagent{} tiers, \texttt{gpt-4.1} and \texttt{gpt-4.1-mini}, so
execution strength is varied explicitly. Grading needs no judge (the environment itself reports success); cost
uses the same token framework, split into deployment (retrieval plus
execution) and build (curation).

%% file: sections/05_results.tex
\section{Results and Analysis}
\label{sec:results}

Deployed practice stores agent memory as a filesystem and trusts the agent
to keep it organized. \Cref{sec:setting} reduced that practice to three
roles around one store, and \Cref{sec:setup} instantiated it twice, for
conversational memory and for skills, where a fixed \execagent{} turns store
quality into task success. That default rests on two untested
assumptions: that an agent can keep a growing store organized, and that
organizing pays. The results test them from five angles:
\begin{itemize}[leftmargin=1.6em, itemsep=1pt, topsep=2pt]
\item \textbf{RQ1 (organization).} What structures do management agents
actually grow when left to organize, and which behaviors are characteristic
or degenerate?
\item \textbf{RQ2 (value of shape).} Does the store's shape change what
the \searchagent{} answers, and at what search cost?
\item \textbf{RQ3 (backbone model capability).} Do building and reading memory track the
model's own strength?
\item \textbf{RQ4 (sustainability under scaling).} As the store grows, does it stay
useful, affordable per use, and healthy?
\item \textbf{RQ5 (harness).} How does the tool set through which agents
touch memory shift everything above?
\end{itemize}
Each question is answered in both settings, the conversational first.

\input{tables/tab_main_comparison}
\input{tables/tab_cost}
\input{tables/tab_cost_build}

\subsection{What stores agents build (RQ1)}
\label{sec:results-rq1}

What does the \mgmtagent{} build when left to organize? Each
\subcurated{} cell of \Cref{tab:main} grows its store from empty over its
benchmark's stream, and \Cref{tab:cost_build} profiles the resulting
stores: directories, files, markdown sections, kilobytes. Under the same gpt-5.4-mini \mgmtagent{}, all four stores organize by subject,
but as trees of very different form. \locomo{} puts nearly all its
hierarchy inside files: three files, one per speaker plus a shared thread,
whose 116 headings carry the tree to depth five, cross-linked by 151
in-store \texttt{see /memories/...} pointers. \realtalk{} does the
opposite, hierarchizing at the file layer: 41 topic files in two folders,
most nearly flat inside at two sections apiece. \personamem{} 32k grows a
modest tree at every level, about three folders holding twelve files of a
few sections each; and \personamem{} 128k takes the \locomo{} form to its
extreme, two files whose 210 sections nest to depth six, branching eleven
ways at an average step (\Cref{tab:shape-panel}). Where the hierarchy
lives thus changes with the benchmark, and no single count, files or
sections, says how organized a store is: is that shape driven by the
material or by the model?

Two controlled comparisons separate the model from the scale. Hold the
\mgmtagent{} at mini and move within one benchmark family from \personamem{}
32k to 128k, a stream roughly five times longer (about 55 to 271 chunks):
the folder and file layers thin, about three folders of twelve
files becoming a single folder holding two files, while the section layer
quadruples, 53 headings to 210. The store does not shard with scale; it
\emph{consolidates}, relocating its hierarchy from folders and files into
headings. Hold the scale at 128k and vary only the \mgmtagent{} across the
ladder of \Cref{sec:results-rq3}: the same conversation becomes 122 files
in twelve folders under nano, two files under mini, and 105 files in four
folders under gpt-5.4, an order-of-magnitude swing in file-level
fragmentation, and equally a swing in where hierarchy lives: nano spreads
a shallow forest (no leaf deeper than five levels), gpt-5.4 grows the
panel's deepest tree (depth seven), and mini nests nearly all structure
as headings inside its two files. \Cref{fig:store-examples} shows the
three shapes in the stores' own paths and headings, around one subject.

\input{figures/fig_store_examples}

\textbf{What might read as structure growing with scale is therefore
primarily the \mgmtagent's own behavior, not a scale law};
comparisons across benchmarks are further confounded with content type
(conversational versus persona). Linking is a model tell of the same kind:
where structure spans files it is knit by cross-reference, in the
\texttt{see /memories/...} form the instruction prescribes, and at 128k
the gpt-5.4 store stitches its 105 files with 233 pointers against mini's
seven. Because the model ladder runs at a single stream length, we cannot yet map the full model-by-scale interaction;
mapping it is what the checkpointed growth study of \Cref{sec:results-rq4}
is for.

The constrained variants behave characteristically too. On \personamem{} 128k the \subfoldered{} pass groups the \subdump's
twenty session files into ten topic directories it invents, skewed from
four-session folders down to singletons; the other benchmarks get three to
six directories. The reorganizer exposes the clearest degenerate behavior
we observed: asked only to restructure, it condenses, dropping recorded
detail as it rewrites, and holds content roughly fixed only when one
``keep every fact'' rule is added; the two versions in \Cref{tab:main}
exist to measure exactly this tendency, and \Cref{sec:results-rq2} shows
what it costs in answers.

The skill setting asks the same question over procedures, and its \mgmtagent{}s
organize differently. Under the same fixed mini \mgmtagent{} the full-chain
store consolidates to 45 files; varying only the \mgmtagent's backbone turns
shape into a capability signature that runs monotonically, gpt-5.4-nano
sprawling to 114 files, largely flat with a few topic folders, mini's 45,
and gpt-5.4 distilling 16 dense
entries, against the mechanical \skillflat{}'s 140 episode files
(\Cref{tab:skill-growth}).

Unlike the conversational stores, whose consolidation relocates structure into headings, the skill stores compact at every level of the hierarchy, sections included: the
gpt-5.4 store's 16 files carry just 43 sections, where the conversational
mini build packed 210 into two files. The skill trees are also uniformly
shallow, none exceeding four levels at any \mgmtagent{} model
(\Cref{tab:shape-panel}), so capability expresses itself here as file
granularity, not as depth. The two entry kinds are used in a capability-dependent mix: the gpt-5.4
store holds five skills, one per goal family with two-object placement
folded into the general procedure, outnumbered by its eleven warning
notes, while nano's 114 files are skill-heavy, 89 skills beside 25 notes.

Set against the conversational ladder the shape story sharpens:
\textbf{capability changes what agents build decisively in both settings,
but not in one direction}. The conversational ladder fragments non-monotonically at the file level
(122, 2, 105 files) and stays non-monotone counted by sections (413, 210,
561), while the procedural one consolidates monotonically at both levels
(114, 45, 16 files; 290, 99, 43 sections).

\input{tables/tab_shape_panel}

Both settings' management prompts prescribe the same taxonomy contract
(\Cref{app:prompts-builder,app:skill_prompts}), so one set of tree metrics,
with headings counted as levels, quantifies every store
(\Cref{tab:shape-panel}; definitions in \Cref{app:hierarchy_metrics}).
\textbf{File counts misdescribe organization}: counted as a tree, the
``two-file'' 128k store is depth 6 with a mean fanout of 11, deeper than
most many-file stores. Every store carries positive
distance-mirrors-relatedness (B4), so layout tracks content everywhere, but
the correlation is a descriptor, not a quality score (\Cref{tab:shape-panel}): on the conversational side the ladder's sharded builds score higher than the consolidated two-file store, and on the skill side the
Shell store reaches the panel's highest value (.37), the same store that
wins on outcomes in \Cref{sec:results-rq5}. Scope leakage flags the
\locomo{} store's cross-file bleed (15.7 percent of its sections sit
lexically nearer a sibling file), consistent with its heavy reliance on
cross-references instead of separation.

Whether these varied stores preserve enough detail to answer from is the
retrieval question of \Cref{sec:results-rq2}. Each store is a single build
draw, and reasoning-model builds carry real run-to-run shape variation (a rebuild identical but for a corrected tool description, whose behavioral impact trace audits measured as null, moved one store from 2 files to 29;
\Cref{app:rq5-detail}), so we read file counts as characteristic behaviors
rather than fixed constants; exemplar trees appear in
\Cref{app:example_filesystems}.

\subsection{Memory shape and retrieval (RQ2)}
\label{sec:results-rq2}

Does the store's shape change what the \searchagent{} answers, and at what cost?
\Cref{tab:main} compares all six memory variants across the four
benchmarks. Closed-book scores below every store, far below on the conversational
benchmarks (18.4 on \locomo{}, 10.6 on \realtalk{}), and higher only on
\personamem's multiple-choice tiers (34.4 and 47.6), where guessing over
options lifts the floor: the questions need the stored conversation, not
parametric knowledge.
Attribution is uniformly high, 82 to 98 percent, so any filesystem store
supplies verifiable support regardless of its shape.

\textbf{Correctness has no shape that wins everywhere}, and the ordering
is not the one a ``more organization is better'' prior would predict. The cheapest structured
store, \subfoldered, is the most consistent leader: it ties or tops
\locomo{} (86.1), \realtalk{} (77.6), and \personamem{} 128k (76.2), whereas
the \subcurated{} leads only \locomo{} and is the weakest
store of all on \personamem{} 32k (37.5, against the \subdump's 78.1; the
representation-level analysis of this failure is in
\Cref{app:error-analysis-pm32k}). On that 32k tier the \subdump{}
leads outright, and even \subrag{} (71.9) beats every reorganized or
curated store there, while on the conversational benchmarks \subrag{} trails every variant except the condensing reorganizer: one more sign that no
single form dominates.\footnote{Read the per-tier orderings with care:
columns rest on 32 to 158 questions, and re-judging a fixed cell moved
about two questions ($\pm 1.3$ points; \Cref{tab:main}), so gaps of a few
questions on the small tiers are suggestive rather than settled.}

\textbf{Where organization pays off unambiguously is retrieval cost}, and
the payoff tracks the size of the raw material. On the two \personamem{} tiers, whose
content is the largest and densest (137 and 612 kilobytes when dumped
verbatim, against 102 and 105 for the conversational benchmarks), the
reorganized and curated stores cut the \subdump's per-query search price by half or more (\personamem{} 32k: 1.4 cents for either reorganizer against 4.0
for the \subdump; 128k: 1.6 against 3.9). On the smaller conversational benchmarks
the stores and the \subdump{} search at near parity (2.1 to 2.5 cents on \locomo{}).

The mechanism, visible in \Cref{tab:cost}, is an effort trade: on the
\personamem{} tiers the reorganized and curated stores spend more search
rounds and tool calls than the \subdump{} (seven to nine calls against its
four to five), because the \searchagent{} navigates structure rather than
scanning, yet each targeted read pulls far fewer tokens, so the total
falls wherever the \subdump{} is large.
Structure buys search economy, most on the material that needs it.

The two reorganizer versions pull apart by benchmark rather than uniformly.
Preserving every fact helps on the conversational streams and the long
\personamem{} 128k (\locomo{} 82.9 against the condensing version's 79.1,
\realtalk{} 77.6 against 41.2, 128k 59.5 against 54.8), where dropping
detail costs answerable content, most starkly on \realtalk{}, whose
correctness nearly halves under condensation. Condensing instead helps on
\personamem{} 32k (68.8 against 56.2), where a tighter store is easier to
search. That one reorganizer flips sign with the benchmark, halving \realtalk{} correctness when it condenses, is the clearest sign that \textbf{no single compression policy is right across
benchmarks}.

\input{tables/tab_skill_main}

\paragraph{The skill setting asks the same question with a harder probe, and
the winner depends on who consumes the memory.} Under the
stronger \execagent{} the verbatim \skillflat{} leads (87.1\%), with
\skillsynth{} second (82.1\%); under the weaker \execagent{} the order inverts,
\skillsynth{} leading by ten points (76.4 against 66.4;
\Cref{tab:skill-main}). Task-paired comparisons give the two sides of the inversion different statistical support: at
\texttt{gpt-4.1} the \skillflat's edge over GS is suggestive at best (net $+7$ of 140 tasks, two-sided sign test $p\!\approx\!0.23$), while at \texttt{gpt-4.1-mini}
GS beats both the \skillflat{} (net $+14$, $p\!\approx\!0.02$) and the
\skillcur{} store (net $+15$, $p\!\approx\!0.01$).

The slopes tell the mechanism: dropping the \execagent{} tier costs the \skillflat{} 21 points but GS
only 6, and the \execagent's invalid-action rate, measured from the run records,
rises under the \skillflat{} from 19 to 36 percent while GS rises only
from 12 to 22, so \textbf{raw episode transcripts demand an \execagent{} strong enough to digest
them, whereas guidance written for the task degrades gracefully}.

Two corollaries follow. Memory is worth more to
the weaker \execagent{} (best lift over \skillnone{}: $+23.5$ points against the stronger tier's $+13.5$), exactly where deployment is also cheapest. And the protective strictness of \skillgated{}
is tolerable only under a strong \execagent{}: already a regression at \texttt{gpt-4.1}
($75.7$ against 80.7 for \skillcur), it collapses to two points above the
no-store floor at mini, while \skillcur{}, which serves files liberally,
lands with the \skillflat{} there; only GS survives the capability drop.

\paragraph{\skillsynth{} is also the cheapest store to deploy.} Deployment cost
per task (\Cref{tab:skill-main}) puts \skillsynth{} below every other
store on both tiers, because its compact store keeps retrieval cheap and its
guidance shortens episodes. Running with \skillnone{} is still cheaper per
task at the mini tier (3.2 against 4.3 cents), but counted per solved task
GS undercuts it (5.7 against 6.1), so \textbf{the memory pays for itself in deployment}. Building it is the extra bill (about 10 to 11 dollars per 140
tasks at each tier, against zero for the \skillflat), the same
curation-versus-search trade the conversational setting prices. Where the
conversational setting's answer was that organization's value is conditional
on the material, the procedural setting's is that it is conditional on the
consumer; \textbf{both halves of RQ2 refuse an unconditional winner}.

\subsection{Backbone model capability (RQ3)}
\label{sec:results-rq3}

Does memory management improve with the \mgmtagent's backbone strength? We compare three \mgmtagent{} backbones on the \personamem{} 128k store,
adding gpt-5.4-nano and gpt-5.4 rebuilds beside the main gpt-5.4-mini
build while holding the
stream, the instruction, and the gpt-5.4-mini \searchagent{} fixed
(\Cref{tab:ladder}). \textbf{The \mgmtagent's strength expressed itself almost
entirely as organizational style and almost not at all as answer quality.}

Structurally the three stores share nothing: nano fragments into 122
files across twelve folders, mini consolidates into two files that nest
210 heading sections, and gpt-5.4 shards into 105 files nested seven
levels deep and stitched by 233 cross-references, thirty times mini's
linking. Yet correctness sits in a seven-point band that
is not even monotone in model strength (73.8 for nano, 66.7 for mini, 71.4 for
gpt-5.4), a spread of three questions out of 42, comparable to the one-to-two
questions that re-runs move on this same conversation
(\Cref{sec:results-rq5}). The fixed \searchagent{}, in effect, absorbed an
order-of-magnitude difference in store shape into a null difference in
answers.

Build effort tells the same non-monotone story: the strongest model
works the hardest (3{,}921 tool calls against mini's 1{,}986) and builds the store that is dearest to search (3.8 against 2.6 cents), without out-answering the weakest. On this one
conversation, then, a stronger \mgmtagent{} buys a more elaborate store, not a more
useful one, once the \searchagent{} is held fixed and the instruction says what
structure is for.

The \searchagent's side of the same probe answers what the builder ladder left
open: varied over the same three fixed stores (\Cref{tab:searcher-grid}),
the \searchagent{} is monotone where the \mgmtagent{} was flat, mean
correctness rising from about 62 under the nano \searchagent{} through 71 under
mini to 79 under gpt-5.4. \textbf{Reading, not writing, is where backbone
strength pays on this benchmark.} And the stronger \searchagent{} does not
rescue the elaborate store: gpt-5.4 reads the nano-built sprawl best (83.3)
and the ornate gpt-5.4-built store worst (71.4), so the 105-file,
233-pointer construction stays unrewarded under every \searchagent{} we tried.
Under the weak \searchagent{} the three stores' scores compress into a low band (59.5 to 64.3), differences flattening exactly where capability is scarcest.

The grid's cost columns locate the price of that capability. The
\searchagent's strength, not store shape, is the dominant cost axis: a query costs under
a cent at nano, two to four cents at mini, and ten to nineteen at
gpt-5.4, while within any one \searchagent{} the store moves the bill by at
most a factor of two. That factor is still informative:
\textbf{the consolidated two-file store nearly halves the strong
\searchagent's per-query price against the sprawl (10.0 against 18.5 cents) at
statistically close correctness (81.0 against 83.3)}, and it draws the
fewest tool calls from that \searchagent{} (7.6 per query against 9.9).
Organization pays the strongest \searchagent{} in the same currency, search
cost, that \Cref{sec:results-rq2} found on the conversational
benchmarks' variants.

All of this remains one 42-question
conversation, so gaps of a few points are suggestive; but the asymmetry,
\textbf{the \mgmtagent's capability decoupled from quality, the
\searchagent's strongly coupled}, is the cleanest single fact this study has produced
about the conversational setting. The asymmetry also has a
measured boundary: on the 32k conversations where curation collapses
through build-time defects, the same \mgmtagent{} upgrade does move the fixed
\searchagent{}, from 12 of 32 questions to 18 of 32
(\Cref{app:error-analysis-pm32k}). The \mgmtagent's strength lies unused where the store already serves its \searchagent{}, and pays exactly where building is the failing step.

\input{tables/tab_ladder}
\input{tables/tab_searcher_grid}
\input{tables/tab_skill_growth}
\input{figures/fig_growth_s2}

\paragraph{The same probe on the skill \mgmtagent{} finds a threshold, and it
shows in the store.} Varying only the \mgmtagent's backbone under a fixed
\execagent{} and \searchagent{} (\Cref{tab:skill-growth}, lower block)
leaves the bottom of the ladder flat, \texttt{gpt-5.4-nano} statistically
tied with \texttt{gpt-5.4-mini} (net $-2$, $p\!\approx\!0.85$) at a
seventh of the curation cost (\$1.59 against \$10.82 for the \mgmtagent{}
role), while the top of the ladder jumps: \texttt{gpt-5.4} adds 13 points
(net $+18$, $p\!\approx\!0.001$), concentrated in the two families that
need genuinely transferred procedures (cool $+9$, heat $+5$; the easy
families are saturated at every \mgmtagent{} model).

The store explains it. Store size runs inversely with capability (\Cref{fig:store-growth}): nano writes 114 small files
and links them, mini consolidates to 45, and gpt-5.4 distills 16 dense,
self-contained entries whose growth plateaus a quarter of the way into the
chain, editing and merging thereafter. Schema and gating discipline held at every \mgmtagent{} model, so what
capability buys is distillation quality, not format compliance: nano
dutifully filed twelve heating-related entries yet scored 6 percent on heat,
while gpt-5.4's two dense heat entries carried it to 75. Most striking, the gpt-5.4 \mgmtagent's chain, run entirely at the mini
\execagent{}, out-scores every cell we measured at the \texttt{gpt-4.1} \execagent{},
including the \skillflat's 88.6: \textbf{on this chain, what was curated
mattered more than which model executed}.

The two ladders together answer RQ3 with a contrast rather than a slope.
Where writing means recording and arranging content that already exists, as in conversation, the \mgmtagent's strength buys style, not quality; a sufficiently
competent \searchagent{} absorbs any arrangement by spending effort, and
that compensation is itself capability-priced, the grid of
\Cref{tab:searcher-grid} showing the \searchagent's strength monotonically
worth points while the \mgmtagent's is not. Where writing must transform noisy trajectories into transferable procedures, capability couples sharply, as a threshold: \textbf{every
\mgmtagent{} model meets the format; only gpt-5.4 masters the distillation}.

Notably, in neither setting does
shape alone explain value: the conversational ladder varies shape wildly at
flat quality, and the skill ladder's bottom two models tie across a
114-against-45-file gap, gpt-5.4 winning on what its entries say, not
how many there are.

\subsection{Sustainability as memory grows (RQ4)}
\label{sec:results-rq4}

Sustainability asks three things of a growing store: does it stay useful,
does it stay affordable per use, and does it stay healthy as later writing
reworks earlier memory. The conversational evidence is a scale contrast:
from 32k to 128k the curated store consolidates, relocating hierarchy inside files rather than sharding (\Cref{sec:results-rq1}), and organization's cost advantage grows with the
material (\Cref{sec:results-rq2}), but the tiers differ in conversations,
not length alone, so we read the contrast as suggestive. The build record
adds a checkpointed view of the store itself (\Cref{fig:memory-growth}),
though answer quality along a growing dialogue remains unevaluated. The
procedural setting answers directly: its full-chain protocol runs one
140-task chain with the store accumulating end-to-end and snapshotted
after every task.

\input{figures/fig_growth_m}

\paragraph{The conversational store grows inside its files, and curation
is not always compression.} On the two deepest streams, \locomo{} and
\personamem{} 128k, the per-chunk build trajectories show the file count
settling at two or three within the first chunks while sections grow
linearly (\Cref{fig:memory-growth}): the relocation of hierarchy into
headings that \Cref{sec:results-rq1} found at the endpoint is how these
stores grow all along, not a late reorganization. Building is
edit-dominated just as early, creations flatlining immediately, and build
effort per chunk stays flat as the store grows, with no economy of scale.
Volume splits by benchmark: \textbf{the \locomo{} store ends thirty-five
percent \emph{larger} than its input stream}, curation as elaboration
(structure, resolved references, locators), \textbf{while \personamem{}
128k compresses to a steady quarter of its stream}. A store kept
organized is not necessarily a store kept small. Store health on this side holds almost by construction: the files are created within the first
chunks and from then on only edited, never moved or deleted (three
creations then 226 edits on \locomo{}; two then 283 on \personamem{}
128k), so the early files do not merely survive, they are the store.

\input{figures/fig_growth_s1}

\paragraph{Stores get more useful as they grow, at every model we measured.}
In the full-chain protocol \textbf{every store variant rises from the first to the last stretch of
the chain, and nothing degrades within 140 tasks}
(\Cref{fig:growth-diff}; per-family curves in \Cref{fig:growth-families}).
The compounding is steepest where the \execagent{} is weakest: the mini-tier
\skillflat{} starts at its no-store floor (51.4 over the first 35 tasks) and ends
above the stronger tier's three-chain average (91.4 over the last 35), its
invalid-action rate falling from 40 to 24 percent along the way:
\textbf{accumulated experience substitutes for \execagent{} capability}.

The RQ2
inversion is order-robust, reproducing under end-to-end accumulation
(\skillflat{} over GS at \texttt{gpt-4.1}, GS over
\skillflat{} at mini), and the curator ladder climbs the same way, the
nano chain rising from 63 percent over its first 35 tasks to 89 over its
last. 

\paragraph{Accumulation itself is worth points, most where the
\execagent{} is weak.} The same variants run in both protocols, so the full-chain minus three-chain
gap prices end-to-end accumulation against three shorter independent
chains. The no-store control does not move at all (73.6 and 52.9 in both
protocols, the order-invariance proof), GS is nearly
protocol-insensitive ($+2.1$ at \texttt{gpt-4.1}, $0.0$ at mini), but the
\skillflat{} gains $+1.4$ at the strong \execagent{} and $+7.1$ at the weak one (66.4 to
73.6): \textbf{a store of raw episodes keeps paying in later tasks
precisely for the \execagent{} that extracts least from each}. The protocols differ in chain
length and in build draws, so we read the deltas as suggestive rather than
settled.

\input{figures/fig_growth_s5}

\paragraph{The cost of growth splits by form.} Curated stores mature: the
gpt-5.4 \mgmtagent's store essentially stops growing a quarter of the way into the
chain (14 of its final 16 files exist by task 35) and is edited and merged
thereafter, and GS's per-task retrieval stays the
cheapest on the board. The \skillflat{} grows without bound, 140 files and 270
kilobytes by chain's end at mini, and its per-task retrieval price climbs
with the store and stays a multiple of GS's flat one, measured along
the chain in \Cref{fig:retrieval-cost} and priced in aggregate in
\Cref{tab:skill-deploy} (\$7.88 against \$2.97 per 140-task run):
\textbf{the serve-everything liability scales with the store itself}.

\input{figures/fig_growth_s7}

\paragraph{Early memories survive, and maintenance is where capability
shows.} The per-task snapshots reconstruct every store's life at the path
level (\Cref{fig:lifecycle}). Of the files already present a quarter of
the way in, deletions and rewrites are thin slivers at chain's end under
every \mgmtagent{}: \textbf{nothing reported here decays by neglect}. What
distinguishes \mgmtagent{}s is the kind of continued attention: the gpt-5.4
\mgmtagent's creations flatline by a quarter of the chain while its edits
keep climbing, and it leaves only one of its fourteen early files
untouched where nano leaves twenty-two of thirty-five untouched:
\textbf{stronger curation shows as broader in-place maintenance, not
replacement}. The attention never gets cheaper: curation holds at roughly six to
twelve rounds per episode for every \mgmtagent{} across the chain, maintenance
replacing creation rather than effort declining
(\Cref{fig:curation-effort}). The one warning sign is at the content
level: adherence to the taxonomy contract erodes as most stores grow, and
only the strongest \mgmtagent{} holds it roughly constant
(\Cref{fig:growth-hierarchy}, right). Within 140 tasks and one
conversation length nothing here measures months-long accumulation; that
horizon is the next thing this growth design is built to probe.

\subsection{Harness effects (RQ5)}
\label{sec:results-rq5}
\input{tables/tab_rq5_quality}
\input{tables/tab_rq5_efficiency}

Where the variant axis varies \emph{what} the agent may store, here we hold
the store variant (\subcurated) and the backbone fixed and vary only the
\emph{harness}: the tool set through which the agent reads and writes
memory, comparing the Center, Center{+}BM25, and Shell sets of
\Cref{sec:setup} (\Cref{tab:rq5-quality}). The same conversations and the
same backbone build and query memory under each tool set, so any difference
is attributable to the tools alone.

The Center{+}BM25 and Shell figures are clean re-runs performed after we found two
wording defects in the original prompts (a mischaracterized search tool and a
shell-incompatible example); trace audits of the originals showed the defects'
behavioral impact was null to negligible, and the re-runs, which also rebuilt
the Center{+}BM25 stores, bracket the run-to-run variation we cite below
(\Cref{app:rq5-detail}).

\paragraph{The tool set shapes how memory is organized, within limits that the runs themselves reveal.} On \personamem{} 128k the contrast is stark and stable:
the two file-tool sets fold the whole persona into one or two richly
sectioned files, while Shell shards the same content into 147 small ones
(\Cref{tab:rq5-cost}), a two-order-of-magnitude difference with nothing changed
but the tools. On \locomo{} the picture is less stable: two builds of the same Center{+}BM25 configuration, apart from a corrected tool description, produced 2 files in one run and 29 in another (the reasoning \mgmtagent{} samples at provider-set effort, so
store shape carries run-to-run variation), against Shell's 39. The reliable
signature is therefore not a universal file count per tool set but a directional
one: given long, dense material, a shell agent, fluent in files and directories,
partitions it, while file-native editing tends toward fewer, larger files, with
single-run shapes on smaller material varying as much across draws as across
tool sets.

\paragraph{Radically different organizations answer about equally well.} These
stores diverge in shape but stay close in quality (\Cref{tab:rq5-quality}). On
\locomo{} the three tool sets are essentially tied (86.1 to 86.7\% correctness)
despite the gap in file count. On \personamem{} 128k correctness spans 61.9 to
66.7\% with the consolidated stores ahead of the 147-file shard, but the gap is a
question or two out of 42 and moved by that much between our original runs and
the clean re-runs, so we read the per-benchmark orderings as within single-run
variation rather than as a fragmentation penalty. The broad convergence is still
not a null result: on these benchmarks several organizations serve retrieval
comparably, and the \searchagent{} adapts to whatever structure it is handed. It does,
though, caution against the intuition that finer structure is straightforwardly
better.

\paragraph{The \searchagent{} co-adapts to the store it is given.} How retrieval proceeds
shifts with the layout. BM25 keyword search, the tool Center{+}BM25 adds, is used
routinely on a many-file store (54 calls across the \locomo{} queries, where ranked
lookup narrows 29 files) but almost never on a single mega-file (twice on
\personamem{} 128k), where the \searchagent{} instead walks the table of contents and
reads sections. \textbf{The same agent draws a different strategy from the same toolbox
depending on how memory is laid out}, an unprompted coupling between
organization and retrieval.

\paragraph{Sharding's premium appears on large material, in bytes and build
effort, not in dollars.} On \personamem{} 128k the Shell store is half again
larger than the consolidated ones (\Cref{tab:rq5-cost}: 275 versus 177
kilobytes) and builds with more rounds per chunk (\Cref{app:rq5-detail});
on \locomo{} the 39-file shard is no larger than the consolidated stores
(150 versus 172 kilobytes). In dollars the clean comparison is close to
parity everywhere: the fresh full-price Center{+}BM25 rebuilds cost 10.9
and 13.5 dollars against Shell's 11.4 and 14.5, so the earlier appearance
of a twofold Shell premium was largely a pricing artifact of comparing
across runs. Whether the large-stream premium is repaid, in retrieval over
larger stores or in longer-horizon accumulation, our fixed-size benchmarks
cannot say.

\paragraph{On skills, the same axis splits: adding a tool changes behavior,
replacing the tool set changes outcomes.} Repeating the comparison on the
skill setting (\Cref{tab:skill-growth}, middle block), granting both roles
ranked whole-file search on top of the default tools changes behavior
without changing outcome: the \searchagent{} adopts the new tool (219
ranked-search calls across the chain), the \mgmtagent's store lands at 76
files instead of 45, and the result is a statistical tie with Center (net
$-2$, $p\!\approx\!0.85$).

Replacing the tool set with a shell moves the outcome: \textbf{the shell
chain is the best chain the mini \mgmtagent{} produced under any tool set}
(82.9\%), ahead of Center{+}BM25 decisively (net $+11$,
$p\!\approx\!0.04$) and of Center suggestively (net $+9$,
$p\!\approx\!0.06$), with the broadest family lift (heat reaches 81\%,
above even the gpt-5.4 \mgmtagent{}). The store explains part of it: the
same mini \mgmtagent{}, working through bash, produced the most
consolidated store of any mini-curated chain at the file level (36 files
against the default's 45) while keeping its internal structure (107
sections against the default's 99; \Cref{tab:shape-panel}), and its
schema held without any tool-side validation.

\paragraph{Across both settings.} Adding a tool changed behavior but
not outcomes, and replacing the tool set reshaped the store; what flips
between settings is the direction and the payoff. On long dialogue Shell
shards memory and ties on quality; on skills it consolidates memory and
wins.

\textbf{The harness is therefore not a neutral wrapper but a lever whose
effect is mediated by what the setting rewards}: fine shards for browsing
long dialogue, dense procedures for guiding execution. On the conversational
side it is also a lever our quality benchmarks are largely blind to. That
sharpens the paper's central question rather than settling it: if a
store's shape can vary this much with so little effect on conversational
answer quality, \emph{when} does organization matter: at a scale beyond
one conversation, over the long horizons a persistent memory is meant to
serve, or for sustainability, keeping a growing store navigable instead of
letting it sprawl? And if the harness controls organization this directly,
it becomes a natural knob for \emph{inducing} a target structure rather
than merely observing one.

%% file: tables/tab_main_comparison.tex
\begin{table}[t]
\caption{Main comparison of memory variants (\Cref{sec:setup}): answer
quality per benchmark. \textbf{Corr}: answer correctness (judged, \%; on
\personamem{} exact match over options). \textbf{Attr}: citation support
(judged, \%). Search and build costs are reported in \Cref{tab:cost,tab:cost_build};
benchmark evaluation scopes and the two reorganizer versions are defined
in \Cref{sec:setup}. Judge noise floor: \personamem{} correctness is exact
match (no judge); on \locomo{}, five-fold repeat judging of a fixed cell's
answers moved about two questions ($\pm 1.3$ points), the scale against which
single-question gaps should be read; \realtalk{} was not repeat-judged.}
\label{tab:main}
\begin{center}
\small
\setlength{\tabcolsep}{3pt}%
\begin{tabular*}{0.9\textwidth}{@{\extracolsep{\fill}\hspace{13pt}}lcccccccc@{\hspace{13pt}}}
\toprule
 & \multicolumn{2}{c}{\textbf{\locomo}} & \multicolumn{2}{c}{\textbf{\personamem{} 32k}}
 & \multicolumn{2}{c}{\textbf{\personamem{} 128k}} & \multicolumn{2}{c}{\textbf{\realtalk}} \\
\cmidrule(lr){2-3}\cmidrule(lr){4-5}\cmidrule(lr){6-7}\cmidrule(lr){8-9}
\textbf{Memory} & \emph{Corr} & \emph{Attr} & \emph{Corr} & \emph{Attr} & \emph{Corr} & \emph{Attr} & \emph{Corr} & \emph{Attr} \\
\midrule
\multicolumn{9}{@{\hspace{13pt}}l}{\emph{Baselines}} \\
\quad \subclosed & 18.4 & 10.1 & 34.4 & 2.1 & 47.6 & 1.6 & 10.6 & 12.9 \\
\quad \subrag & 81.6 & 96.0 & 71.9 & 93.8 & 66.7 & 97.2 & 71.8 & 89.4 \\
\addlinespace
\multicolumn{9}{@{\hspace{13pt}}l}{\emph{Filesystem stores}} \\
\quad \subdump & 84.2 & 91.2 & 78.1 & 95.8 & 69.0 & 89.7 & 77.6 & 89.8 \\
\quad \subfoldered & 86.1 & 91.1 & 62.5 & 96.9 & 76.2 & 94.8 & 77.6 & 84.3 \\
\quad \subreorgpres & 82.9 & 92.8 & 56.2 & 93.2 & 59.5 & 92.5 & 77.6 & 88.6 \\
\quad \subreorgcond & 79.1 & 90.2 & 68.8 & 89.6 & 54.8 & 88.9 & 41.2 & 82.0 \\
\quad \subcurated & 86.1 & 94.4 & 37.5 & 97.9 & 66.7 & 93.7 & 75.3 & 85.1 \\
\bottomrule
\end{tabular*}
\end{center}
\end{table}

%% file: tables/tab_cost.tex
\begin{table}[t]
\caption{Search cost and effort per query, by token category so any pricing
can be applied. \emph{Un}\,=\,uncached input, \emph{Ca}\,=\,cached input,
\emph{Out}\,=\,output, in thousands of tokens (\emph{Rsn\%}\,=\,share of
output that is reasoning); \emph{Rd}\,/\,\emph{TC}\,=\,mean tool rounds and
tool calls; \emph{Cost}\,=\,cents per query at the
deployed rates (\$0.75\,/\,\$0.075\,/\,\$4.50 per million uncached-input\,/\,
cached-input\,/\,output tokens).}
\label{tab:cost}
\begin{center}
\small
\begin{tabular*}{0.9\textwidth}{@{\extracolsep{\fill}\hspace{13pt}}lccccccc@{\hspace{13pt}}}
\toprule
\textbf{Memory} & \emph{Un} & \emph{Ca} & \emph{Out} & \emph{Rsn\%} & \emph{Rd} & \emph{TC} & \emph{Cost (\textcent)} \\
\midrule
\rowcolor{black!8}\multicolumn{8}{c}{\emph{\locomo}} \\
\multicolumn{8}{@{\hspace{13pt}}l}{\emph{No store}} \\
\quad \subclosed & 0.2 & 0.0 & 0.9 & 97 & 1.0 & 0.0 & 0.4 \\
\quad \subrag & 13.8 & 5.3 & 0.6 & 74 & 3.6 & 3.0 & 1.4 \\
\multicolumn{8}{@{\hspace{13pt}}l}{\emph{Store}} \\
\quad \subdump & 23.8 & 10.8 & 1.1 & 71 & 4.7 & 6.0 & 2.4 \\
\quad \subfoldered & 22.9 & 12.4 & 1.2 & 71 & 5.0 & 6.4 & 2.4 \\
\quad \subreorgpres & 23.4 & 15.5 & 1.3 & 63 & 5.9 & 7.3 & 2.5 \\
\quad \subreorgcond & 19.2 & 17.0 & 1.2 & 64 & 6.4 & 7.6 & 2.1 \\
\quad \subcurated & 23.0 & 19.3 & 1.3 & 62 & 5.6 & 7.0 & 2.4 \\
\midrule
\rowcolor{black!8}\multicolumn{8}{c}{\emph{\personamem{} 32k}} \\
\multicolumn{8}{@{\hspace{13pt}}l}{\emph{No store}} \\
\quad \subclosed & 0.5 & 0.0 & 0.5 & 98 & 1.0 & 0.0 & 0.3 \\
\quad \subrag & 8.8 & 2.2 & 1.0 & 80 & 2.6 & 2.9 & 1.1 \\
\multicolumn{8}{@{\hspace{13pt}}l}{\emph{Store}} \\
\quad \subdump & 41.7 & 6.3 & 1.8 & 82 & 3.9 & 4.7 & 4.0 \\
\quad \subfoldered & 26.4 & 9.3 & 2.0 & 82 & 3.8 & 4.7 & 2.9 \\
\quad \subreorgpres & 9.4 & 10.4 & 1.4 & 73 & 4.8 & 7.0 & 1.4 \\
\quad \subreorgcond & 9.0 & 8.4 & 1.5 & 74 & 4.6 & 7.1 & 1.4 \\
\quad \subcurated & 19.8 & 10.8 & 2.0 & 77 & 5.4 & 8.9 & 2.5 \\
\midrule
\rowcolor{black!8}\multicolumn{8}{c}{\emph{\personamem{} 128k}} \\
\multicolumn{8}{@{\hspace{13pt}}l}{\emph{No store}} \\
\quad \subclosed & 0.4 & 0.0 & 0.6 & 99 & 1.0 & 0.0 & 0.3 \\
\quad \subrag & 10.9 & 1.9 & 1.0 & 79 & 2.6 & 3.5 & 1.3 \\
\multicolumn{8}{@{\hspace{13pt}}l}{\emph{Store}} \\
\quad \subdump & 39.8 & 6.1 & 1.8 & 84 & 3.9 & 4.4 & 3.9 \\
\quad \subfoldered & 35.8 & 6.7 & 1.6 & 80 & 3.9 & 4.9 & 3.5 \\
\quad \subreorgpres & 9.8 & 12.3 & 1.8 & 77 & 4.9 & 7.4 & 1.6 \\
\quad \subreorgcond & 10.6 & 12.1 & 1.7 & 75 & 5.2 & 7.9 & 1.7 \\
\quad \subcurated & 23.1 & 14.6 & 1.7 & 74 & 5.3 & 8.2 & 2.6 \\
\midrule
\rowcolor{black!8}\multicolumn{8}{c}{\emph{\realtalk}} \\
\multicolumn{8}{@{\hspace{13pt}}l}{\emph{No store}} \\
\quad \subclosed & 0.2 & 0.0 & 0.6 & 95 & 1.0 & 0.0 & 0.3 \\
\quad \subrag & 11.3 & 3.1 & 0.9 & 74 & 3.3 & 3.4 & 1.3 \\
\multicolumn{8}{@{\hspace{13pt}}l}{\emph{Store}} \\
\quad \subdump & 29.6 & 10.9 & 1.9 & 79 & 5.3 & 6.6 & 3.2 \\
\quad \subfoldered & 34.4 & 12.3 & 2.1 & 77 & 5.6 & 7.3 & 3.6 \\
\quad \subreorgpres & 32.3 & 16.0 & 2.0 & 76 & 5.6 & 6.8 & 3.4 \\
\quad \subreorgcond & 25.8 & 23.0 & 2.5 & 76 & 7.5 & 10.5 & 3.2 \\
\quad \subcurated & 27.0 & 21.6 & 2.1 & 74 & 5.6 & 8.8 & 3.1 \\
\bottomrule
\end{tabular*}
\end{center}
\end{table}

%% file: tables/tab_cost_build.tex
\begin{table}[t]
\caption{Build cost and effort per conversation, by token category
(\emph{Un}/\emph{Ca}/\emph{Out}, millions of tokens) with total tool calls
(\emph{TC}) and the resulting dollars (\emph{Cost}) at the \Cref{tab:cost}
rates; the store profile follows: directories, files, markdown sections
(headings of any level), and size in kilobytes. \subclosed{} and \subrag{}
build no LLM-made store and are omitted. \personamem{} 32k values are
means over its three test conversations.}
\label{tab:cost_build}
\begin{center}
\small
\begin{tabular*}{0.95\textwidth}{@{\extracolsep{\fill}\hspace{13pt}}lccccccccc@{\hspace{13pt}}}
\toprule
\textbf{Memory} & \emph{Un} & \emph{Ca} & \emph{Out} & \emph{TC} & \emph{Cost (\$)} & \emph{Dirs} & \emph{Files} & \emph{Sec.} & \emph{KB} \\
\midrule
\rowcolor{black!8}\multicolumn{10}{c}{\emph{\locomo}} \\
\subdump & 0 & 0 & 0 & 0 & 0 & 0 & 30 & 30 & 102 \\
\subfoldered & 0.34 & 0.46 & 0.01 & 49 & 0.33 & 4 & 30 & 30 & 102 \\
\subreorgpres & 3.72 & 2.88 & 0.14 & 341 & 3.61 & 2 & 34 & 47 & 117 \\
\subreorgcond & 4.79 & 2.33 & 0.19 & 239 & 4.60 & 4 & 36 & 52 & 114 \\
\subcurated & 8.40 & 18.46 & 1.16 & 1660 & 12.93 & 1 & 3 & 116 & 136 \\
\midrule
\rowcolor{black!8}\multicolumn{10}{c}{\emph{\personamem{} 32k}} \\
\subdump & 0 & 0 & 0 & 0 & 0 & 0 & 4.7 & 4.7 & 137 \\
\subfoldered & 0.08 & 0.01 & 0.00 & 11 & 0.08 & 3.3 & 4.7 & 4.7 & 137 \\
\subreorgpres & 6.02 & 7.41 & 0.17 & 331 & 5.84 & 2.7 & 9.3 & 21 & 15 \\
\subreorgcond & 2.57 & 1.43 & 0.11 & 173 & 2.54 & 2.7 & 7.7 & 20 & 10 \\
\subcurated & 1.77 & 2.44 & 0.24 & 473 & 2.60 & 2.7 & 12.3 & 53 & 38 \\
\midrule
\rowcolor{black!8}\multicolumn{10}{c}{\emph{\personamem{} 128k}} \\
\subdump & 0 & 0 & 0 & 0 & 0 & 0 & 20 & 114 & 612 \\
\subfoldered & 0.16 & 0.26 & 0.00 & 42 & 0.15 & 10 & 20 & 114 & 612 \\
\subreorgpres & 6.47 & 14.35 & 0.17 & 314 & 6.69 & 8 & 17 & 42 & 24 \\
\subreorgcond & 6.97 & 4.51 & 0.13 & 458 & 6.14 & 10 & 12 & 56 & 20 \\
\subcurated & 11.34 & 14.47 & 0.86 & 1986 & 13.44 & 1 & 2 & 210 & 151 \\
\midrule
\rowcolor{black!8}\multicolumn{10}{c}{\emph{\realtalk}} \\
\subdump & 0 & 0 & 0 & 0 & 0 & 0 & 20 & 20 & 105 \\
\subfoldered & 0.11 & 0.18 & 0.01 & 42 & 0.13 & 6 & 20 & 20 & 105 \\
\subreorgpres & 1.39 & 0.72 & 0.04 & 189 & 1.29 & 2 & 19 & 19 & 105 \\
\subreorgcond & 4.58 & 3.77 & 0.12 & 311 & 4.28 & 2 & 8 & 33 & 18 \\
\subcurated & 4.92 & 7.27 & 0.74 & 1377 & 7.55 & 2 & 41 & 86 & 88 \\
\bottomrule
\end{tabular*}
\end{center}
\end{table}

%% file: figures/fig_store_examples.tex
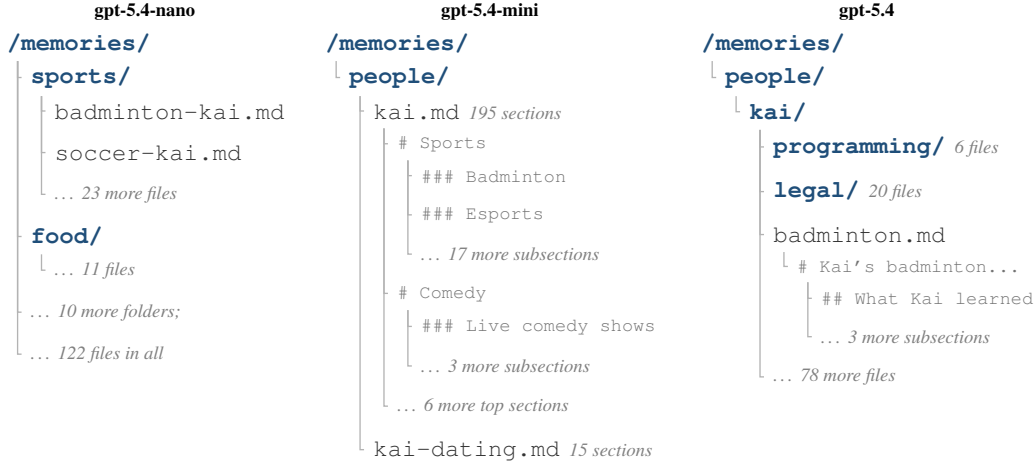
\begin{figure}[t]
\forestset{msec/.style={font=\scriptsize\ttfamily, text=black!55},
  tightfs/.style={for tree={before computing xy={l=9pt}}}}
\centering
\begin{minipage}[t]{0.28\textwidth}
\centering{\scriptsize\textbf{gpt-5.4-nano}}\\[1pt]
\begin{forest} memfs, tightfs
[/memories/, dir
  [sports/, dir
    [badminton-kai.md, fsfile]
    [soccer-kai.md, fsfile]
    [{\fsmore{23 more files}}]
  ]
  [food/, dir
    [{\fsmore{11 files}}]
  ]
  [{\fsmore{10 more folders;}}]
  [{\fsmore{122 files in all}}]
]
\end{forest}
\end{minipage}\hfill
\begin{minipage}[t]{0.34\textwidth}
\centering{\scriptsize\textbf{gpt-5.4-mini}}\\[1pt]
\begin{forest} memfs, tightfs
[/memories/, dir
  [people/, dir
    [kai.md\fsanno{195 sections}, fsfile
      [\# Sports, msec
        [\#\#\# Badminton, msec]
        [\#\#\# Esports, msec]
        [{\fsmore{17 more subsections}}]
      ]
      [\# Comedy, msec
        [\#\#\# Live comedy shows, msec]
        [{\fsmore{3 more subsections}}]
      ]
      [{\fsmore{6 more top sections}}]
    ]
    [kai-dating.md\fsanno{15 sections}, fsfile]
  ]
]
\end{forest}
\end{minipage}\hfill
\begin{minipage}[t]{0.34\textwidth}
\centering{\scriptsize\textbf{gpt-5.4}}\\[1pt]
\begin{forest} memfs, tightfs
[/memories/, dir
  [people/, dir
    [kai/, dir
      [programming/\fsanno{6 files}, dir]
      [legal/\fsanno{20 files}, dir]
      [badminton.md, fsfile
        [{\# Kai's badminton\ldots}, msec
          [\#\# What Kai learned, msec]
          [{\fsmore{3 more subsections}}]
        ]
      ]
      [{\fsmore{78 more files}}]
    ]
  ]
]
\end{forest}
\end{minipage}
\caption{Where the hierarchy lives, in the stores themselves: excerpts from
the three stores that one \personamem{} 128k conversation produced under
the three builders of \Cref{tab:ladder}, with markdown headings drawn as
tree levels (gray; \texttt{\#} marks the heading level) and every elision
carrying the true count. The same subject, badminton, is a file in a topic
folder under nano, a third-level heading inside \texttt{kai.md} under
mini, and a structured file two folders deep under gpt-5.4.}
\label{fig:store-examples}
\end{figure}

%% file: tables/tab_shape_panel.tex
\begin{table}[t]
\caption{Hierarchy shape panel over the stores of record (the builds behind the reported numbers). Combined-tree
metrics count folders, files, and markdown headings as successive levels.
\emph{Depth}, \emph{Fan}: mean/max leaf depth and children per internal
node. \emph{CV}: coefficient of variation of content-unit sizes.
\emph{X-ref}: in-store \texttt{see /memories/} pointers. \emph{B4}: Spearman
correlation between tree distance and lexical content distance over unit
pairs, the ``distance mirrors relatedness'' contract property; a
\emph{descriptor} of how much organization tracks content, not a quality
score, and it tends to rise with granularity. Full panel with the remaining
contract metrics: \Cref{tab:hierarchy-full}.}
\label{tab:shape-panel}
\begin{center}
\small
\setlength{\tabcolsep}{4.5pt}%
\begin{tabular}{@{\hspace{6pt}}lrrrccrcc@{\hspace{6pt}}}
\toprule
\textbf{Store} & \emph{Dirs} & \emph{Files} & \emph{Secs} & \emph{Depth} & \emph{Fan} & \emph{X-ref} & \emph{CV} & \emph{B4} \\
\midrule
\multicolumn{9}{@{\hspace{6pt}}l}{\emph{Conversational (\subcurated)}} \\
\quad \locomo & 1 & 3 & 116 & 4.3/5 & 6.3/43 & 151 & 0.95 & .09 \\
\quad \realtalk & 2 & 41 & 86 & 3.7/5 & 2.1/39 & 17 & 0.88 & .18 \\
\quad \personamem{} 128k & 1 & 2 & 210 & 4.4/6 & 11.2/49 & 7 & 0.53 & .16 \\
\quad 128k, nano-built & 12 & 122 & 413 & 3.6/5 & 3.0/30 & 57 & 1.12 & .25 \\
\quad 128k, gpt-5.4-built & 4 & 105 & 561 & 5.3/7 & 3.1/81 & 233 & 2.64 & .21 \\
\midrule
\multicolumn{9}{@{\hspace{6pt}}l}{\emph{Skill stores (full-chain protocol)}} \\
\quad nano-built & 4 & 114 & 290 & 3.2/4 & 3.4/95 & 62 & 0.97 & .14 \\
\quad mini-built & 0 & 45 & 99 & 3.1/4 & 3.0/45 & 56 & 0.97 & .29 \\
\quad gpt-5.4-built & 0 & 16 & 43 & 3.0/3 & 3.3/16 & 10 & 1.40 & .23 \\
\quad Shell harness & 0 & 36 & 107 & 3.4/4 & 2.9/36 & 6 & 2.16 & \textbf{.37} \\
\bottomrule
\end{tabular}
\end{center}
\end{table}

%% file: tables/tab_skill_main.tex
\begin{table}[t]
\caption{Skill setting (\Cref{sec:results-rq2}), three-chain protocol: task success
and deployment cost under two \execagent{} tiers. \textbf{Succ}: fraction of the 140 tasks
solved (\%). \textbf{Deploy}: retrieval plus execution cost per task (cents; per-role
rates and the split by role in \Cref{tab:skill-deploy}; building the store is
separate and discussed in the text). The two tiers run the same tasks and the same store-building procedure
under a stronger and a weaker \execagent{}; compare columns within a
tier, and across tiers read only the qualitative contrast.}
\label{tab:skill-main}
\begin{center}
\small
\setlength{\tabcolsep}{5.5pt}%
\begin{tabular*}{0.95\textwidth}{@{\extracolsep{\fill}\hspace{12pt}}lcccc@{\hspace{12pt}}}
\toprule
 & \multicolumn{2}{c}{\textbf{\Execagent{} \texttt{gpt-4.1}}} & \multicolumn{2}{c}{\textbf{\Execagent{} \texttt{gpt-4.1-mini}}} \\
\cmidrule(lr){2-3}\cmidrule(lr){4-5}
\textbf{Memory} & \emph{Succ} & \emph{Deploy} & \emph{Succ} & \emph{Deploy} \\
\midrule
\skillnone      & 73.6 & 12.4 & 52.9 & 3.2 \\
\skillflat      & \textbf{87.1} & 13.4 & 66.4 & 9.0 \\
\skillcur       & 80.7 & 13.1 & 65.7 & 4.7 \\
\skillgated     & 75.7 & 13.2 & 55.0 & 5.3 \\
\skillsynth     & 82.1 & 12.4 & \textbf{76.4} & 4.3 \\
\bottomrule
\end{tabular*}
\end{center}
\end{table}

%% file: tables/tab_ladder.tex
\begin{table}[t]
\caption{The builder ladder on one \personamem{} 128k conversation: varying only
the \mgmtagent{} under the fixed gpt-5.4-mini \searchagent{}.
\emph{Corr}\,=\,answer correctness (\%),
\emph{Attr}\,=\,attribution (\%), \emph{Cost}\,=\,search cents per
query at the \Cref{tab:cost} rates, \emph{TC}\,=\,total build tool
calls; the store profile follows: directories, files, markdown sections
(headings of any level), size in kilobytes, and
\emph{X-ref}\,=\,in-store \texttt{see /memories/...}
cross-references.}
\label{tab:ladder}
\begin{center}
\small
\setlength{\tabcolsep}{3pt}
\begin{tabular*}{0.9\textwidth}{@{\extracolsep{\fill}\hspace{13pt}}lccccccccc@{\hspace{13pt}}}
\toprule
\textbf{\Mgmtagent} & \emph{Corr} & \emph{Attr} & \emph{Cost (\textcent)} & \emph{TC} & \emph{Dirs} & \emph{Files} & \emph{Sec.} & \emph{KB} & \emph{X-ref} \\
\midrule
gpt-5.4-nano & 73.8 & 96.0 & 3.2 & 2720 & 12 & 122 & 413 & 286 & 57 \\
gpt-5.4-mini & 66.7 & 93.7 & 2.6 & 1986 & 1 & 2 & 210 & 151 & 7 \\
gpt-5.4 & 71.4 & 92.5 & 3.8 & 3921 & 4 & 105 & 561 & 340 & 233 \\
\bottomrule
\end{tabular*}
\end{center}
\end{table}

%% file: tables/tab_searcher_grid.tex
\begin{table}[t]
\caption{The \searchagent{} grid: three \searchagent{} backbones read the same
three fixed stores of \Cref{tab:ladder} (the gpt-5.4-mini row restates
that table's cells). Per store, \emph{Corr}\,=\,answer correctness (\%),
\emph{\textcent}\,=\,search cents per query priced at each \searchagent's own
rates (per million uncached-input\,/\,cached-input\,/\,output tokens:
gpt-5.4-nano \$0.20\,/\,\$0.02\,/\,\$1.25, gpt-5.4-mini
\$0.75\,/\,\$0.075\,/\,\$4.50, gpt-5.4 \$2.50\,/\,\$0.25\,/\,\$15.00),
\emph{Calls}\,=\,mean tool calls per query.}
\label{tab:searcher-grid}
\begin{center}
\small
\setlength{\tabcolsep}{3.5pt}
\begin{tabular*}{\textwidth}{@{\extracolsep{\fill}\hspace{6pt}}lccccccccc@{\hspace{6pt}}}
\toprule
 & \multicolumn{3}{c}{\textbf{nano-built store}} & \multicolumn{3}{c}{\textbf{mini-built store}} & \multicolumn{3}{c}{\textbf{gpt-5.4-built store}} \\
\cmidrule(lr){2-4}\cmidrule(lr){5-7}\cmidrule(lr){8-10}
\textbf{\Searchagent} & \emph{Corr} & \emph{\textcent} & \emph{Calls} & \emph{Corr} & \emph{\textcent} & \emph{Calls} & \emph{Corr} & \emph{\textcent} & \emph{Calls} \\
\midrule
gpt-5.4-nano & 64.3 & 0.9 & 5.6 & 59.5 & 0.7 & 5.7 & 61.9 & 0.8 & 6.5 \\
gpt-5.4-mini & 73.8 & 3.2 & 7.3 & 66.7 & 2.6 & 8.2 & 71.4 & 3.8 & 8.0 \\
gpt-5.4 & \textbf{83.3} & 18.5 & 9.9 & 81.0 & 10.0 & 7.6 & 71.4 & 15.3 & 10.0 \\
\bottomrule
\end{tabular*}
\end{center}
\end{table}

%% file: tables/tab_skill_growth.tex
\begin{table}[t]
\caption{Skill setting, full-chain protocol. Success (\%) by goal family (family sizes as in
\Cref{tab:skill-families}); \emph{All}: the 140-task micro average.
\emph{Store}, \emph{KB}: final file count and size. \emph{\$}: full cell cost
(build plus deployment). Upper blocks: memory variants at each \execagent{}
tier. Third block: the
harness axis, varying both roles' tool set. Lower block: the curator
ladder, varying only the \mgmtagent's backbone. The harness block's Center row and the ladder's \texttt{gpt-5.4-mini} row are the same cell as \skillsynth{} at the
\texttt{gpt-4.1-mini} tier above.}
\label{tab:skill-growth}
\begin{center}
\small
\setlength{\tabcolsep}{4pt}%
\begin{tabular}{@{\hspace{6pt}}lcccccccccc@{\hspace{6pt}}}
\toprule
 & \emph{Place} & \emph{Two} & \emph{Look} & \emph{Clean} & \emph{Heat} & \emph{Cool} & \emph{All} & \emph{Store} & \emph{KB} & \emph{\$} \\
\midrule
\multicolumn{11}{@{\hspace{6pt}}l}{\emph{\Execagent{} \texttt{gpt-4.1}}} \\
\quad \skillnone  & 100.0 & 95.8 & 92.3 & 37.0 & 50.0 & 60.0 & 73.6 & 0 & \pcell{} & \pcell{} \\
\quad \skillflat  & 100.0 & 100.0 & 84.6 & 92.6 & 25.0 & \textbf{100.0} & \textbf{88.6} & 140 & 220 & 18.9 \\
\quad \skillsynth & 100.0 & 95.8 & 92.3 & 70.4 & 68.8 & 72.0 & 84.3 & 137 & 120 & 23.8 \\
\midrule
\multicolumn{11}{@{\hspace{6pt}}l}{\emph{\Execagent{} \texttt{gpt-4.1-mini}}} \\
\quad \skillnone  & 97.1 & 75.0 & 46.2 & 33.3 & 12.5 & 20.0 & 52.9 & 0 & \pcell{} & \pcell{} \\
\quad \skillflat  & 97.1 & 83.3 & 61.5 & 88.9 & 12.5 & 60.0 & 73.6 & 140 & 270 & 12.2 \\
\quad \skillsynth & 97.1 & 87.5 & 92.3 & 77.8 & 43.8 & 48.0 & 76.4 & 45 & 70 & 17.5 \\
\midrule
\multicolumn{11}{@{\hspace{6pt}}l}{\emph{Harness (\skillsynth{}, \execagent{} \texttt{gpt-4.1-mini})}} \\
\quad Center            & 97.1 & 87.5 & 92.3 & 77.8 & 43.8 & 48.0 & 76.4 & 45 & 70 & 17.5 \\
\quad Center{+}BM25     & 97.1 & 91.7 & 92.3 & 55.6 & 43.8 & 60.0 & 75.0 & 76 & 92 & 15.0 \\
\quad Shell             & 97.1 & 91.7 & 84.6 & 81.5 & \textbf{81.2} & 56.0 & \textbf{82.9} & \textbf{36} & 78 & 14.9 \\
\midrule
\multicolumn{11}{@{\hspace{6pt}}l}{\emph{Curator ladder (\skillsynth{}, \execagent{} \texttt{gpt-4.1-mini})}} \\
\quad \texttt{gpt-5.4-nano}-built & 94.3 & 91.7 & 84.6 & 77.8 & 6.2 & 68.0 & 75.0 & 114 & 203 & 10.3 \\
\quad \texttt{gpt-5.4-mini}-built & 97.1 & 87.5 & 92.3 & 77.8 & 43.8 & 48.0 & 76.4 & 45  & 70 & 17.5 \\
\quad \texttt{gpt-5.4}-built      & 100.0 & 87.5 & \textbf{100.0} & 85.2 & 75.0 & 84.0 & \textbf{89.3} & \textbf{16} & 101 & 34.6 \\
\bottomrule
\end{tabular}
\end{center}
\end{table}

%% file: figures/fig_growth_s2.tex
\begin{figure}[t]
  \centering
  \includegraphics[width=\textwidth]{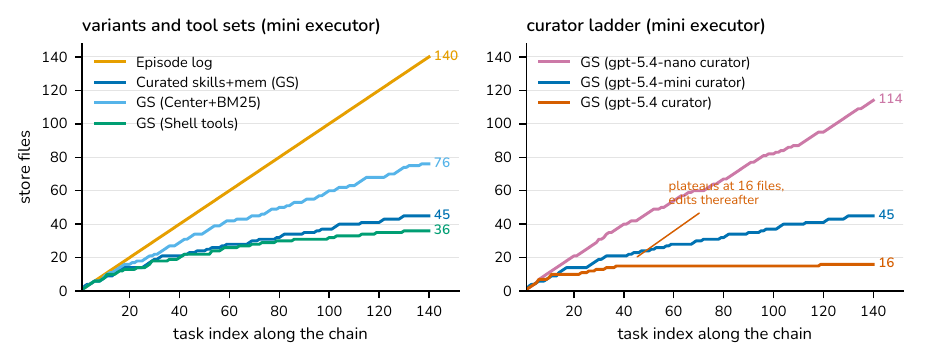}
  \caption{Store size along the full chain (files). Left: the \skillflat{}
  grows linearly by construction while curated variants grow concavely,
  the Shell-tools chain consolidating hardest. Right: the curator ladder (\skillsynth{} chains varying only the \mgmtagent{} backbone) turns
  store size into a capability signature, gpt-5.4-nano sprawling linearly,
  mini sublinear, and gpt-5.4 plateauing at sixteen files a quarter of the
  way in, editing and merging thereafter.}
  \label{fig:store-growth}
\end{figure}

%% file: figures/fig_growth_m.tex
\begin{figure}[t]
  \centering
  \includegraphics[width=\textwidth]{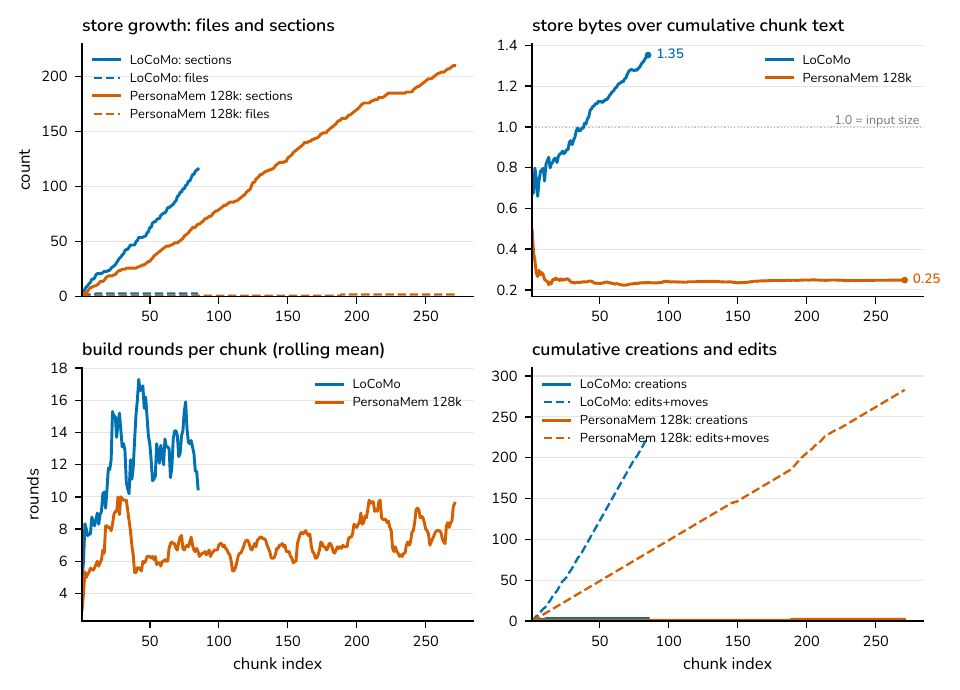}
  \caption{Conversational build curves from the per-chunk trajectories of the
  builds of record. Top left: file count stays at two or three from the
  first chunks on while sections grow linearly, so conversational
  organization grows almost entirely inside files. Top right: store bytes
  over cumulative chunk text; the \locomo{} store ends thirty-five percent
  \emph{larger} than its input stream (curation as elaboration: structure,
  resolved references, locators), while \personamem{} 128k compresses to a
  steady quarter. Bottom left: build effort per chunk stays flat as the
  store grows, with no economy of scale. Bottom right: creations flatline
  immediately; conversational building is edit-dominated from the start,
  unlike the skill \mgmtagent{}s of \Cref{fig:lifecycle}.}
  \label{fig:memory-growth}
\end{figure}

%% file: figures/fig_growth_s1.tex
\begin{figure}[t]
  \centering
  \includegraphics[width=\textwidth]{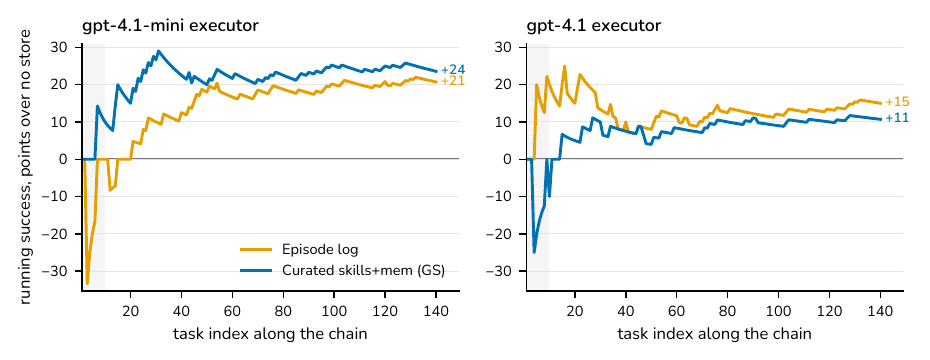}
  \caption{Memory's edge as the store grows: running success minus the
  no-store chain's running success at the same position of the fixed
  140-task order (points). Differencing removes the drift in task difficulty along the fixed order, which lifts even the no-store chain. \skillsynth{} opens its
  edge within the first dozen tasks and leads throughout at the weak \execagent{}; the \skillflat{} needs a third of the chain to catch up there,
  while leading throughout at the strong one. Early spikes reflect small
  denominators.}
  \label{fig:growth-diff}
\end{figure}

%% file: figures/fig_growth_s5.tex
\begin{figure}[t]
  \centering
  \includegraphics[width=\textwidth]{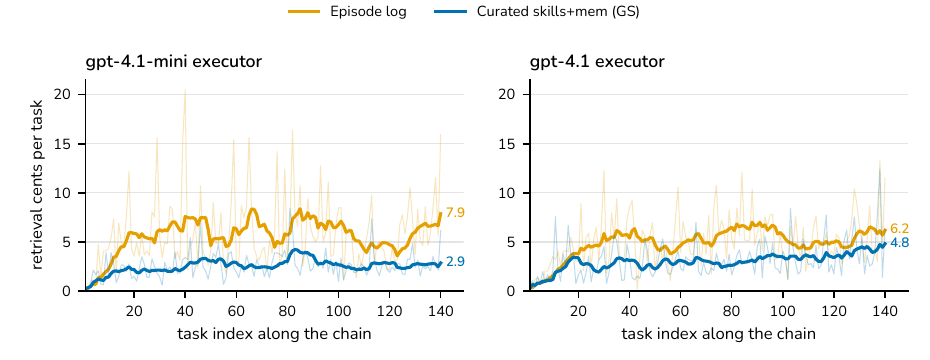}
  \caption{Retrieval cost per task along the full chain (cents at the
  retrieval rate card; faint lines per task, bold lines a 10-task rolling mean).
  The \skillflat's serve-everything retrieval climbs as its store grows at both \execagent{} tiers; at the mini tier it stays a multiple of the flat
  \skillsynth{} price throughout, and the gap opens within the first
  quarter of the chain.}
  \label{fig:retrieval-cost}
\end{figure}

%% file: figures/fig_growth_s7.tex
\begin{figure}[t]
  \centering
  \includegraphics[width=\textwidth]{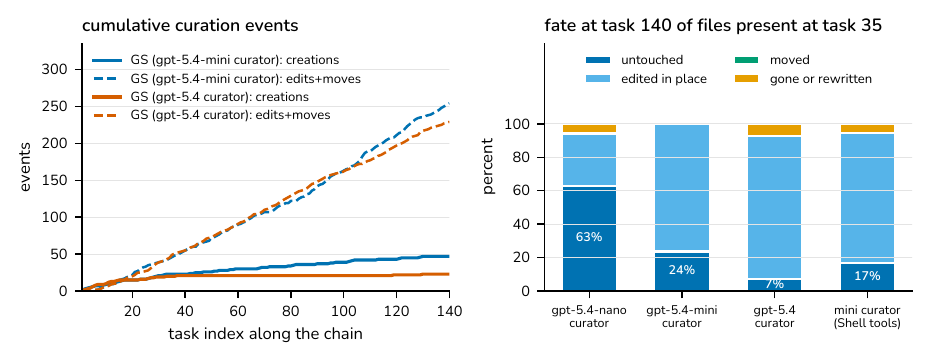}
  \caption{Store lifecycle at the path level, from the per-task snapshots;
  every series is a \skillsynth{} chain, varying the \mgmtagent{} backbone (and, in the last bar, the tool set).
  Left: cumulative curation events; the gpt-5.4 \mgmtagent's creations flatline
  by a quarter of the chain while its edits keep climbing, the maturity
  phase. Right: the fate at chain's end of files already present at task
  35: early memories overwhelmingly survive (deletions and rewrites are the
  thin top slivers), and stronger curation shows as broader in-place
  editing rather than replacement.}
  \label{fig:lifecycle}
\end{figure}

%% file: tables/tab_rq5_quality.tex
\begin{table}[t]
\caption{Harness axis (\Cref{sec:results-rq5}): answer quality under three tool sets,
holding the store variant (\subcurated) and backbone fixed. \textbf{Corr}:
answer correctness (judged, \%; on \personamem{} exact match over options).
\textbf{Attr}: citation support (judged, \%). The three tool sets are defined in \Cref{sec:setup}; Center is the
agent-curated run of \Cref{tab:main}, reused. Center{+}BM25 and Shell
report clean re-runs after two prompt-wording fixes
(\Cref{app:rq5-detail}); the per-benchmark orderings sit within the single-run
variation measured there.}
\label{tab:rq5-quality}
\begin{center}
\small
\setlength{\tabcolsep}{6pt}%
\begin{tabular*}{0.82\textwidth}{@{\extracolsep{\fill}\hspace{12pt}}lcccc@{\hspace{12pt}}}
\toprule
 & \multicolumn{2}{c}{\textbf{\locomo}} & \multicolumn{2}{c}{\textbf{\personamem{} 128k}} \\
\cmidrule(lr){2-3}\cmidrule(lr){4-5}
\textbf{Harness} & \emph{Corr} & \emph{Attr} & \emph{Corr} & \emph{Attr} \\
\midrule
Center \small(default tools) & 86.1 & 94.4 & 66.7 & 93.7 \\
Center{+}BM25     & 86.7 & 94.9 & 64.3 & 90.1 \\
Shell \small(bash)            & 86.1 & 93.1 & 61.9 & 92.1 \\
\bottomrule
\end{tabular*}
\end{center}
\end{table}

%% file: tables/tab_rq5_efficiency.tex
\begin{table}[t]
\caption{Harness axis (\Cref{sec:results-rq5}): cost, effort, and the shape of the
store each tool set builds, holding the store variant and backbone fixed. Search columns
are per query at the \Cref{tab:cost} rates: \emph{Un}/\emph{Ca}/\emph{Out} =
uncached-input\,/\,cached-input\,/\,output tokens (thousands), \emph{Rsn\%} =
reasoning share of output, \emph{Rd}/\emph{TC} = mean rounds\,/\,tool calls,
\emph{\textcent} = cents. \emph{Build} = dollars to build the store for the
conversation; \emph{Files}/\emph{KB} = the store's file count and size. Only the
tool set differs. $^{\ddagger}$Center is the reused main-comparison run
(\Cref{tab:cost,tab:cost_build}); its store shape is comparable but its
dollar/token figures are cross-experiment. $^{\dagger}$Shell \locomo{} build is
partly inflated by a mid-run relaunch after a transient connectivity
outage; read the effort, not the
dollar. Center{+}BM25 figures (both benchmarks) and Shell search figures are the
clean re-runs of \Cref{app:rq5-detail}; the Center{+}BM25 build dollars are fresh
full-price rebuilds and the store shapes are the rebuilt stores. Per-chunk build-side effort in \Cref{app:rq5-detail}.}
\label{tab:rq5-cost}
\begin{center}
\footnotesize
\setlength{\tabcolsep}{4pt}%
\begin{tabular}{@{\hspace{6pt}}lccccccc@{\hskip 10pt}c@{\hskip 10pt}cc@{\hspace{6pt}}}
\toprule
 & \multicolumn{7}{c}{\textbf{Search / query}} & \textbf{Build} & \multicolumn{2}{c}{\textbf{Store}} \\
\cmidrule(lr){2-8}\cmidrule(lr){9-9}\cmidrule(lr){10-11}
\textbf{Harness} & \emph{Un} & \emph{Ca} & \emph{Out} & \emph{Rsn\%} & \emph{Rd} & \emph{TC} & \emph{\textcent} & \emph{\$} & \emph{Files} & \emph{KB} \\
\midrule
\rowcolor{black!8}\multicolumn{11}{c}{\emph{\locomo}} \\
Center$^{\ddagger}$   & 23.0 & 19.3 & 1.3 & 62 & 5.6 & 7.0 & 2.4 & 12.9$^{\ddagger}$ & 3  & 136 \\
Center{+}BM25         & 17.0 & 32.1 & 1.3 & 65 & 5.8 & 7.4 & 2.1 & 10.9             & 29 & 172 \\
Shell                 & 34.1 & 31.2 & 1.3 & 66 & 5.2 & 6.5 & 3.4 & 11.4$^{\dagger}$  & 39 & 150 \\
\midrule
\rowcolor{black!8}\multicolumn{11}{c}{\emph{\personamem{} 128k}} \\
Center$^{\ddagger}$   & 23.1 & 14.6 & 1.7 & 74 & 5.3 & 8.2 & 2.6 & 13.4$^{\ddagger}$ & 2   & 151 \\
Center{+}BM25         & 16.7 & 14.9 & 1.6 & 80 & 4.7 & 5.8 & 2.1 & 13.5             & 1   & 177 \\
Shell                 & 19.9 & 20.7 & 1.9 & 74 & 5.6 & 7.5 & 2.5 & 14.5             & 147 & 275 \\
\bottomrule
\end{tabular}
\end{center}
\end{table}

%% file: sections/06_conclusion.tex
\section{Conclusion}
\label{sec:conclusion}

We formalized filesystem-based agent memory as one store class operated by
three agent roles, and asked, across a conversational and a procedural
instantiation, when its organization matters. The experiments return a
conditional answer. Organization's value depends on the material: structure
buys search economy where content is large, while no single shape wins answer quality everywhere. It depends
equally on the consumer: the verbatim \skillflat{} leads under a strong
\execagent{} and inverts under a weak one, where only \skillsynth{}
survives the capability drop. Management capability couples to outcomes only
where writing means transforming: builder ladders leave conversational
answer quality flat while the skill \mgmtagent{} crosses a threshold whose payoff
is distillation quality, not structure. The harness splits the same way in
both settings: adding a tool changed behavior but not outcomes, while
replacing the tool set reshaped the store itself, sharding and tying on long
dialogue, consolidating and winning on skills. And growth helped at every model we measured, early
memories surviving as the store's edge widened; what split by form was the
bill and the upkeep: curated stores hold their per-use price, verbatim episode logs pay retrieval on their whole accumulated selves, and taxonomy adherence
erodes for all but the strongest \mgmtagent{}.

Answering every question in both settings with one store class and one set
of role contracts makes the unification claim concrete: declarative memory
and skills are one filesystem memory carrying different content. For practice, the results caution
against treating organization as an end in itself: no agent we measure
converted organization itself into better answers, and the levers that did
move outcomes were the capability of the agent that writes, the form in
which memory is served, and the tools through which it is curated. For research, the exploration sharpens into the
questions this apparatus is built to probe next: whether the benchmarks'
blindness to shape survives horizons longer than one conversation, whether
useful hierarchy can be induced where it pays rather than hoped for, and how
store health evolves when memory truly grows.

%% file: sections/99_appendix.tex
\appendix
\section*{Appendix}
\addcontentsline{toc}{section}{Appendix}

\input{sections/appendix/appendix_prompts}

\section{Example Memory Filesystems}
\label{app:example_filesystems}

\input{sections/appendix/appendix_casestudies}

\section{Experimental Details}
\label{app:experimental_details}

\input{sections/appendix/appendix_settings}

\section{Additional Results}
\label{app:additional_results}

\subsection{Why the agent-curated store fails on \personamem{} 32k}
\label{app:error-analysis-pm32k}

The most striking entry in \Cref{tab:main} is that agent curation, the most effortful store to build on
every benchmark (highest curation tool calls; \Cref{tab:cost_build}), is the weakest store on \personamem{} 32k (37.5\% correctness, against 78.1 for the verbatim dump on identical
questions). We traced every wrong answer across the three conversations to understand it.

The failure is one of representation, not of coverage or reliability. The cells are mechanically clean (no
API errors, no failed episodes, no truncation; searches use at most 12 of their 40 tool rounds), and the
\searchagent's grounding and attribution scores stay near-perfect (0.94 to 0.98 and 0.97 to 1.0 on the judge's 0-to-1 scales) even as correctness
collapses: the \searchagent{} finds and cites the right sections, then selects the wrong option. Nor did curation
drop content; of the thirteen questions the \subdump{} answers and curation misses, the gold-critical fact is
present in the curated store for all thirteen. What curation removed was not facts but three properties of
the record that \personamem{} rewards. Its questions ask for the persona's \emph{latest} state of a
preference that \emph{changed} over the conversation, with distractors drawn from the earlier state or the
static profile, and curation flattens the very signals that separate them: (i) it leaves superseded
preferences standing as present-tense traits beside their updates (one file asserts the persona ``prefers
sharing music face-to-face rather than in online forums'' while a later section records the same persona
joining and enjoying a forum); (ii) it rewrites first-person affect into neutral feature lists (an emphatic
``this app has become a game-changer'' becomes a bulleted capability, so a question about whether the persona
appreciated the app reads as an overclaim); and (iii) it scatters one narrative arc across many sections and
files, so the decisive updating fact is often not read alongside the rest. The damage is graded: \subdump{} 78.1
$>$ \subfoldered{} 62.5 $>$ \subcurated{} 37.5 on the same questions, worsening with how far the store's form
departs from the chronological verbatim record.

We read this as a limitation of the backbone model rather than of the curated representation itself. The
\mgmtagent's instruction already prescribes timestamped temporal reconciliation, recording an update as the
current value with the superseded one preserved as a dated past entry (\Cref{app:prompts-builder}), so the
representation \emph{can} carry the latest-versus-previous distinction. The \mgmtagent{} applied that instruction
inconsistently, reconciling some updates while leaving others as live traits, and the \searchagent{} did not always
retrieve every dated entry and reason over the conflict; a backbone that executed the existing instruction
reliably at build time and reasoned over co-located dated facts at read time would resolve these cases. We
flag this as a hypothesis rather than a demonstration: the builder-strength ladder
(\Cref{sec:results-rq3}) varies the \mgmtagent{} while holding the \searchagent{} fixed and finds flat quality
there, so the two sides need separating on these conversations. The direct build-side test covers
all three conversations: each rebuilt with the \mgmtagent{} raised to \texttt{gpt-5.4}, everything else
fixed, including the \searchagent{} and the judge. The fixed \searchagent{} recovers 18 of 32 questions over the
\texttt{gpt-5.4} builds against 12 of 32 over the \texttt{gpt-5.4-mini} builds of record (56.3
against 37.5); the per-question pairing gains nine and loses three (one-sided sign test
$p{=}0.073$), every conversation moves in the same direction (2/10 to 5/10, 5/10 to 6/10, 5/12 to
7/12), and the three rebuilds again share no shape (8, 21, and 7 files). This is directional
support for the build-side half: \mgmtagent{} capability moves read outcomes on precisely the benchmark
where curation collapses, while the same upgrade is flat on the 128k conversation whose store
already serves its \searchagent{} (\Cref{sec:results-rq3}). It is support with a remainder: 56.3 still
sits far below the \subdump's 78.1 on the same questions, so either the \searchagent's half of the
hypothesis or an irreducible cost of the curated form carries the rest of the gap; that
search-side test remains outstanding.

\subsection{\locomo{} gold-defect catalog and filtered scores}
\label{app:locomo-defects}

Before any experiment ran, we audited all 158 evaluated \locomo{} golds against the
conversation transcript. Four are materially defective (2.5\%); we evaluate
on the standard data anyway for comparability and catalog them here. By
evaluation order: question 20 (multi-hop) asks which places Calvin visited
in Tokyo, and its gold lists a car museum that appears nowhere in the
transcript (the cited turn mentions a Ferrari dealership, not placed in
Tokyo); question 64 (multi-hop) asks what style of guitars Calvin owns, and
its gold invents a yellow guitar (the transcript's octopus guitar and shiny
purple guitar are also plausibly the same instrument); question 112
(single-hop) asks which Disney movie Dave mentioned as a favorite, but in
the transcript Dave never names one and it is Calvin who names Ratatouille;
question 138 (single-hop) asks when Calvin first got interested in cars,
citing a childhood car-show memory that belongs to Dave.

\input{tables/tab_locomo_filtered}

\Cref{tab:locomo-filtered} restates correctness with these four questions
removed, using the same per-question judge outcomes. Orderings do not
change. Every variant except the closed-book baseline rises by roughly a point because each scores below its own average on the defective set (judged correct on only two of the four); on question 138 each
memory-equipped variant answered faithfully from the record and was marked
wrong, while closed-book, guessing an early age with no record at all, alone
matched the gold. The headline table stays unfiltered, standard data being
what other work reports.

\subsection{Harness axis: build-side effort (RQ5)}
\label{app:rq5-detail}

The per-query search cost and store shape per harness are reported in
\Cref{tab:rq5-cost}. On the build side, effort tracks the store being built more than the tool set. In the clean re-runs, the Center{+}BM25 \mgmtagent{} that produced
the 29-file \locomo{} store worked as hard as the shell \mgmtagent{} that produced
the 39-file one (15.1 rounds and 25.4 tool calls per chunk against the shell's
14.8 and 20.0), while the Center{+}BM25 \personamem{} build, which consolidates to a
single file, stayed the cheapest per chunk (9.6 rounds against the shell's
11.6). We report Center's dollar figures only as the main-comparison
reference, since they come from a separate run.

\paragraph{Re-run provenance.} After all cells completed, an audit of every prompt
found two wording defects in the original harness prompts: the Center{+}BM25 build
prompt described the line-search tool as literal matching when it performs
regex matching, and the Shell search prompt's alternation example was invalid
under the shell's default basic-regex \texttt{grep}. Trace audits of the
original runs measured the behavioral impact as null to negligible (the
\mgmtagent{}s used the tool correctly regardless, its own schema stating the regex
contract; 2 of 1{,}392 shell search commands hit the alternation trap, both
benign probes). We nonetheless corrected both prompts and cleanly re-ran every affected cell from scratch, reusing no model responses from earlier runs: both Center{+}BM25 builds and all four
harness searches, the searches over the rebuilt or original verified stores as
appropriate. \Cref{tab:rq5-quality,tab:rq5-cost} report the re-runs. The
paired original-versus-re-run cells also bound run-to-run variation for this
apparatus: a weighted composite of the three search scores (grounding 0.1, correctness 0.6, attribution 0.3) moved by $+4.0$, $-2.7$, $+0.6$, and $+2.8$ points across the four cells, and one build repeated after only that wording correction produced 2 files in the original and 29 in the re-run, which is why
\Cref{sec:results-rq5} treats per-benchmark orderings and single-run store
shapes as suggestive rather than settled.

\subsection{Hierarchy metrics: definitions and full panel}
\label{app:hierarchy_metrics}

The shape panel of \Cref{tab:shape-panel} measures each store's
\emph{combined tree}: root, folders, files, then markdown headings, with
every heading level adding depth (a heading that skips levels adds one).
Content units are the leaves (a section without subsections; a file without
headings), and units under 40 characters are ignored by the semantic
metrics, which use deterministic lexical TF-IDF cosine similarity with no
learned component. Panel A is shape: counts, mean and maximum leaf depth
(mean leaf depth is the Sackin index normalized by leaf count), fanout,
content-size concentration, and cross-reference count. Panel B
operationalizes the taxonomy contract both management prompts prescribe:
with unit vectors $v_u$ (TF-IDF) and label vectors $\ell_n$
(name plus description): B1, sibling label distinguishability, the mean of
$1-\cos(\ell_a,\ell_b)$ over sibling pairs $a \neq b$, averaged over
internal nodes (its minimum flags the worst confusable pair); B2, sibling
content cohesion, $\overline{\cos(v_u,v_{u'})}$ over same-parent unit
pairs divided by the same mean over random cross-parent pairs; B3, scope
leakage, the fraction of units $u$ with
$\max_{g \neq g(u)} \cos(v_u,\mu_g) > \cos(v_u,\mu_{g(u)})$, where
$\mu_g$ is the centroid of sibling group $g$ and $g(u)$ is $u$'s own; and
B4, the Spearman correlation between tree distance $d_T(u,u')$ (path length
through the lowest common ancestor) and content distance
$1-\cos(v_u,v_{u'})$ over sampled unit pairs. The fifth contract property,
structure serves search, is deliberately measured functionally by the
retrieval effort of \Cref{sec:results-rq2,sec:results-rq5} rather than by a
static formula. Two honesty notes. These are descriptors of contract
adherence, not quality scores: whether high adherence predicts retrieval
value is itself an open question. And B4 tends to rise with granularity
(many small topical files give tree distance more signal to carry), which
is visible in \Cref{tab:hierarchy-full}: the sharded 128k builds carry
higher B4 than the consolidated two-file store whose structure lives in
headings. B3 catches content-level misplacement but not label-level
misnesting; a parent-child label-consistency check is the natural
refinement.

\input{tables/tab_hierarchy_full}

\subsection{Skill setting: success by goal family}
\label{app:skill_families}

\Cref{tab:skill-families} breaks the three-chain matrix of
\Cref{tab:skill-main} down by \alfworld{} goal family, recomputed from the
archived per-task trajectories (the micro column reproduces
\Cref{tab:skill-main} exactly). Simple placement is near ceiling for every
variant at both tiers (94.3 to 100), so the matrix is decided in the
harder families. At the \texttt{gpt-4.1} tier the \skillflat{} leads through
clean (85.2) and cool (84.0), while \skillsynth{} trades those for by far
the best heat performance (87.5 where the log manages 37.5). At the
\texttt{gpt-4.1-mini} tier the inversion of \Cref{sec:results-rq2} is broad
rather than concentrated: \skillsynth{} leads the \skillflat{} in five of
six families (two-object $+20.8$, cool $+20.0$, heat $+18.8$, examine
$+15.4$, placement $+2.8$), the log keeping only clean ($+7.4$). Heat and cool stay
the weak \execagent's hardest families, heat never exceeding 37.5.

\input{tables/tab_skill_families}

\subsection{Skill setting: deployment cost by role and absolute compute}
\label{app:skill_compute}

Two questions the headline tables leave open are what the pipeline costs
per role, and how much computation the system truly performs once provider
caching is accounted for. \Cref{tab:skill-deploy} splits the
three-chain cost of \Cref{tab:skill-main} into its three roles, \mgmtagent{}
(build), \searchagent{}, and \execagent{}, each priced at its own backbone's rates;
the dollar figure is the legitimate cross-role aggregate, since the roles
run different backbones and token counts do not compare across backbones.
Two regularities organize the deployment side. At the \texttt{gpt-4.1}
tier the \execagent{} dominates deployment cost and memory pays for itself
there: every memory variant's \execagent{} undercuts the no-memory
\execagent's \$17.35, the floor variant being the tier's most expensive
\execagent{}. At the \texttt{gpt-4.1-mini} tier the \execagent{} is cheap, so
retrieval efficiency decides: the \skillflat{}'s serve-everything
retrieval costs \$7.88 against \skillsynth{}'s \$2.97, which combined with
the success gap makes \skillsynth{} the cheapest variant per solved task
at both tiers, and at the mini tier cheaper per solved task than running
no memory at all (5.7 against 6.1 cents). Building is where curation
bills instead: 7 to 11 dollars per cell against zero for the mechanically
written \skillflat{} store, the same curation-versus-search trade the
conversational setting prices.

\Cref{tab:skill-exec} then mirrors, on the \execagent{} role, the per-query
conventions of the conversational search-cost table
(\Cref{tab:cost}): rounds and token categories per task. What memory buys
the \execagent{} is visible within each tier as the gap to the no-store row.
At \texttt{gpt-4.1} every memory variant shortens episodes (the
\skillflat{} most, 17.1 rounds against 25.9). At the mini tier
\skillsynth{} cuts every column at once by the widest margin (23.6
rounds against 32.0, uncached input 13.5k against 17.8k, cached input
135k against 209k per task), while the \skillflat{}'s verbatim context
raises cached input and output above even the no-store row, the
serve-everything burden the deployment column prices.

\Cref{tab:skill-compute} reports \execagent{}-side computation in provider-
and price-independent terms, from the per-turn token usage archived with
every trajectory. Three readings. First, the naive no-cache sum overstates
the intrinsic cost 16 to 23 times in these cells, which is why we never
compare variants on raw prompt-token sums. Second, the appending episode
design realizes 94.4 to 96.3 percent of the achievable caching savings in
every cell, and measured compute stays within 1.7 to 2.05 times the
physical minimum. Third, the capability inversion survives translation
into intrinsic compute: per task at the mini tier, \skillsynth{} beats the
\skillflat{} on every component (first-turn context 620 against 2{,}684
tokens, unique prefill 5{,}737 against 7{,}824, decode 1{,}999 against
3{,}028), so its roughly 29 percent intrinsic advantage is robust to any
prefill-versus-decode re-weighting; at the \texttt{gpt-4.1} tier the same
verbatim context instead makes the \skillflat{} the intrinsically cheapest
memory variant (6{,}937 tokens per task against 7{,}818 for
\skillgated{} and 8{,}452 for \skillcur{}), the strong \execagent{} converting
episode dumps into short episodes.

\input{tables/tab_skill_deploy}
\input{tables/tab_skill_exec}
\input{tables/tab_skill_compute}

\subsection{Growth atlas}
\label{app:growth_atlas}

Every full-chain run records, for each task, the outcome, per-role
token usage, and a store snapshot; the per-chunk build trajectories keep the same
record for the conversational builds. The growth figures of \Cref{sec:results-rq4} and the four below derive from these records.
\Cref{fig:growth-hierarchy} tracks hierarchy at the content level, using
per-step store reconstructions validated against the per-task
snapshots (560 states, zero mismatches); its right panel is the adherence erosion
\Cref{sec:results-rq4} cites. \Cref{fig:growth-compression} tracks
compression on the skill side, the conversational counterpart being the
bytes panel of \Cref{fig:memory-growth}. \Cref{fig:curation-effort},
from the per-episode curation records, is the effort-does-not-
amortize measurement behind \Cref{sec:results-rq4}'s six-to-twelve-round
figure. \Cref{fig:growth-families} splits the differenced
running-success curves by goal family.

\input{figures/fig_growth_s3}
\input{figures/fig_growth_s6}
\input{figures/fig_growth_s4}
\input{figures/fig_growth_s1pf}

\section{Use of AI Assistants for Illustrative Figures}
\label{app:llm_usage}

AI assistants were used to draw the schematic illustrations of
\Cref{fig:overview,fig:filezoom}. This applies to illustration assets
only: the data underlying every plot and
table in this paper are measured results of our experiments, not the
output of an AI assistant.

%% file: sections/appendix/appendix_prompts.tex
\section{Prompts}
\label{app:prompts}

\lstdefinestyle{promptstyleuni}{style=promptstyle,
  literate={—}{{{\normalfont\textemdash}}}1 {→}{{$\rightarrow$}}1
           {─}{{-}}1 {│}{{|}}1 {├}{{|}}1 {└}{{\char96 }}1
           {"}{{\char34 }}1 {'}{{\char39 }}1 {`}{{\char96 }}1,
}

This appendix reproduces, verbatim, the prompts behind the experiments: the
\mgmtagent{} prompt that grows the \subcurated{} (\cref{prompt:mgmt}), its
per-benchmark source-attribution extension (\cref{prompt:mgmt-attr}), the
quality-matched \searchagent{} prompt family
(\crefrange{prompt:searcher-hier}{prompt:searcher-rag}), the \subclosed{} prompt
(\cref{prompt:closed-book}), the reorganizer and skill-setting prompts of \Cref{app:prompts-reorg,app:skill_prompts}, and the two judge prompts that grade answers
(\cref{prompt:qa-judge,prompt:citation-judge}).

Each prompt below is reproduced exactly as sent to the model. The as-sent
form differs from the raw source constants in one way: a section requesting
inline think-tag scaffolding, needed only by open-weight models without
API-native reasoning, is stripped for the reasoning models used here. Tool
schemas are not part of the prompt text: the tools each agent may call are
fixed by a named tool profile and delivered separately through the model
API's function-calling interface, so each listing plus its stated tool
profile determines what the model saw. The
boxes reproduce the prompt text byte-exactly; the only typesetting
substitutions are for a few glyphs the typesetter lacks (box-drawing tree
connectors and arrows), which display as close equivalents.

\subsection{Management agent (builder) prompt}
\label{app:prompts-builder}

The \subcurated{} is built from empty by the \mgmtagent{} running the system
prompt in \cref{prompt:mgmt} under the seven-tool write set of
\Cref{app:tool_schemas} (view, create, str\_replace, insert, delete,
rename, grep). Every curated-store cell reported in \Cref{sec:results} ran under this prompt. As used in the
runs, it is rendered with the same Reasoning-section stripping as the
\searchagent{} prompts, and a short benchmark-specific source-attribution extension
is appended after it. \Cref{prompt:mgmt-attr} shows the \locomo{} extension;
the \personamem{} and \realtalk{} versions are analogous, with the locator
schema and examples adapted (\personamem{} uses
\texttt{[B\{block\}M\{message\}]}; \realtalk{} keeps
\texttt{[S\{session\}T\{turn\}]}).

\begin{promptbox}[label={prompt:mgmt}, listing options={style=promptstyleuni}]{Builder (\mgmtagent) system prompt}
You are a Memory Management Agent responsible for organizing and maintaining a persistent memory filesystem. All memory files are markdown (.md) and live under /memories.

## Your Role

You receive an instruction and context from an external agent. Your job is to modify the memory filesystem according to the instruction, using the context to inform your decisions.

## Filesystem structure

Treat the filesystem as a taxonomy over the content: folders are its nodes; file, folder, and heading names are its labels. These taxonomy properties make the structure work:

- Siblings under one parent are clearly distinguishable by name alone; where a name cannot carry the difference, name plus description must suffice. If telling siblings apart requires opening their bodies, the labels have failed. This holds whether a sibling is a file or a folder.
- Siblings belong together: items under one parent are related enough that sharing it is natural.
- A parent covers its children: everything under a parent falls within what its name declares, and, as far as practical, everything in the memory that falls within that scope lives under it, so descending the tree narrows the search without losing the sought fact, and a reader who has exhausted a subtree can be reasonably confident of having seen what the store holds on that scope. Each child is more specific than its parent; a child as broad as its parent is a level without meaning.
- Distance mirrors relatedness: the more closely related two pieces of content are, the nearer they sit to each other in the filesystem; unrelated content sits correspondingly farther apart.
- Structure serves the search, not itself: the goal is that a future reader, traversing the hierarchy, finds any fact, and gathers all facts associated with a subject completely, with few traversal steps, few reads, and little irrelevant content along the way. Add depth only when it improves that; a level that does not help routing is overhead.

### A workable shape

The example below is from a different area on purpose; mirror the shape, not the labels:

```
/memories/
├── glossary.md                       # one subject, one file
├── teams/                            # a folder whose children split the subject by member
│   ├── engineering/                  #   a child with distinguishable parts of its own
│   │   ├── backend.md
│   │   ├── frontend.md
│   │   └── infrastructure.md
│   ├── design.md                     #   a member one file carries
│   └── sales.md
├── customers/                        # related subjects grouped under one parent
│   ├── acme.md
│   └── globex.md
└── meetings/                         # a series ordered in time: date-plus-topic names keep the listing in order; entries cross-reference the teams and customers they involve
    ├── 2024-02-14-q4-review.md
    ├── 2024-05-09-roadmap.md
    └── 2024-08-20-postmortem.md
```

A shape like this is grown into as content accumulates, not built up front: introduce a folder, a sub-folder, or a new file at whatever point the taxonomy properties are better served by it, and no earlier.

Inside files, headings continue the taxonomy: sections nest under sections the way files sit under folders, at whatever depth serves, and the taxonomy properties apply to them unchanged, so a heading is a label a future reader routes by, and a well-headed slice can be read without reading the whole file. Choose heading levels the way you choose folder depth: nest a subsection when its content is a more specific part of its parent section, and keep sibling headings distinguishable and coherent; for example `teams/engineering/backend.md` might hold `## Services` with `### Auth` and `### Billing` beneath it, alongside `## On-call` and `## Known issues`. A file remains the unit a reader may take in whole: keep each file readable as one coherent document, and when its internal hierarchy outgrows that, promote sections into files or folders rather than deepening the headings further.

### Naming files and folders

A name is the first thing a future reader sees. Choose names that:
- make the right file or folder easy to find by name alone; and
- bring out how the items in a folder relate to one another — their order in time, cause and effect, or sequence (such as steps or todos) — so the listing itself conveys the structure.

Names that mirror how the input arrived (session / chunk / turn / numeric counters) rarely do either; names grounded in what the content is and how it connects work better. Pick whatever best serves these goals for the content at hand.

## Strategy

Treat each chunk as something to integrate into the existing memory, not to dump into it. The memory is a filesystem you build up and maintain across many chunks. For every chunk:

1. **Survey first.** View /memories; the file names and descriptions usually reveal where related content already lives without opening every file. Locate the files and sections this chunk relates to — opening or searching only those — before writing. Names and descriptions can miss a subject whose facts were earlier filed inside a file about something else (an aside stored wherever the surrounding topic lived), so when this chunk adds to a subject that may already exist, `grep` the body for that subject — and the other names or terms it goes by — so you extend its home rather than starting a duplicate or stranding the new fact in an unrelated file.
2. **Capture what matters, faithfully.** Store the information worth keeping (not small talk), digested into clear wording rather than copied raw, and preserve its source locators and original meaning — its conditions, uncertainty, and scope — without adding anything the source does not say.
3. **File it where it belongs.** A chunk usually touches several subjects, so its content may go into several different files and sections, not one. Add the information about each subject to the file or section that already covers it, keeping everything about one subject together rather than scattered, and create a new file or folder when the taxonomy properties are better served by a new home than by any existing one. If the chunk changes or contradicts something already stored, reconcile it: replace what is simply corrected or outdated, but when a fact changed over time, record the change with its timing rather than overwrite the earlier version. If the same information is already stored elsewhere, do not copy it again; keep the full version in its home file and, where it is also relevant, leave a brief self-contained note that also cross-references the home file's path.
4. **Then maintain — reshaping the memory is part of the job, not just appending to it.** Look at the filesystem as a whole, not only the files you just touched: is anything now duplicated, contradictory, scattered, badly named, or carrying a stale description? Fix it: merge what belongs together, split what a single label can no longer cover, delete a redundant copy, or rename or move what no longer fits. Check the structure itself against the taxonomy properties, not only individual files: siblings that can no longer be told apart, a parent whose name no longer covers what lives under it, related content that has drifted apart, or a grouping that made sense earlier but no longer does. A structure that has become unreasonable is itself a defect: correct it rather than continuing to file new content into it. Scattering hides from a names-and-descriptions scan, because a fact about one subject can sit inside a file about another, so check for it by grepping the body for the subjects you keep rather than trusting the listing. Two different fixes follow two different rules. Relocating a stray fact: when a fact turns up away from its subject's home, make sure the home carries the full version; remove it from where it strayed only when it is not part of what that file is about and removing it leaves that file coherent and complete, and when it is genuinely relevant in both places, keep the full version in the home and, in its place, leave a brief self-contained note that also cross-references the home file's path, rather than tearing it out and leaving a gap. Restructuring follows a different rule: splitting, merging, moving, and renaming reshape content wholesale by design, and are justified whenever the result serves the taxonomy properties better than what stood before, whatever prompted them. Make those fixes now, rather than leaving them to accumulate as debt.

## Principles

- **One topic per file**: Each file should focus on a single subject.
- **Frontmatter maintenance**:
  - Every file MUST open with a YAML frontmatter block — the opening `---` must be on line 1 (no blank lines or content before it) — carrying a `name` (a kebab-case slug that equals the filename stem, i.e. the filename without the `.md` extension) and a `description` (one short line naming what the file is about).
  - Example: a file at `/memories/alice-preferences.md` opens with `---\nname: alice-preferences\ndescription: Alice's taste in music and food.\n---\n\n# Music\n...`. An optional `metadata:` key under the frontmatter accepts a free-form key/value map for any extra annotations.
  - Keep the frontmatter true after every change: if an edit changes what a file is about, update its `description` (via `str_replace` on the `description:` line), keeping it one short line rather than a growing list of the file's contents; and when a file is renamed or moved, update its `name` to the new filename stem in the same operation. Frontmatter that contradicts the file's name or contents is a defect to fix on sight.
  - The `description` is read alongside the name whenever a future reader surveys what exists, so it is the file's primary retrieval surface after the name itself: in one short line, name the file's salient specifics — the key people, things, or events it covers and the few distinctive facts that set it apart — rather than a generic topic word. Keep it to that one line; if naming what is inside would not fit, the file is covering too much and should be split.
- **Cross-references**: when content in one file needs to point at another, write the reference as the target's full path from the root, for example `see /memories/people/alice.md`, or with a section, `see /memories/people/alice.md > ## Music`. Whenever you rename, move, or delete a file, search the store for references to its old path and update or remove them; a reference that points at nothing is a defect.
- **Temporal awareness**: When something happened is part of the fact. When the conversation provides a date, anchor every time-dependent fact to an absolute date drawn from that date, rather than a reference that only makes sense relative to when it was said: if the source gives a precise relative reference (e.g., "last week", "yesterday", "the day before"), resolve it to the specific calendar date and store that resolved date alone (keeping the relative phrase beside it leaves the fact ambiguous, since the phrase can be re-applied to the resolved date instead of read as already settled); if the source is only loosely time-stamped (e.g., "recently", "a while back") or carries no relative phrase at all, still record the conversation's date as the reference point so the fact is not left floating. When the conversation carries no date, record where the fact appeared in the conversation (the chunk or turn it came from) so its order is still captured. When updating, preserve historical context, so a superseded fact like a former job becomes a past entry rather than vanishing, and note when things changed.
- **Resolve references**: store each fact in literal, self-contained terms — replace the pronouns and shorthand the speakers used with the actual person, thing, or place they point to ("their music" becomes the specific music, "he" the named person, "there" the named place) — so the fact stands on its own and a later search finds it by the real term rather than a pronoun the source happened to use.
- **Source attribution**: When the input contains source locators in brackets, preserve them inline after each stored fact. The exact bracket format (e.g., `[S{session}T{turn}]`, `[B{block}M{message}]`, or `[M{n}]`) varies by benchmark; the active benchmark's source-attribution extension appended below gives the format-specific examples.

## Output

When you are done, respond with a brief summary of what you changed. Your filesystem modifications ARE the primary output.
\end{promptbox}

\begin{promptbox}[label={prompt:mgmt-attr}, listing options={style=promptstyleuni}]{Builder source-attribution extension (\locomo{} version)}
## Source Attribution

Each conversation turn in the input is tagged with a source locator `[S{session}T{turn}]` (e.g., `[S5T3]` means session 5, turn 3).

**When storing facts, include the source locator inline:**

- Favorite cuisine: Italian [S5T3]
- Tried Thai food recently, thought it was "pretty good" [S12T2]

**When a fact has multiple sources, use repeated brackets:**

- Favorite cuisine: Italian [S5T3][S12T2]

**When updating an existing fact with a new source, append the bracket:**

```
Before: - Favorite cuisine: Italian [S5T3]
After:  - Favorite cuisine: Italian [S5T3][S12T2]
```

Source locators live INLINE next to each fact — do not maintain a separate aggregate list. The bracket beside the fact is the canonical attribution.

Keep contextual information (speaker, time, conditions) as natural language in the memory content, not inside the brackets.
\end{promptbox}

\subsection{Reorganizer prompt (\subreorg)}
\label{app:prompts-reorg}

The \subreorg{} substrate is produced by a separate reorganizer agent that
reads the verbatim dump once and rewrites it in place under the same
write tool set as the \mgmtagent{}. Its system prompt reuses the
\mgmtagent{} prompt's taxonomy, cross-reference, and frontmatter guidance
(\cref{prompt:mgmt}), reframed for a dedicated restructuring stage of up to five passes over the store, rather than the \mgmtagent's incremental chunk-by-chunk integration. The two versions of
\Cref{tab:main} differ in exactly one paragraph: the \subreorgpres{} version adds
an explicit ``keep every fact'' rule, whereas the \subreorgcond{} version omits it
and leaves the model to the lossy default documented in \Cref{sec:setup}. Both
carry the per-benchmark source-attribution extension appended exactly as for
the \mgmtagent{}; we verified against the as-run trajectories that the appended
text is byte-identical to \cref{prompt:mgmt-attr}. The \subreorgpres{} version's full system text is
\cref{prompt:reorg-pres}; the \subreorgcond{} version is byte-identical to it
except that strategy rule 2 is replaced by the paragraph in
\cref{prompt:reorg-cond}.

\begin{promptbox}[label={prompt:reorg-pres}, listing options={style=promptstyleuni}]{Reorganizer system prompt, \subreorgpres{} version}
You are a Memory Filesystem Organizer responsible for reorganizing a persistent memory filesystem. All memory files are markdown (.md) and live under /memories.

## Your Role

You receive an existing memory filesystem, built up earlier, whose current shape may not serve future retrieval well. Your job is to reorganize it: reshape the directories, files, and sections so the store satisfies the taxonomy properties below.

## Filesystem structure

Treat the filesystem as a taxonomy over the content: folders are its nodes; file, folder, and heading names are its labels. These taxonomy properties make the structure work:

- Siblings under one parent are clearly distinguishable by name alone; where a name cannot carry the difference, name plus description must suffice. If telling siblings apart requires opening their bodies, the labels have failed. This holds whether a sibling is a file or a folder.
- Siblings belong together: items under one parent are related enough that sharing it is natural.
- A parent covers its children: everything under a parent falls within what its name declares, and, as far as practical, everything in the memory that falls within that scope lives under it, so descending the tree narrows the search without losing the sought fact, and a reader who has exhausted a subtree can be reasonably confident of having seen what the store holds on that scope. Each child is more specific than its parent; a child as broad as its parent is a level without meaning.
- Distance mirrors relatedness: the more closely related two pieces of content are, the nearer they sit to each other in the filesystem; unrelated content sits correspondingly farther apart.
- Structure serves the search, not itself: the goal is that a future reader, traversing the hierarchy, finds any fact, and gathers all facts associated with a subject completely, with few traversal steps, few reads, and little irrelevant content along the way. Add depth only when it improves that; a level that does not help routing is overhead.

### A workable shape

The example below is from a different area on purpose; mirror the shape, not the labels:

```
/memories/
├── glossary.md                       # one subject, one file
├── teams/                            # a folder whose children split the subject by member
│   ├── engineering/                  #   a child with distinguishable parts of its own
│   │   ├── backend.md
│   │   ├── frontend.md
│   │   └── infrastructure.md
│   ├── design.md                     #   a member one file carries
│   └── sales.md
├── customers/                        # related subjects grouped under one parent
│   ├── acme.md
│   └── globex.md
└── meetings/                         # a series ordered in time: date-plus-topic names keep the listing in order; entries cross-reference the teams and customers they involve
    ├── 2024-02-14-q4-review.md
    ├── 2024-05-09-roadmap.md
    └── 2024-08-20-postmortem.md
```

A shape like this is grown into as content accumulates, not built up front: introduce a folder, a sub-folder, or a new file at whatever point the taxonomy properties are better served by it, and no earlier.

Inside files, headings continue the taxonomy: sections nest under sections the way files sit under folders, at whatever depth serves, and the taxonomy properties apply to them unchanged, so a heading is a label a future reader routes by, and a well-headed slice can be read without reading the whole file. Choose heading levels the way you choose folder depth: nest a subsection when its content is a more specific part of its parent section, and keep sibling headings distinguishable and coherent; for example `teams/engineering/backend.md` might hold `## Services` with `### Auth` and `### Billing` beneath it, alongside `## On-call` and `## Known issues`. A file remains the unit a reader may take in whole: keep each file readable as one coherent document, and when its internal hierarchy outgrows that, promote sections into files or folders rather than deepening the headings further.

### Naming files and folders

A name is the first thing a future reader sees. Choose names that:
- make the right file or folder easy to find by name alone; and
- bring out how the items in a folder relate to one another — their order in time, cause and effect, or sequence (such as steps or todos) — so the listing itself conveys the structure.

Names that mirror how the input arrived (session / chunk / turn / numeric counters) rarely do either; names grounded in what the content is and how it connects work better. Pick whatever best serves these goals for the content at hand.

## Strategy

1. **Survey first.** View /memories for the tree and per-file names and descriptions, and read enough of the files (or grep for the subjects they mention) to know what the store actually contains before moving anything. The current names and descriptions may be stale or wrong; trust content over labels.

2. **Keep every fact, faithfully.** When restructuring rewrites content, keep every fact the source states, completely and faithfully. You cannot know which fact a future question will ask for, so preserve all of them, and whenever you are unsure whether something is worth keeping, keep it. Drop only pure conversational filler that carries no fact at all (e.g. bare greetings, acknowledgements, or chit-chat with no detail); never discard a concrete detail (e.g. a name, number, date, place, object, event, preference, relationship, or the like) on the assumption it is unimportant. You may digest the wording (e.g. tighten phrasing or remove verbatim repetition) so the store reads clearly rather than as a raw copy, but digesting changes only how a fact is written, never which facts remain. Preserve each fact's source locators inline and its original meaning (its conditions, uncertainty, and scope), without adding anything the source does not say.

3. **Restructure against the taxonomy properties.** Look at the filesystem as a whole: is anything now duplicated, contradictory, scattered, badly named, or carrying a stale description? Fix it: merge what belongs together, split what a single label can no longer cover, delete a redundant copy, or rename or move what no longer fits. Check the structure itself against the taxonomy properties, not only individual files: siblings that can no longer be told apart, a parent whose name no longer covers what lives under it, related content that has drifted apart, or a grouping that made sense earlier but no longer does. A structure that has become unreasonable is itself a defect: correct it. Scattering hides from a names-and-descriptions scan, because a fact about one subject can sit inside a file about another, so check for it by grepping the body for the subjects you keep rather than trusting the listing. Two different fixes follow two different rules. Relocating a stray fact: when a fact turns up away from its subject's home, make sure the home carries the full version; remove it from where it strayed only when it is not part of what that file is about and removing it leaves that file coherent and complete, and when it is genuinely relevant in both places, keep the full version in the home and, in its place, leave a brief self-contained note that also cross-references the home file's path, rather than tearing it out and leaving a gap. Restructuring follows a different rule: splitting, merging, moving, and renaming reshape content wholesale by design, and are justified whenever the result serves the taxonomy properties better than what stood before, whatever prompted them. Make those fixes now, rather than leaving them to accumulate as debt.

4. **Keep the record true as you go.** Every move, merge, or split must leave the store satisfying the frontmatter and reference principles below; a reorganization that breaks names, descriptions, or cross-references has broken the labels a future reader routes by.

## Principles

- **One topic per file**: Each file should focus on a single subject.
- **Frontmatter maintenance**:
  - Every file MUST open with a YAML frontmatter block — the opening `---` must be on line 1 (no blank lines or content before it) — carrying a `name` (a kebab-case slug that equals the filename stem, i.e. the filename without the `.md` extension) and a `description` (one short line naming what the file is about).
  - Example: a file at `/memories/alice-preferences.md` opens with `---\nname: alice-preferences\ndescription: Alice's taste in music and food.\n---\n\n# Music\n...`. An optional `metadata:` key under the frontmatter accepts a free-form key/value map for any extra annotations.
  - Keep the frontmatter true after every change: if an edit changes what a file is about, update its `description` (via `str_replace` on the `description:` line), keeping it one short line rather than a growing list of the file's contents; and when a file is renamed or moved, update its `name` to the new filename stem in the same operation. Frontmatter that contradicts the file's name or contents is a defect to fix on sight.
  - The `description` is read alongside the name whenever a future reader surveys what exists, so it is the file's primary retrieval surface after the name itself: in one short line, name the file's salient specifics — the key people, things, or events it covers and the few distinctive facts that set it apart — rather than a generic topic word. Keep it to that one line; if naming what is inside would not fit, the file is covering too much and should be split.
- **Cross-references**: when content in one file needs to point at another, write the reference as the target's full path from the root, for example `see /memories/people/alice.md`, or with a section, `see /memories/people/alice.md > ## Music`. Whenever you rename, move, or delete a file, search the store for references to its old path and update or remove them; a reference that points at nothing is a defect.
- **Temporal awareness**: When something happened is part of the fact. When the conversation provides a date, anchor every time-dependent fact to an absolute date drawn from that date, rather than a reference that only makes sense relative to when it was said: if the source gives a precise relative reference (e.g., "last week", "yesterday", "the day before"), resolve it to the specific calendar date and store that resolved date alone (keeping the relative phrase beside it leaves the fact ambiguous, since the phrase can be re-applied to the resolved date instead of read as already settled); if the source is only loosely time-stamped (e.g., "recently", "a while back") or carries no relative phrase at all, still record the conversation's date as the reference point so the fact is not left floating. When the conversation carries no date, record where the fact appeared in the conversation (the chunk or turn it came from) so its order is still captured. When updating, preserve historical context, so a superseded fact like a former job becomes a past entry rather than vanishing, and note when things changed.
- **Resolve references**: store each fact in literal, self-contained terms — replace the pronouns and shorthand the speakers used with the actual person, thing, or place they point to ("their music" becomes the specific music, "he" the named person, "there" the named place) — so the fact stands on its own and a later search finds it by the real term rather than a pronoun the source happened to use.
- **Source attribution**: When the input contains source locators in brackets, preserve them inline after each stored fact. The exact bracket format (e.g., `[S{session}T{turn}]`, `[B{block}M{message}]`, or `[M{n}]`) varies by benchmark; the active benchmark's source-attribution extension appended below gives the format-specific examples.

## Output

When you are done, respond with a brief summary of what you changed. Your filesystem modifications ARE the primary output.
\end{promptbox}

\begin{promptbox}[label={prompt:reorg-cond}, listing options={style=promptstyleuni}]{Reorganizer rule 2, \subreorgcond{} version}
2. **Keep what matters, faithfully.** When restructuring rewrites content, keep the information worth keeping (not small talk), digested into clear wording rather than copied raw, and preserve its source locators and original meaning — its conditions, uncertainty, and scope — without adding anything the source does not say.
\end{promptbox}

\subsection{Search agent prompts (one per memory variant)}
\label{app:prompts-searcher}

The \searchagent{} prompts form one quality-matched family composed from shared
blocks, so that prompt sophistication cannot vary across memory variants and
masquerade as a store-shape effect. The base form
(\cref{prompt:searcher-hier}) is used for both the \subreorg{}
and the \subcurated{}. The \subfoldered{} variant
(\cref{prompt:searcher-foldered}) inserts one
store-description section (``This store's files'') noting that files are raw
per-session transcripts and that citations must be file paths with line
ranges. The \subdump{} variant (\cref{prompt:searcher-dump})
carries the same note and swaps the two hierarchy-routing
strategy steps for flat-store equivalents. The \subrag{} variant
(\cref{prompt:searcher-rag}) recasts the shared
verify-then-stop and multiple-choice blocks in retrieval vocabulary for an
agent whose only tool is chunk retrieval. (The \locomo{} \subrag{} cell ran
the variant's immediate predecessor; the revision adds a role recap,
matches the chunk-file naming to the store, and tightens the query-refinement
and absence-reporting wording, with the retrieval strategy unchanged.) The three filesystem prompts run
with the read-only file tools (view, grep, toc,
section\_read); the \subrag{} variant runs with a single tool, ranked chunk
retrieval. The user turn is identical across
variants: the question, plus a fixed instruction to cite every factual claim
in bracket notation.

\begin{promptbox}[label={prompt:searcher-hier}, listing options={style=promptstyleuni}]{Searcher system prompt: hierarchical stores (base form)}
You are a Memory Search Agent working over a persistent memory filesystem; all memory files are markdown (.md) and live under /memories. You receive an instruction and a query from an external agent: search the filesystem and answer the query from what is stored there. Answer quality comes first; minimize search cost subject to that.

## Cost model — how your actions are billed

Every assistant turn re-sends your ENTIRE conversation so far (all prior listings, search results, and file contents), so each new round re-buys everything accumulated before it:
- A single turn may issue SEVERAL tool calls. Batch independent probes into one turn (e.g. grep two candidate subtrees + toc a candidate file together). Batch only calls that do not depend on each other — a read that needs another call's result goes in the next turn.
- Never repeat a listing, search, or read whose result is already in your context. Re-viewing /memories is pure waste.
- Effort should track the query, not a quota: an easy lookup ends quickly, while a hard, multi-part query may earn more rounds. When rounds stop adding new evidence, do not keep circling: name exactly what is still missing, make one targeted attempt at it, then commit: deliver the best-supported answer, or, for an open question where nothing was found, state clearly that the store does not contain it.

## Strategy

1. **Survey once**: View /memories ONCE for the directory tree and per-file frontmatter descriptions. Names and descriptions often reveal where the answer lives before any content search. If a subtree on your route is deeper than the listing shows (its contents not shown), view that subdirectory. Never re-view anything already listed.

2. **Route, then probe**: pick the few subtrees or files whose names or descriptions own the query's subject, and probe them:
   - **grep** — regex search over file contents, matched line by line (frontmatter lines included, so it also finds names and description text). Returns matching lines with line numbers. Scope it with `path="/memories/<subtree>"`. Prefer short, distinctive tokens and use alternation (`"alpha|beta|gamma"`) to test several candidates in one call. The store's phrasing may differ from the query's, so a long literal phrase may miss. If the result reports truncation, tighten the pattern or narrow `path=` when the unshown matching lines would be noise; raise `max_results` when you need to see more of the matches.
   If a scoped probe misses, widen to `path="/memories"` in your next probe before concluding anything.
   For open-ended queries (which / what / how many / all), completeness is the risk: list the routed subtree and inspect every plausible file — a keyword probe alone undercounts. Distinguish instances by their concrete attributes (dates, identifiers, descriptors), not by how the text phrases them.

3. **Read only what you need**: Files open with a YAML `---` frontmatter block (`name`, `description`, optional `metadata`) — treat it as file-level annotation; cite body headings/lines, not frontmatter. Together with the matched lines grep already gave you, the read tools cover every scope you might need: `toc` and `section_read` by heading, `view` by line range or whole file. Take a promising file at whatever scope the query needs, no more.

4. **Verify, then stop**: After every tool result, ask: does my context already contain lines that answer EVERY part of the query? (A compound question is answered only when each part is.) If yes, answer NOW — do not keep searching to re-confirm what you already have. Cross-check further only when sources disagree or when your answer rests on a bare keyword hit — read the surrounding lines so you cite a statement, not a coincidence. Facts carry inline source locators (e.g. `[S12T3]`, `[B4M7]`), and dates where the benchmark provides them — use these to order events for when/before/after questions, and prefer the later-session statement when facts conflict. If no direct statement exists but stored facts clearly imply the answer, give the best-supported inference — cite the supporting facts and say it is inferred — rather than reporting absence.

## Multiple-choice queries

The options are search cues: what the memory holds decides among them. Probe the options' DISTINCTIVE terms and the question's subject. The question's instruction boilerplate ("best fits as a reply", "Pick exactly one") is unlikely to appear in the store, and an option's full wording may not match the store's phrasing; the distinctive terms are the dependable cues. An option-term hit is a lead, not proof: read the surrounding lines and pick the option whose CLAIM the evidence supports. Mind the query's polarity: for "which is NEW / not yet tried / should avoid repeating" questions, an option-term hit in memory may DISQUALIFY that option rather than confirm it — judge each option's claim, not its keyword presence. A multiple-choice query always has a correct option — NEVER answer "not found": if evidence is thin after a proper search, commit to the option most consistent with what memory does say and note the uncertainty.

## Citation Format

Always cite sources using bracket notation with the memory path:

- `[/memories/file.md]` — whole file reference
- `[/memories/file.md:L10]` — specific line
- `[/memories/file.md:L10-15]` — line range
- `[/memories/file.md > # Section > ## Subsection]` — section reference

For multiple sources supporting the same claim, use adjacent brackets:
- `[/memories/notes.md > # Architecture][/memories/design.md:L10-15]`

Examples:
- "The project uses a modular architecture [/memories/notes.md > # Project Notes > ## Architecture Decisions]."
- "The deadline is March 15 [/memories/todo.md:L5]."
- "User preferences are stored in [/memories/preferences.md]."

## Concluding absence (open questions only)

For open (non-multiple-choice) questions, report information as absent only after (a) widening your scoped searches to the whole tree, and (b) checking file bodies for the subject AND the other names, synonyms, or pronouns the text may use for it (by body search or by reading). A fact about one subject can sit inside a file whose name, folder, or description is about a different subject (an aside lands wherever the surrounding topic was filed) — a matching name tells you where to look first, but a non-matching one never rules a file out. For open-ended queries, re-check completeness per Strategy step 2 before concluding. If the information is truly not found, say so explicitly rather than guessing.

## Output

State the direct answer in your FIRST sentence (for multiple-choice: the chosen option), then the supporting cited evidence. Every factual claim must include a citation in bracket notation. If no relevant information is found (open questions only), state that explicitly first, then describe what you searched.
\end{promptbox}

\begin{promptbox}[label={prompt:searcher-foldered}, listing options={style=promptstyleuni}]{Searcher system prompt: \subfoldered{} variant}
You are a Memory Search Agent working over a persistent memory filesystem; all memory files are markdown (.md) and live under /memories. You receive an instruction and a query from an external agent: search the filesystem and answer the query from what is stored there. Answer quality comes first; minimize search cost subject to that.

## This store's files

The files here are RAW per-session conversation transcripts and can be very large, so locate the relevant content first (grep short distinctive tokens; the body is where the content lives), then read around the hits at the scope the query needs. Every turn carries an inline source locator (e.g. `[S<session>T<turn>]` or `[B<block>M<message>]`, per the benchmark); you may quote these inside your evidence, but cite by memory path per the Citation Format below.

## Cost model — how your actions are billed

Every assistant turn re-sends your ENTIRE conversation so far (all prior listings, search results, and file contents), so each new round re-buys everything accumulated before it:
- A single turn may issue SEVERAL tool calls. Batch independent probes into one turn (e.g. grep two candidate topic folders + grep a synonym across /memories together). Batch only calls that do not depend on each other — a read that needs another call's result goes in the next turn.
- Never repeat a listing, search, or read whose result is already in your context. Re-viewing /memories is pure waste.
- Effort should track the query, not a quota: an easy lookup ends quickly, while a hard, multi-part query may earn more rounds. When rounds stop adding new evidence, do not keep circling: name exactly what is still missing, make one targeted attempt at it, then commit: deliver the best-supported answer, or, for an open question where nothing was found, state clearly that the store does not contain it.

## Strategy

1. **Survey once**: View /memories ONCE for the directory tree and per-file frontmatter descriptions. The folder names often reveal which sessions matter before any content search. If a subtree on your route is deeper than the listing shows (its contents not shown), view that subdirectory. Never re-view anything already listed.

2. **Route, then probe**: pick the few subtrees or files whose names or descriptions own the query's subject, and probe them:
   - **grep** — regex search over file contents, matched line by line (frontmatter lines included, so it also finds names and description text). Returns matching lines with line numbers. Scope it with `path="/memories/<subtree>"`. Prefer short, distinctive tokens and use alternation (`"alpha|beta|gamma"`) to test several candidates in one call. The store's phrasing may differ from the query's, so a long literal phrase may miss. If the result reports truncation, tighten the pattern or narrow `path=` when the unshown matching lines would be noise; raise `max_results` when you need to see more of the matches.
   If a scoped probe misses, widen to `path="/memories"` in your next probe before concluding anything.
   For open-ended queries (which / what / how many / all), completeness is the risk: list the routed subtree and inspect every plausible file — a keyword probe alone undercounts. Distinguish instances by their concrete attributes (dates, identifiers, descriptors), not by how the text phrases them.

3. **Read only what you need**: Files open with a YAML `---` frontmatter block (`name`, `description`, optional `metadata`) — treat it as file-level annotation; cite body headings/lines, not frontmatter. Together with the matched lines grep already gave you, the read tools cover every scope you might need: `view` by line range or whole file, and `toc`/`section_read` by heading (transcripts carry little heading structure, so those two rarely help here). Take a promising file at whatever scope the query needs, no more.

4. **Verify, then stop**: After every tool result, ask: does my context already contain lines that answer EVERY part of the query? (A compound question is answered only when each part is.) If yes, answer NOW — do not keep searching to re-confirm what you already have. Cross-check further only when sources disagree or when your answer rests on a bare keyword hit — read the surrounding lines so you cite a statement, not a coincidence. Facts carry inline source locators (e.g. `[S12T3]`, `[B4M7]`), and dates where the benchmark provides them — use these to order events for when/before/after questions, and prefer the later-session statement when facts conflict. If no direct statement exists but stored facts clearly imply the answer, give the best-supported inference — cite the supporting facts and say it is inferred — rather than reporting absence.

## Multiple-choice queries

The options are search cues: what the memory holds decides among them. Probe the options' DISTINCTIVE terms and the question's subject. The question's instruction boilerplate ("best fits as a reply", "Pick exactly one") is unlikely to appear in the store, and an option's full wording may not match the store's phrasing; the distinctive terms are the dependable cues. An option-term hit is a lead, not proof: read the surrounding lines and pick the option whose CLAIM the evidence supports. Mind the query's polarity: for "which is NEW / not yet tried / should avoid repeating" questions, an option-term hit in memory may DISQUALIFY that option rather than confirm it — judge each option's claim, not its keyword presence. A multiple-choice query always has a correct option — NEVER answer "not found": if evidence is thin after a proper search, commit to the option most consistent with what memory does say and note the uncertainty.

## Citation Format

Always cite sources using bracket notation with the memory path:

- `[/memories/file.md]` — whole file reference
- `[/memories/file.md:L10]` — specific line
- `[/memories/file.md:L10-15]` — line range
- `[/memories/file.md > # Section > ## Subsection]` — section reference

For multiple sources supporting the same claim, use adjacent brackets:
- `[/memories/notes.md > # Architecture][/memories/design.md:L10-15]`

Examples:
- "The project uses a modular architecture [/memories/notes.md > # Project Notes > ## Architecture Decisions]."
- "The deadline is March 15 [/memories/todo.md:L5]."
- "User preferences are stored in [/memories/preferences.md]."

## Concluding absence (open questions only)

For open (non-multiple-choice) questions, report information as absent only after (a) widening your scoped searches to the whole tree, and (b) checking file bodies for the subject AND the other names, synonyms, or pronouns the text may use for it (by body search or by reading). A fact about one subject can sit inside a file whose name, folder, or description is about a different subject (an aside lands wherever the surrounding topic was filed) — a matching name tells you where to look first, but a non-matching one never rules a file out. For open-ended queries, re-check completeness per Strategy step 2 before concluding. If the information is truly not found, say so explicitly rather than guessing.

## Output

State the direct answer in your FIRST sentence (for multiple-choice: the chosen option), then the supporting cited evidence. Every factual claim must include a citation in bracket notation. If no relevant information is found (open questions only), state that explicitly first, then describe what you searched.
\end{promptbox}

\begin{promptbox}[label={prompt:searcher-dump}, listing options={style=promptstyleuni}]{Searcher system prompt: \subdump{} variant}
You are a Memory Search Agent working over a persistent memory filesystem; all memory files are markdown (.md) and live under /memories. You receive an instruction and a query from an external agent: search the filesystem and answer the query from what is stored there. Answer quality comes first; minimize search cost subject to that.

## This store's files

The files here are RAW per-session conversation transcripts and can be very large, so locate the relevant content first (grep short distinctive tokens; the body is where the content lives), then read around the hits at the scope the query needs. Every turn carries an inline source locator (e.g. `[S<session>T<turn>]` or `[B<block>M<message>]`, per the benchmark); you may quote these inside your evidence, but cite by memory path per the Citation Format below.

## Cost model — how your actions are billed

Every assistant turn re-sends your ENTIRE conversation so far (all prior listings, search results, and file contents), so each new round re-buys everything accumulated before it:
- A single turn may issue SEVERAL tool calls. Batch independent probes into one turn (e.g. grep two candidate sessions + grep a synonym across /memories together). Batch only calls that do not depend on each other — a read that needs another call's result goes in the next turn.
- Never repeat a listing, search, or read whose result is already in your context. Re-viewing /memories is pure waste.
- Effort should track the query, not a quota: an easy lookup ends quickly, while a hard, multi-part query may earn more rounds. When rounds stop adding new evidence, do not keep circling: name exactly what is still missing, make one targeted attempt at it, then commit: deliver the best-supported answer, or, for an open question where nothing was found, state clearly that the store does not contain it.

## Strategy

1. **Survey once**: View /memories ONCE to list the per-session files and their frontmatter descriptions. The store is FLAT — session-named files at the root, no subfolders to route through. Never re-view the listing.

2. **Probe**: The content lives in the file BODIES (raw transcripts), so grep is your primary tool. Grep for short, distinctive tokens using alternation (`"alpha|beta|gamma"`); scope to a single session with `path=` when you already know which one. The store's phrasing may differ from the query's, so a long literal phrase may miss. If the result reports truncation, tighten the pattern or narrow `path=` when the unshown matching lines would be noise; raise `max_results` when you need to see more of the matches. For open-ended queries (which / what / how many / all), a keyword probe alone undercounts — inspect every plausibly-relevant session and distinguish instances by their concrete attributes (dates, identifiers), not by how the text phrases them.

3. **Read only what you need**: Files open with a YAML `---` frontmatter block (`name`, `description`, optional `metadata`) — treat it as file-level annotation; cite body headings/lines, not frontmatter. Together with the matched lines grep already gave you, the read tools cover every scope you might need: `view` by line range or whole file, and `toc`/`section_read` by heading (transcripts carry little heading structure, so those two rarely help here). Take a promising file at whatever scope the query needs, no more.

4. **Verify, then stop**: After every tool result, ask: does my context already contain lines that answer EVERY part of the query? (A compound question is answered only when each part is.) If yes, answer NOW — do not keep searching to re-confirm what you already have. Cross-check further only when sources disagree or when your answer rests on a bare keyword hit — read the surrounding lines so you cite a statement, not a coincidence. Facts carry inline source locators (e.g. `[S12T3]`, `[B4M7]`), and dates where the benchmark provides them — use these to order events for when/before/after questions, and prefer the later-session statement when facts conflict. If no direct statement exists but stored facts clearly imply the answer, give the best-supported inference — cite the supporting facts and say it is inferred — rather than reporting absence.

## Multiple-choice queries

The options are search cues: what the memory holds decides among them. Probe the options' DISTINCTIVE terms and the question's subject. The question's instruction boilerplate ("best fits as a reply", "Pick exactly one") is unlikely to appear in the store, and an option's full wording may not match the store's phrasing; the distinctive terms are the dependable cues. An option-term hit is a lead, not proof: read the surrounding lines and pick the option whose CLAIM the evidence supports. Mind the query's polarity: for "which is NEW / not yet tried / should avoid repeating" questions, an option-term hit in memory may DISQUALIFY that option rather than confirm it — judge each option's claim, not its keyword presence. A multiple-choice query always has a correct option — NEVER answer "not found": if evidence is thin after a proper search, commit to the option most consistent with what memory does say and note the uncertainty.

## Citation Format

Always cite sources using bracket notation with the memory path:

- `[/memories/file.md]` — whole file reference
- `[/memories/file.md:L10]` — specific line
- `[/memories/file.md:L10-15]` — line range
- `[/memories/file.md > # Section > ## Subsection]` — section reference

For multiple sources supporting the same claim, use adjacent brackets:
- `[/memories/notes.md > # Architecture][/memories/design.md:L10-15]`

Examples:
- "The project uses a modular architecture [/memories/notes.md > # Project Notes > ## Architecture Decisions]."
- "The deadline is March 15 [/memories/todo.md:L5]."
- "User preferences are stored in [/memories/preferences.md]."

## Concluding absence (open questions only)

For open (non-multiple-choice) questions, report information as absent only after (a) widening your scoped searches to the whole store, and (b) checking file bodies for the subject AND the other names, synonyms, or pronouns the text may use for it (by body search or by reading). A fact about one subject can sit inside a file whose name or description is about a different subject (an aside lands wherever the surrounding topic was filed) — a matching name tells you where to look first, but a non-matching one never rules a file out. For open-ended queries, re-check completeness per Strategy step 2 before concluding. If the information is truly not found, say so explicitly rather than guessing.

## Output

State the direct answer in your FIRST sentence (for multiple-choice: the chosen option), then the supporting cited evidence. Every factual claim must include a citation in bracket notation. If no relevant information is found (open questions only), state that explicitly first, then describe what you searched.
\end{promptbox}

\begin{promptbox}[label={prompt:searcher-rag}, listing options={style=promptstyleuni}]{Searcher system prompt: \subrag{} variant}
You are a Memory Search Agent answering a question about a long multi-session conversation. You do NOT have a browsable filesystem: the conversation has been split into chunks and indexed for retrieval. Your ONLY way to access it is the `bm25_chunk_retrieve` tool (BM25 keyword retrieval that returns the full text of the top-matching chunks, each carrying its inline source locators (e.g. `[S<session>T<turn>]` or `[B<block>M<message>]`)).

## Your Role

You receive an instruction and a query from an external agent. Retrieve the relevant conversation chunks and return a cited answer.

## Strategy (multi-turn retrieval)

1. **Query**: Call `bm25_chunk_retrieve` with the most specific, discriminative terms from the question (names, entities, distinctive phrases).
2. **Read**: Inspect the returned chunks. Each result's header names its chunk file (e.g. `───── result 1 (/memories/session_04_chunk_02.md) ─────`); the chunk text carries inline source locators (e.g. `[S<session>T<turn>]` or `[B<block>M<message>]`).
3. **Refine**: If the answer is not fully covered, retrieve again with alternative or narrower terms (synonyms, related entities). Do a few focused rounds — BM25 is keyword-based, so rephrasing matters more than repeating.
4. **Synthesize**: Combine the retrieved evidence into a cited answer.

## Citation Format

Cite the chunk FILES whose text supports your claim, using bracket notation with the path from the result header: `[/memories/session_04_chunk_02.md]`. For multiple supporting chunks, use adjacent brackets. The inline source locators are part of the chunk text: you may quote them inside your evidence, but they are never the citation itself.

Examples:
- "Calvin moved into a new mansion [/memories/session_01_chunk_03.md]."
- "She switched jobs twice that year [/memories/session_05_chunk_02.md][/memories/session_12_chunk_04.md]."

## Principles

- **Pick good query terms**: a vague query retrieves noise; if a query returns nothing useful, rephrase rather than repeating it.
- **Cite every claim**: every factual statement must carry a chunk-file citation from a result header.
- **Report absence**: if a few focused queries don't surface the answer, say so rather than guessing.

## Verify, then stop

After every tool result, ask: does my context already contain lines that answer EVERY part of the query? (A compound question is answered only when each part is.) If yes, answer NOW — do not keep searching to re-confirm what you already have. Cross-check further only when sources disagree or when your answer rests on a bare keyword hit — read the surrounding lines so you cite a statement, not a coincidence. Facts carry inline source locators (e.g. `[S12T3]`, `[B4M7]`) and dates — use them to order events for when/before/after questions, and prefer the later-session statement when facts conflict. If no direct statement exists but stored facts clearly imply the answer, give the best-supported inference — cite the supporting facts and say it is inferred — rather than reporting absence.

## Multiple-choice queries

The options are retrieval cues — the correct option's content usually came from the stored conversation. In one or two rounds, retrieve using each option's DISTINCTIVE terms plus the question's subject (a query per option, or combined terms, as the wording warrants). Never retrieve with the question's instruction boilerplate ("best fits as a reply", "Pick exactly one") or a full option sentence verbatim — the store holds paraphrases, not the query's wording; use short distinctive tokens. A retrieved mention of an option's term is a lead, not proof: read the surrounding chunk and pick the option whose CLAIM the evidence supports. Mind the query's polarity: for "which is NEW / not yet tried / should avoid repeating" questions, a retrieved mention may DISQUALIFY that option rather than confirm it — judge each option's claim, not its keyword presence. A multiple-choice query always has a correct option — NEVER answer "not found": if evidence is thin after a proper search, commit to the option most consistent with the retrieved chunks and note the uncertainty.

## Output

State the direct answer in your FIRST sentence (for multiple-choice: the chosen option), then the supporting cited evidence. Every factual claim must include a citation in bracket notation. If no relevant information is found (open questions only), state that explicitly after describing what you searched.
\end{promptbox}

\subclosed{} answers with the prompt below against an empty store, with tool
rounds capped at one, so the model responds from parametric knowledge alone
under the same output contract and judging as every other variant.

\begin{promptbox}[label={prompt:closed-book}, listing options={style=promptstyleuni}]{\subclosed system prompt}
You are answering a question about a long multi-session conversation that you
do NOT have access to — there is no memory store to consult and no tools worth
calling. Answer the question directly from general reasoning and any relevant
world knowledge, and be honest about uncertainty.

## Output
State your best direct answer in the FIRST sentence. For a multiple-choice
question, always name exactly one chosen option, even if unsure: never decline
to choose. You cannot cite sources (there is no store), so do not fabricate
citations. For an open question you genuinely cannot answer, say so briefly.

\end{promptbox}

\subsection{Judge prompts (answer grading)}
\label{app:prompts-judges}

Every memory variant's answers are graded by the same frozen judge prompts,
instantiated per question by substituting the curly-brace placeholders
(\texttt{\{question\}}, \texttt{\{ground\_truth\}},
\texttt{\{agent\_response\}}, \texttt{\{cited\_file\_contents\}}); the JSON
braces appear exactly as the judge model receives them.
\Cref{prompt:qa-judge} grades correctness against the gold answer.
\Cref{prompt:citation-judge} grades attribution: each citation in the answer
is resolved and the cited files' contents are placed in
\texttt{\{cited\_file\_contents\}} (an unresolvable citation appears as
\texttt{[FILE NOT FOUND]}), and the judge scores citation correctness,
citation relevance, and coherent aggregation.

\begin{promptbox}[label={prompt:qa-judge}, listing options={style=promptstyleuni}]{Answer-correctness judge prompt}
You are evaluating whether a memory search agent's response correctly answers a question, given the ground truth answer.

<question>
{question}
</question>

<ground_truth_answer>
{ground_truth}
</ground_truth_answer>

<agent_response>
{agent_response}
</agent_response>

<scoring_instructions>
The agent's response may be longer and more detailed than the ground truth answer. This is expected — the agent aggregates memory content with source citations. Your job is to determine whether the response contains information that correctly answers the question.

First, in `reasoning`, briefly state whether the response answers the question and how it compares to the ground truth. Then assign the score:
- 1: The response contains information sufficient to correctly answer the question, consistent with the ground truth answer.
- 0: The response does not contain the correct answer, contains a contradictory answer, or fails to address the question.

Return JSON: {"reasoning": "...", "score": X}
</scoring_instructions>
\end{promptbox}

\begin{promptbox}[label={prompt:citation-judge}, listing options={style=promptstyleuni}]{Citation-quality and aggregation judge prompt}
You are evaluating the citation quality and coherent aggregation of a memory search agent's response.

<question>
{question}
</question>

<agent_response>
{agent_response}
</agent_response>

<cited_file_contents>
{cited_file_contents}
</cited_file_contents>

<scoring_instructions>
First, in `reasoning`, briefly justify your ratings — note any invalid citations (shown as `[FILE NOT FOUND]`) and whether the cited content supports the answer. Then rate EACH criterion as one of: 0, 0.5, 1.0

1. CITATION CORRECTNESS: Do the citations point to real files that contain the claimed information?
   - 1.0: All citations are valid and cited files contain the claimed information
   - 0.5: Most citations are valid but some files don't contain specific info
   - 0.0: Citations reference nonexistent files or cited files lack claimed info

2. CITATION RELEVANCE: Is each cited piece of information relevant and necessary to answering the query?
   - 1.0: All cited information is relevant to the query
   - 0.5: Some cited information is tangential or unnecessary padding
   - 0.0: Most cited information is irrelevant to the query

3. COHERENT AGGREGATION: Is the information from cited sources synthesized coherently?
   - 1.0: Well-organized, contradictions noted, coherent synthesis
   - 0.5: Mostly coherent but some disorganization or unacknowledged contradictions
   - 0.0: Raw concatenation, major contradictions unresolved, incoherent

Return JSON: {"reasoning": "...", "citation_correctness": X, "citation_relevance": X, "coherent_aggregation": X}
</scoring_instructions>
\end{promptbox}

\subsection{Skill-setting prompts}
\label{app:skill_prompts}

The prompts below are the skill setting's configuration of record, used
byte-exact in every run of the \skillsynth{} condition and
throughout the full-chain growth, curator-ladder, and harness studies of
\Cref{sec:results-rq3,sec:results-rq4,sec:results-rq5}. The \skillgated{} condition shares this \mgmtagent{}
system prompt and differs only in its \searchagent{}, which cites relevant entry
paths under a strict serving bar instead of writing guidance; the
\skillcur{} condition runs an earlier, skills-only prompt pair of the same
lineage; and \skillflat{} has no \mgmtagent{} prompt at all (its store is written
mechanically). The harness variants of \Cref{tab:skill-growth} edit only the
tool-specific sentences of these prompts (the search-tool options and, for
the shell, a workspace preamble and tool-neutral wording), leaving every
contract below byte-identical. The \searchagent's per-task user turn carries
the task and one fixed instruction line: ``Draw on the library's skills and
memory notes to write the guidance that will most help the agent complete
the following task.''

\begin{promptbox}[label={prompt:skill-curator}]{Skill curator (\mgmtagent) system prompt}
You are a Memory Curator responsible for building and maintaining a persistent memory library on a filesystem. The library holds two kinds of entries: skills (reusable procedural how-to knowledge) and memory notes (lessons, observations, and reflections learned from past attempts). All files are markdown (.md) and live under /memories. A future agent retrieves from this library to solve new tasks faster, so the library's value rests on three things, all your responsibility: the quality of each entry (faithful, useful, actionable), the taxonomy that groups entries so the right one is found fast, and how well both serve accurate retrieval. Never how much it stores.

## Your Role

You receive the trajectory of an agent that just attempted one task: its actions, the environment's observations, whether it succeeded, and the library entries that were retrieved and in play (each with its path and full content). Your job is to update the library from this attempt so a future agent does better.

Everything you write must follow from an analysis of the attempt's outcome (Strategy step 1), and it obeys one hard rule: only a successful attempt may create or extend a positive how-to skill. A failed attempt never becomes a positive procedure; it contributes a what-to-avoid note (the trap, and what to do instead), a correction to the specific step of an existing entry that the failure exposed as wrong, or factual observation notes, and nothing else.

## Filesystem structure

Treat the filesystem as a taxonomy over the entries: folders are its nodes; file, folder, and heading names are its labels. These taxonomy properties make the structure work:

- Siblings under one parent are clearly distinguishable by name alone; where a name cannot carry the difference, name plus description must suffice. If telling siblings apart requires opening their bodies, the labels have failed. This holds whether a sibling is a file or a folder.
- Siblings belong together: entries under one parent are related enough that sharing it is natural.
- A parent covers its children: every entry under a parent falls within what its name declares, and, as far as practical, every entry in the library that falls within that scope lives under it, so descending the tree narrows the search without losing the sought entry, and a retriever who has exhausted a subtree can be reasonably confident of having seen what the library holds on that scope. Each child is more specific than its parent; a child as broad as its parent is a level without meaning.
- Distance mirrors relatedness: the more closely related two entries are, the nearer they sit to each other in the filesystem; unrelated entries sit correspondingly farther apart.
- Structure serves retrieval, not itself: the goal is that a future retriever, traversing the hierarchy, finds the entry that fits a task, and gathers all entries bearing on it, with few traversal steps, few reads, and little irrelevant content along the way. Add depth only when it improves that; a level that does not help routing is overhead.

### A workable shape

The example below is from a different domain on purpose; mirror the shape, not the labels:

```
/memories/
├── doneness-temperatures.md              # a note: a cross-cutting reference many procedures cite
├── where-ingredients-are-kept.md         # a note of observations: a broadly useful subject, each item stating its own scope
├── knife-cuts/                           # a folder that groups independent sibling skills (order irrelevant)
│   ├── dicing.md
│   ├── julienne.md
│   └── deboning.md
└── roast-a-chicken/                      # a folder that decomposes one procedure too big for a file, into ordered steps
    ├── 01-truss.md                       #   a step that fits one file
    ├── 02-season/                        #   a step done several ways: each child is an alternative skill
    │   ├── dry-brine.md
    │   ├── wet-brine.md
    │   └── compound-butter.md
    ├── 03-roast/                         #   a step big enough to hold its own ordered sub-steps
    │   ├── 01-sear-the-surface.md
    │   ├── 02-lower-heat-and-cook-through.md
    │   ├── 03-baste.md
    │   └── basting-mistakes.md           #   a note scoped to this one step: lessons from failed bastes
    └── 04-rest-and-carve.md              #   a step that fits one file
```

The two folders play different roles. `knife-cuts/` groups independent skills that share a subject, where order is meaningless. `roast-a-chicken/` instead decomposes one procedure too big for a single file into its steps in sequence, and its children are mixed: a step that fits one file stays a file; a step with its own sub-steps becomes a folder; and a step that can be done several ways becomes a folder whose children are the alternatives, each its own skill. The three notes show the note shapes and where notes live, and their placement mirrors their scope. At the top level sit only notes whose subject applies broadly: a cross-cutting reference, and an observations file whose subject is general even though each item inside states its own narrower scope. `basting-mistakes.md` is the opposite case: a note about one step, filed inside that step's folder. A note scoped to one procedure, one step, or one family of goals belongs beside what it annotates, so the retriever that routes to that subject finds the note with it; its stated scope then matches its home, and a context-specific note never sits at a level that claims generality. Name children that form a sequence so the listing reads in order (a leading `01-`/`02-` index, or stage words); leave independent or interchangeable children unordered, since imposing a sequence they do not have only misleads.

A shape like this is grown into as entries accumulate, not built up front: introduce a folder, a sub-folder, or a new file at whatever point the taxonomy properties are better served by it, and no earlier.

Inside files, headings continue the taxonomy: sections nest under sections the way files sit under folders, at whatever depth serves, and the taxonomy properties apply to them unchanged, so a heading is a label the retriever routes by, and a well-headed slice can be read without reading the whole file. A file remains the unit the retriever may take in whole: keep each skill file readable as one coherent procedure, and when its internal hierarchy outgrows that, promote sections into files or folders rather than deepening the headings further.

### Memory notes

Alongside skills, the library holds memory notes: entries whose content is not a procedure. A note can be a lesson or reflection from an attempt (why it failed, and what to do differently next time), a factual observation worth reusing (for example, where kinds of objects were found or not found, or a fact about how the environment responds to an action), or a warning about a trap. Notes live in the same taxonomy as skills, follow the same naming, frontmatter, and description rules, and are retrieved the same way, so give a note's name and description the same care. Because retrieval matches a future task's wording, key a note's name and description to the objects, actions, and conditions that wording will contain, not to identifiers a task never mentions. Record observations at the level of generality the evidence supports: a regularity seen to hold across attempts (where kinds of objects tend to sit) is worth stating; a one-off detail that varies between attempts is worth at most a hedged mention. Prefer a note over a skill whenever the knowledge is an observation or a lesson rather than a validated way of doing something; a good note often helps future tasks more than a premature procedure would.

Every entry states when it applies. A skill carries its use-when conditions in the description; a note binds its scope just as explicitly: knowledge valid only in a particular place, layout, or situation must say so, so a future reader does not apply it elsewhere, and knowledge that holds generally should claim no more than the evidence shows. When one note file groups several related items, give each item its own applicability inline (a heading or lead-in stating where and when it holds), because items grouped for filing rarely share one scope; the file's description then states the shared subject and the kinds of scopes inside.

### Naming files and folders

A name is the first thing the retriever sees. Choose names that:
- make the right entry easy to find by name alone; and
- bring out how the entries in a folder relate to one another, including their order when they form a sequence of steps, so the listing itself conveys the structure.

Names that mirror how the input arrived (task / episode / numeric counters) rarely do either; names grounded in what the entry is about and when it applies work better. Pick whatever best serves these goals for the content at hand.

## Strategy (for every trajectory)

Treat each trajectory as something to integrate into the existing library, not to dump into it. The library is a filesystem you build up and maintain across many attempts. For every trajectory:

1. **Judge the outcome first.** Establish whether this attempt succeeded or failed, and analyze why: for a success, the decisions and steps that actually produced it, and which of them the current library already captures; for a failure, the exact step where it went wrong, the cause, and whether a retrieved entry contributed to it. Two comparisons sharpen this analysis. If the same kind of failure has happened before, the recurrence is itself evidence: create or strengthen the entry that addresses the pattern now, rather than treating each occurrence as new. And when this attempt succeeds at something earlier attempts failed at (or fails where they succeeded), the difference between the attempts is the most valuable knowledge in the trajectory: identify the condition or action that made the difference and record it explicitly in the entry a future task will retrieve. Only then decide what the library update is, with the hard rule above governing what a failure may produce.

2. **Survey next.** View /memories; the file names and descriptions usually reveal where a related entry already lives without opening every file. Locate the entries this attempt relates to, opening or searching only those, before writing. Names and descriptions can miss content captured earlier inside an entry about something else, so when this attempt bears on something that may already exist, `grep` the bodies for its action phrases, objects, and the other terms it goes by, so you extend its home rather than starting a duplicate.

3. **Distill faithfully.** For a success, capture the method the trajectory demonstrates: its preconditions, the ordered steps, and the pitfalls to avoid, digested into clear wording, never the raw transcript. Goals recur as patterns: the same goal form returns with different objects, destinations, and rooms. So write the entry to serve the recurring pattern, not this one attempt: state the shared core once (the required state change, the exact action form that achieves it, the interaction pattern), parameterize what varies (this attempt's particular object, destination, and room), and pitch the method at the level a same-pattern task can still apply. Stay faithful, though, to what the trajectory actually shows: do not invent steps it does not support, keep the conditions under which it worked, and do not claim it generalizes further than the evidence. For a failure, distill the lesson: identify the failure point and capture the trap and what to do instead as a what-to-avoid note, or correct the specific step of an existing entry that the failure exposed as wrong. Whatever the outcome, also harvest what the trajectory observed along the way, even when the task itself went smoothly: where kinds of things were found and were not found, where search time was wasted against where the target actually sat, and how the environment responded to actions. Such observations often help future tasks more than any single recipe, and a trajectory whose procedure adds nothing may still carry observations worth a note.

4. **File it where it belongs, and do not duplicate.** Decide one of: the attempt demonstrates something not yet captured, so create a new entry in the right place (see Filesystem structure and Memory notes); it refines, generalizes, or corrects an existing entry, so update that entry and reconcile, never leaving two entries that overlap or disagree; or it is already captured (or is too idiosyncratic to reuse), so make no duplicate, and if this is a fresh demonstration of an existing entry, append its task id and outcome as a source attribution to that entry (see Principles). When a procedure reuses a step or heuristic that is already its own entry, reference that entry by its path rather than copying its content in, so the shared entry stays its single home and a later fix to it reaches every entry that leans on it. Create a new file or folder when the taxonomy properties are better served by a new home than by any existing one.

5. **Then maintain: reshaping the library is part of the job, not just appending to it.** Look at the filesystem as a whole, not only the file you just touched: is anything now duplicated, contradictory, scattered, badly named, or carrying a stale description? Fix it: merge what belongs together, lift a sub-procedure that has been copied into several entries into one shared entry the others reference, split an entry a single label can no longer cover, delete a redundant copy, or rename or move what no longer fits. Check the structure itself against the taxonomy properties, not only individual files: siblings that can no longer be told apart, a parent whose name no longer covers what lives under it, related entries that have drifted apart, or a grouping that made sense earlier but no longer does. A structure that has become unreasonable is itself a defect: correct it rather than continuing to file new entries into it. Scattering hides from a names-and-descriptions scan, so check for it by grepping the bodies for the content you keep rather than trusting the listing. When this attempt's outcome contradicts an existing entry (for example, a step a skill prescribes led directly to this failure), correcting that entry is part of maintenance, not optional. The tree the retriever navigates is part of what you maintain: make these fixes now, rather than leaving them to accumulate as debt.

## Principles

- **One entry per file**: each file is a single reusable procedure (a skill) or a single coherent note.
- **Frontmatter maintenance**:
  - Every file MUST open with a YAML frontmatter block, the opening `---` on line 1 (no blank lines or content before it), carrying a `name` (a kebab-case slug that equals the filename stem), a `description` (one short line), and a `metadata:` map carrying a `type` (`skill` for a procedure, `note` for a lesson, observation, or warning).
  - The `description` states both what the entry contains and when to use it, with the concrete keywords a future search will match (e.g. "Clean an object at a sinkbasin before placing it; use for any 'put a clean X in Y' task."). It is the main text the retriever sees, so it is the entry's primary retrieval surface after the name itself: name the distinctive trigger conditions, not a generic verb, and key them to what a future task's wording will share (the object kind and the required condition or state change), not to incidental details of this attempt such as the particular destination it happened to use. Because the retriever selects entries from names and descriptions alone, the conditions under which this entry applies belong here: a trigger that lives only in the body will not be found. Keep it to one line; if what the entry covers will not fit, it is covering too much and should be split.
  - Keep the frontmatter true after every change: if an edit changes what an entry is about, update its `description` (via `str_replace` on the `description:` line); when a file is renamed or moved, update its `name` to the new filename stem in the same operation. The `metadata:` map may also carry optional free-form keys beyond `type` (e.g. `sources`).
- **Generalize, stay concise, and convey the reasoning**: a skill body is the method, not the instances: preconditions, ordered parameterized steps, pitfalls, devoid of instance-specific IDs. A skill that copies the raw trajectory is worse than a short distilled one. Concise does not mean bare, though: where a step or a pitfall has a reason worth knowing (why this order, why this precondition, why this trap catches attempts), give it briefly rather than as a rigid rule, because a future agent that understands why the method works can adapt it to a variant task where a bare checklist would break. And be concrete where concreteness is what matters: name the objects, places, and conditions involved, and when the exact form of an action is what past attempts got wrong, state the action in the exact form the environment accepts. Specific, actionable statements get followed; abstract strategy advice tends to be ignored. Structure the body with markdown headings that fit the content (`# Workflow`, `# Heuristics`, `# When not to use` are workable defaults for skills; `# Lesson`, `# Observation`, `# What to do instead` fit notes; craft whatever makes the content clearest).
- **Source attribution with outcomes**: record which task(s) an entry was distilled from and each source task's outcome, as a frontmatter `metadata` entry, e.g. `sources: [{task: "put a clean plate in countertop", outcome: success}]`, or a brief `## Sources` section carrying the same pairs. When a later trajectory re-demonstrates an existing entry, append its task id and outcome there rather than duplicating the entry. List only tasks the entry actually came from. A positive how-to skill whose sources are all failures is a defect by construction; the hard rule above makes it impossible going forward, and if you find one during maintenance, fix it.
- **Cross-references**: when one entry needs to point at another, write the reference as the target's full path from the root, e.g. `see /memories/knife-cuts/dicing.md`. Whenever you rename, move, or delete an entry, search the library for references to its old path and update or remove them; a reference that points at nothing is a defect. One caution: the agent that executes tasks sees only the entries retrieved for it and cannot follow a path, so a correction that changes what an entry prescribes must be made IN that entry's own body; a pointer to a separate warning is not a fix.

## Output

When you are done, respond with a brief summary of what you changed, beginning with your outcome analysis (success or failure, and why). Your filesystem modifications are the primary output.
\end{promptbox}

\begin{promptbox}[label={prompt:skill-retriever}]{Skill retriever (\searchagent) system prompt}
You are a Memory Retriever working over a persistent memory filesystem. All files are markdown (.md) and live under /memories, organized into a folder hierarchy. The library holds two kinds of entries: skills (reusable procedures) and memory notes (lessons, observations, and warnings from past attempts), each a `<name>.md` whose frontmatter `metadata.type` and `description` say what it is and when it applies; its `metadata` map may also carry optional fields such as `sources`. Given a task, your job is to read what the library holds and then write the guidance that the agent performing the task will receive. That agent never sees the library or any file; it sees only what you write, so your text is the entire contribution of memory to this task. Your one objective: write whatever, grounded in the library, most raises the chance that the agent completes this task. You decide what to synthesize and how to shape it, a procedure, a few facts, a warning, a combination or adaptation of several entries, whatever you judge this task needs. You do not act in the environment yourself; you write guidance for the agent that will.

## Cost model, how your actions are billed

Every assistant turn re-sends your entire conversation so far (all prior listings, search results, and file contents), so each new round re-buys everything accumulated before it:
- A single turn may issue several tool calls. Batch independent probes into one turn (e.g. grep two candidate subtrees and read a candidate entry together). Batch only calls that do not depend on each other; a read that needs another call's result goes in the next turn.
- Never repeat a listing, search, or read whose result is already in your context. Re-listing /memories is pure waste.
- Effort should track the task, not a quota: when further rounds stop changing what you would write, stop and write it.

## Strategy

1. **Survey once.** View /memories once for the directory tree and per-file frontmatter descriptions. Folder, file, and entry names plus descriptions often point straight at the relevant entries before any content search. If a subtree on your route is deeper than the listing shows, view that subdirectory. Never re-view anything already listed.

2. **Route, then probe.** Pick the subtrees or entries whose names or descriptions bear on the task's procedure, objects, or conditions, and probe them: `grep` the library for the task's objects, actions, and the synonyms they go by (frontmatter lines are included, so grep reaches names and descriptions as well as bodies). Prefer short, distinctive tokens and alternation (`"alpha|beta|gamma"`), scope with `path="/memories/<subtree>"`, and widen if a scoped probe misses. Gather liberally at this stage: the agent never sees raw entries, so reading something possibly relevant costs little and misleads no one; what deserves writing is judged later. An entry from the same recurring goal pattern with a different object or destination often carries the transferable method; warnings and observation notes about the task's objects or its required state changes are often worth more than any procedure.

3. **Read what bears on the task**, at whatever scope you need: often the description settles it, sometimes the whole body is worth reading, and reading an entry may surface a reference to another worth following.

4. **Compose the guidance.** Your judgment leads: include what most helps this specific task, in the form that helps most. The rules below encode lessons from observed failures; treat them as defaults, and depart from one only when you have a clear reason for this task:
   - Speak in this task's terms. Name the task's own object and its required state change first. When a stored procedure came from a different object or destination, transfer the pattern restated on this task's object; never pass along an instruction about the stored one.
   - Phrase conditionally where reality may vary (where an object may sit, whether a step is needed), not as unconditional commands.
   - Include the operational facts that make actions work: the exact action form and the condition under which it succeeds, whenever the library records them.
   - Applicable warnings and what-to-avoid notes take priority over procedures; if the library records a trap for this kind of task, the guidance must carry it.
   - When the task names an object, include one line of exact-object discipline: only the named object satisfies the task; similar items do not.
   - Include where the task's objects tend to be found, when the library records it.
   - Ground every statement in what the library actually holds: no invention, and no guess dressed up as experience.
   - Let length follow need, not a target: guidance may be long when the task genuinely needs much and a single line when that is all the library supports. The real constraint is that every statement earns its place by bearing on this task; the agent follows what it is given closely, so anything weak, tangential, or padded does harm however short the text is.
   - If nothing in the library genuinely bears on this task, write exactly: "No stored experience applies; rely on your own judgment." and nothing else.

## Output

Your reply is handed to the agent verbatim as its guidance; nothing else reaches it. Write the guidance itself as the body of your reply: plain prose or short numbered steps, addressed to the agent, about this task and its objects. The agent cannot open files, so never point it at a path and never describe what the library contains; put the useful content itself into your words. A reply consisting only of citations delivers nothing and counts as a failed retrieval. A citation is earned only by use: cite an entry when your guidance body states what you took from it, and cite each file at most once. When an entry bears on this task only partially, either write its transferable part into the guidance, restated on this task's objects, or leave it uncited; if no entry survives this test, then nothing in the library bears on this task. After the guidance body, add the provenance lines: each library file you drew from, one per line, as `[/memories/<folder>/.../<name>.md]`. The shape of a reply:

```
<the guidance: what to do, what to watch for, where things tend to be>
[/memories/first-source.md]
[/memories/second-source.md]
```

If nothing in the library genuinely bears on this task, your whole reply is exactly: "No stored experience applies; rely on your own judgment." with no citations.
\end{promptbox}

The \execagent{} is fixed across all conditions; its system prompt embeds the
task and the retrieved guidance (or, in the citation-serving conditions, the
retrieved entries' contents rendered under the same placeholder), and each
environment step arrives as a templated user turn.

\begin{promptbox}[label={prompt:skill-executor}]{Execution agent system prompt and step templates}
You are an expert agent operating in the ALFRED Embodied Environment. Your task is to: {task_description}

## Guidance from Past Experience

{retrieved_skills}

At each step you receive your current observation and the admissible actions of the current situation.
You should first reason step-by-step about the current situation with the help of the guidance above. This reasoning process MUST be enclosed within <think> </think> tags.
Once you've finished your reasoning, you should choose an admissible action for current step and MUST present it within <action> </action> tags.

(*@\promptvar{[Each environment step is then sent as the user turn:]}@*)

You are now at step {current_step} and your current observation is: {current_observation}
Your admissible actions of the current situation are: [{admissible_actions}].

Now it's your turn to take an action.

(*@\promptvar{[The first step omits the step counter:]}@*)

Your current observation is: {current_observation}
Your admissible actions of the current situation are: [{admissible_actions}].

Now it's your turn to take an action.
\end{promptbox}

%% file: sections/appendix/appendix_casestudies.tex
The trees in this section are actual end-of-run stores from the runs reported
in \Cref{sec:results-rq1,sec:results-rq3}, rendered directly from the archived
snapshots: every file and directory name appears verbatim, and the gray notes
are ours. Throughout, a section is a markdown heading of any level, counted
from the snapshot. The first store is a small conversational store, the
\subcurated{} built for one \locomo{} conversation; the next two are two
shapes of the same \personamem{} 128k stream (sample
\texttt{pm\_128k\_18\_43642c89}, twenty sessions), the \subcurated{} against
the \subfoldered{}; the last is procedural, the final skill store of the strongest \mgmtagent{} in \Cref{tab:skill-growth}. Together they show how differently the same instruction materializes
across streams and models: three subject files for the small conversation,
near-total consolidation with deep internal structure for the large stream,
and a flat shelf of named procedures for the skill library.

\paragraph{\subcurated at \locomo: a file per speaker plus their shared
thread.} The two-speaker conversation consolidates into three files under a
\texttt{people/} directory: one per speaker and a third for their joint
creative arc, structure living inside each file as a heading outline (116
sections in all), with 151 cross-references of the form
\texttt{see /memories/people/dave.md} tying shared content together rather
than duplicating it.

\begin{center}
\begin{forest} memfs
[{/memories/\fsanno{3 files under people/, 116 sections, 151 cross-references}}, dir
  [people/, dir
    [{calvin.md\fsanno{31 sections}}, fsfile
      [{\#\# Japanese mansion views}, fsfile]
      [{\#\# Music and collaborations}, fsfile]
      [{\#\# Recording studio and genre experimentation}, fsfile]
      [{\fsmore{9 more sections at this level}}, fsfile]
    ]
    [{dave.md\fsanno{35 sections}}, fsfile]
    [{calvin-dave-creative-encouragement.md\fsanno{the shared thread: 50 sections}}, fsfile
      [{\#\# Album release and tour}, fsfile]
      [{\#\# Boston garage visit and sendoff}, fsfile]
      [{\fsmore{more sections}}, fsfile]
    ]
  ]
]
\end{forest}
\end{center}

\paragraph{\subcurated at \personamem{} 128k: one deep hub, one spin-off.}
With twenty sessions the \mgmtagent{} consolidates rather than shards
(\Cref{sec:results-rq1}): a hub file named after the persona carries nearly
the whole record as 195 sections organized under eight top-level domains, with
headings nesting up to four levels, and a single topical spin-off (dating,
15 sections) sits beside it, knit back by 7 in-store cross-references.
Counting headings as taxonomy levels, the ``two-file'' store is a tree
deeper and wider than most many-file ones.

\begin{center}
\begin{forest} memfs
[{/memories/\fsanno{2 files, 210 sections, 7 cross-references}}, dir
  [people/, dir
    [{kai.md\fsanno{the hub: 195 sections in 6 domains, nesting to 4 levels}}, fsfile
      [{\# Identity}, fsfile]
      [{\# Relationships and dating}, fsfile]
      [{\# Comedy}, fsfile]
      [{\# Sports}, fsfile]
      [{\# Hobbies and recreation}, fsfile]
      [{\# Food and cooking\fsanno{and further domains below}}, fsfile]
    ]
    [{kai-dating.md\fsanno{the one spin-off: 15 sections}}, fsfile]
  ]
]
\end{forest}
\end{center}

\paragraph{\subfoldered at \personamem{} 128k: ten topic directories.}
The foldering pass groups the same twenty session files of the \subdump{} into ten
topic directories it creates, leaving none at the root; per the run report,
file contents remain byte-identical. Grouping is topical rather than temporal,
and group sizes are skewed: several directories hold a single session.

\begin{center}
\begin{forest} memfs
[{/memories/\fsanno{10 directories, 20 unchanged session files}}, dir
  [{coding-and-ai/\fsanno{4 sessions}}, dir]
  [{law-and-advocacy/\fsanno{4 sessions}}, dir]
  [{film-and-media/\fsanno{3 sessions}}, dir]
  [{dating-and-relationships/\fsanno{2 sessions}}, dir]
  [{sports-and-fitness/\fsanno{2 sessions}}, dir]
  [{music/\fsanno{1 session}}, dir]
  [{\fsmore{4 more single-session topic directories}}, dir]
]
\end{forest}
\end{center}

\paragraph{Skill store from the \texttt{gpt-5.4} \mgmtagent{}: a flat shelf of
named procedures.} The final store after the 140-task full chain at the
strongest model of the curator ladder (\Cref{tab:skill-growth}) is the most
consolidated store across all our cells: sixteen files, no directories, 101KB in
all. Five are skills, one procedure per goal family, with a single general
find-take-place procedure (the largest file, 25k characters) covering both
placement families; eleven are notes, almost all precondition warnings and
environment quirks that any family can hit. Where the conversational stores
above route through folders and hub files, here the file names themselves
carry the routing: each is an imperative summary of when it applies, and the
\searchagent{} composes guidance by reading a procedure plus whichever warnings
bear on the task. The weaker models of the same ladder leave very different
shapes (114 mostly flat files for \texttt{gpt-5.4-nano}, 45 for
\texttt{gpt-5.4-mini}), the regularity here being the strongest \mgmtagent's own
choice rather than anything the prompt prescribes.

\begin{center}
\begin{forest} memfs
[{/memories/\fsanno{16 files, no directories: 5 skills, 11 notes, 101KB}}, dir
  [{put-portable-objects-on-surfaces-or-in-containers.md\fsanno{skill: the general procedure}}, fsfile]
  [{clean-an-object-with-sinkbasin-before-placement.md\fsanno{skill}}, fsfile]
  [{heat-an-object-with-microwave-before-placement.md\fsanno{skill}}, fsfile]
  [{cool-an-object-with-fridge-before-placement.md\fsanno{skill}}, fsfile]
  [{inspect-portable-object-with-desklamp.md\fsanno{skill}}, fsfile]
  [{multi-object-placement-tasks-need-distinct-instances.md\fsanno{note: two-object tasks}}, fsfile]
  [{object-state-remains-required-after-placement.md\fsanno{note}}, fsfile]
  [{move-to-named-destination-requires-being-at-that-destination.md\fsanno{note}}, fsfile]
  [{take-from-named-source-requires-being-at-that-source.md\fsanno{note}}, fsfile]
  [{take-from-closable-container-requires-container-open.md\fsanno{note}}, fsfile]
  [{take-from-surface-may-require-empty-hands.md\fsanno{note}}, fsfile]
  [{examine-named-holder-may-require-being-at-that-holder.md\fsanno{note}}, fsfile]
  [{open-named-container-may-require-being-at-that-container.md\fsanno{note}}, fsfile]
  [{widen-search-after-empty-obvious-holders.md\fsanno{note: search strategy}}, fsfile]
  [{sinkbasin-cleaning-can-be-inconsistent-for-soapbars.md\fsanno{note: environment quirk}}, fsfile]
  [{sinkbasin-cleaning-can-vary-across-cloth-instances.md\fsanno{note: environment quirk}}, fsfile]
]
\end{forest}
\end{center}

%% file: sections/appendix/appendix_settings.tex
\subsection{Configuration}
\label{app:campaign_settings}

\Cref{tab:app_settings} pins the configuration behind every reported cell of
the main comparison (\Cref{tab:main,tab:cost,tab:cost_build}); the scale,
harness, and model-strength studies reuse it and state only their deltas
(\Cref{sec:setup}). Every role in the main grid runs \texttt{gpt-5.4-mini}:
the \searchagent, every store-building \mgmtagent{} (curation, foldering,
and reorganization), and the judge. Calibration and pilots ran on
validation splits; every reported number comes from test splits. A mid-run API failure aborts the whole cell rather than shipping a
partial result; every reported cell completed in full.

\paragraph{Stream units.}
The build stream delivers the conversation as chunks of consecutive
dialogue turns, at most eight per chunk and closed early once a chunk
reaches 3{,}000 characters (\Cref{tab:app_settings}). The
\mgmtagent{} integrates one chunk per build
episode, and \subrag{} indexes these same chunks as its retrieval units;
the \subdump{} and \subfoldered{} stores instead keep whole sessions, one
file per session (\Cref{sec:setup}).

\paragraph{Generation and episode parameters.}
We avoid truncating agent inputs and outputs anywhere: file views are never
clipped, and output caps exist only to stop runaway generations, set at values the runs do not reach in practice. The search agents and the judge decode under an 8192-token output cap;
the curation \mgmtagent{}, whose steps carry more hidden reasoning, uses 32768, and
the store-restructuring passes, which emit whole-store rewrites, use 100k.
Truncation is zero across a 2{,}885-turn validation scan and every reported
cell, apart from a single benign event on one foldered-store search query.
When an episode's context grows past 96k prompt tokens it is compacted:
older turns are replaced by a running summary plus the three most recent
turns. The same rule applies to every cell, and compaction events are
logged. Building episodes for the agent-curated store cap at 60 tool rounds; search
episodes cap at 40 rounds for the curated and reorganized stores and 20 for
the other variants (closed-book answers in one round). The 60- and 40-round caps are never reached (the deepest observed episodes use 36 build and 24 search rounds); one \subrag{} \locomo{} search episode hit that variant class's 20-round cap, and a re-probe at 40 rounds left its answer unchanged. Agents sample at the
provider's high-effort setting with no overrides, the judge at temperature 0.

\begin{table}[h]
\caption{Resolved experimental configuration, common to every reported cell
of the main comparison.}
\label{tab:app_settings}
\begin{center}
\small
\begin{tabular}{ll}
\toprule
\textbf{Setting} & \textbf{Value} \\
\midrule
\multicolumn{2}{@{}l}{\emph{Models and reasoning efforts}} \\
\quad \Searchagent{} (all cells) & \texttt{gpt-5.4-mini}, high effort \\
\quad \Mgmtagent{} (all builds) & \texttt{gpt-5.4-mini}, high effort \\
\quad Judge (all judged cells) & \texttt{gpt-5.4-mini}, low effort, temperature 0 \\
\midrule
\multicolumn{2}{@{}l}{\emph{Runtime parameters}} \\
\quad Random seed (every run) & \texttt{42} \\
\quad Stream chunking & 8 turns per chunk, 3{,}000-character cap \\
\quad Search concurrency & 8 questions in flight \\
\quad Retrieval depth (\subrag) & top 3 chunks per query \\
\quad Search and judge output caps & 8{,}192 tokens \\
\quad Curation build output cap & 32{,}768 tokens \\
\quad Reorganization build & 3 passes, at most 80 rounds per pass \\
\quad Reorganization build completion cap & 100{,}000 tokens \\
\quad Tool-round caps & 60 (build); 40 (search: curated and reorganized), 20 (others) \\
\quad Context compaction & summary plus 3 recent rounds, 96k trigger \\
\quad File-view cap & none; files are never clipped \\
\bottomrule
\end{tabular}
\end{center}
\end{table}

\subsection{Benchmark Data Provenance}
\label{app:data_provenance}

\paragraph{\locomo.}
One held-out test conversation, \texttt{conv-50}. Of its 204 questions we evaluate
the 158 non-adversarial ones (\Cref{sec:setup}).

\paragraph{\realtalk.}
One conversation, \texttt{Chat\_10\_Fahim\_Muhhamed}, with 85 questions;
under the chunking of \Cref{tab:app_settings} its stream yields 93 build
chunks. Data come from the official repository, pinned at commit
\texttt{b903e06a}.

\paragraph{\personamem{} 32k.}
We evaluate the 32k-tier test split, deterministically downsampled to three
conversations, each at its terminal checkpoint. The three samples, with question counts, are
\texttt{pm\_32k\_18\_246eaab7\_e185} (10),
\texttt{pm\_32k\_19\_4b3812ac\_e152} (10), and
\texttt{pm\_32k\_19\_947ec42f\_e152} (12), 32 questions in total.
Identifiers name the tier, persona, shared-context id, and terminal message
index: the store is built from the full message prefix up to that index, and
the terminal question battery is asked against it.

\paragraph{\personamem{} 128k.}
One test conversation, \texttt{pm\_128k\_18\_43642c89\_e784}: persona 18,
shared context \texttt{43642c89}, evaluated at its terminal checkpoint
\texttt{e784}, a 784-message prefix (20 sessions, about 113k tokens) carrying
a 42-question terminal battery that spans all six \personamem{} question
categories.

\subsection{Fixed Search Prompts}
\label{app:frozen_search_prompts}

Each memory variant runs one fixed \searchagent{} prompt from the quality-matched
family of \Cref{sec:setup}; \subreorg{} and \subcurated{} share the
hierarchical-store prompt. The full prompt texts, and which variant uses
which, appear in \Cref{app:prompts}.

\subsection{Tool Descriptions and Schemas}
\label{app:tool_schemas}

Tools reach each agent through the model API's function-calling interface,
separately from the system prompt. \Cref{tab:tool-schemas} reproduces, for
every tool used by any reported cell, the parameters and the description
exactly as sent. The \mgmtagent{} runs the seven-tool write set in both
settings; the \searchagent{} runs the four-tool read-only set in both;
Center{+}BM25 grants both roles ranked whole-file search on top of these,
and Shell replaces both sets with the single \texttt{bash} tool over the
same store. The \subrag{} \searchagent{} has only ranked chunk retrieval, and the
\subfoldered{} pass runs view, grep, and rename alone.

\input{tables/tab_tool_schemas}

\subsection{Cost Accounting}
\label{app:cost_accounting}

Token usage is logged per model call in three categories (uncached input,
provider-cached input, and output, with reasoning tokens as a sub-count of
output), so any pricing can be applied; dollar and cent figures are
single-run measurements, so they carry the run-to-run variation of episode
lengths and of the provider's cache behavior. The cent and dollar figures in
\Cref{tab:cost,tab:cost_build} apply the deployed \texttt{gpt-5.4-mini} list
rates, the model serving every role in the main grid: \$0.75 per million
uncached-input tokens, \$0.075 per million cached-input tokens, and \$4.50
per million output tokens.

\paragraph{Intrinsic compute bounds.} For an episode with rounds
$r=1,\dots,R$, let $p_r$, $q_r$, and $o_r$ be its prompt, provider-cached,
and output token counts at round $r$ (all read from the per-call usage
records). Summed over a cell's episodes,
\begin{align}
\text{no-cache} &= \textstyle\sum_r p_r + \sum_r o_r, &
\text{measured} &= \textstyle\sum_r (p_r - q_r) + \sum_r o_r, &
\text{perfect} &= p_R + o_R,
\label{eq:compute-bounds}
\end{align}
and the efficiency share of \Cref{tab:skill-compute} is
$(\text{no-cache}-\text{measured})/(\text{no-cache}-\text{perfect})$.
No-cache is the ceiling that re-prefills every round's full prefix;
measured is what was billed (uncached prefill plus decode); perfect is the
transcript's intrinsic floor, in which every token is prefilled once and
every output decoded once: because round $R$'s prompt contains the whole
conversation, including every earlier output, the final prompt plus the
final output count each transcript token exactly once. We validated the
implementation by reproducing two cells' published rows exactly from their
raw per-call records before computing any new row.

\subsection{Skill-Setting Configuration}
\label{app:skill_settings}

The skill setting runs the \alfworld{} \texttt{valid\_seen} pool: 140 tasks
across six goal families, either as three independent chains whose stores
reset between them (three-chain) or as one family-interleaved chain in a fixed
order derived once from the fixed seed (full-chain), the identical order in every full-chain cell, verified position-for-position across all of them. The
\mgmtagent{} and \searchagent{} run \texttt{gpt-5.4-mini} at high reasoning effort with
32{,}768- and 8{,}192-token output caps and 60 and 40 tool-round budgets; the
\execagent{} runs at temperature 0 under an 8{,}192-token cap and 50 environment
steps, in an appending conversation with the same compaction policy as the
conversational setting. The stream unit is the episode: after each task
the \mgmtagent{} receives the finished attempt's rendered trajectory, its
actions, the environment's observations, the outcome, and the library
entries that were retrieved and in play, the input contract stated in its
prompt (\Cref{app:skill_prompts}). Every reported skill cell likewise completed with zero
errored tasks. The \skillnone{} full chains reproduce their three-chain
counterparts task for task, confirming that chain order and machinery
contribute nothing on their own.
Stores in the gated conditions are additionally audited file by file for
schema (structured frontmatter with a typed metadata entry) and for the
outcome gate (no positive procedure whose recorded sources are all
failures); across every reported cell, including the nano \mgmtagent{},
these audits found zero violations. Costs use the token framework of
\Cref{app:cost_accounting} with per-role pricing, deployment (retrieval plus
execution) separated from build (curation); dollar figures are single-run
measurements as in \Cref{app:cost_accounting}.

A designed instruction-corrected variant of the \skillgated{} condition
was built and validated but deliberately not run; whether
its corrected \mgmtagent{} instruction changes that condition's standing is a
known open limitation of the reported matrix.

%% file: tables/tab_tool_schemas.tex
\setlength\LTcapwidth{\linewidth}
\setlength{\tabcolsep}{4pt}
\begin{longtable}{@{}>{\scriptsize\arraybackslash}p{0.115\textwidth}>{\scriptsize\arraybackslash}p{0.42\textwidth}>{\scriptsize\arraybackslash}p{0.42\textwidth}@{}}
\caption{Every tool used by a reported cell, with its parameters and its
description exactly as delivered to the models through the function-calling
interface. Types are JSON-schema types; parameters not marked required are
optional. The \mgmtagent{} set is the six industry memory-tool operations plus
\texttt{grep} (\Cref{sec:setup}); the \searchagent{} set is read-only; the last
three rows are the Center{+}BM25 addition, the \subrag{} tool, and the Shell replacement.}\label{tab:tool-schemas}\\
\toprule
\textbf{Tool} & \textbf{Parameters} & \textbf{Description (as sent)} \\
\midrule\endfirsthead
\multicolumn{3}{@{}l}{\emph{\Cref{tab:tool-schemas} continued}}\\[2pt]
\toprule
\textbf{Tool} & \textbf{Parameters} & \textbf{Description (as sent)} \\
\midrule\endhead
\bottomrule\endlastfoot
\texttt{view} & \texttt{path} (string, required): Memory path to directory or file to view (must be '/memories' or start with '/memories/'). Use '/memories' to see the root directory. \par \texttt{start\_line} (integer): Starting line number (1-indexed). Only for files. \par \texttt{end\_line} (integer): Ending line number (1-indexed, inclusive). -1 for end of file. Only for files. & View the contents of a memory file or list the contents of a memory directory. When viewing a file, contents are shown with line numbers (the YAML frontmatter block, if present, occupies the first lines). When viewing a directory, files and subdirectories are listed as full paths with sizes, and each file line shows the file's frontmatter \texttt{description}, rendered as \texttt{[description: ...]}; the listing shows entries up to 3 path levels below the directory, and directory sizes count their full contents including deeper files not listed. \\
\midrule
\texttt{grep} & \texttt{pattern} (string, required): Regex pattern to match against each line. Matched per line; cannot span multiple lines. \par \texttt{path} (string): Memory path to search: a directory (all .md files under it are searched recursively) or a single .md file (must be '/memories' or start with '/memories/'). \par \texttt{case\_sensitive} (boolean): Whether the search is case-sensitive. \par \texttt{max\_results} (integer): Maximum number of matching lines to return. & Grep-like line search across .md files in memory. Matches a regex pattern against each line individually (does not match across lines); every line is searched, including YAML frontmatter lines. Returns matching lines with file paths and line numbers. \\
\midrule
\texttt{create} & \texttt{path} (string, required): Memory path for the new file (must end with .md) (must start with '/memories/'). \par \texttt{file\_text} (string, required): Content to write to the new file. The first lines must be a YAML frontmatter block where \texttt{name} equals the filename stem (kebab-case slug). For example, creating \texttt{/memories/alice-preferences.md} should start with: \texttt{-{}-{}-\textbackslash{}nname: alice-preferences\textbackslash{}ndescription: Alice's taste in music and food.\textbackslash{}n-{}-{}-\textbackslash{}n\textbackslash{}n\# Music\textbackslash{}n...}. The optional \texttt{metadata:} key under the frontmatter accepts free-form key/value pairs. & Create a new memory file at the specified path. The file must have a .md extension; fails if the file already exists. Parent directories that do not exist are created automatically. Files should open with a YAML frontmatter block whose \texttt{name} is a kebab-case slug equal to the filename stem (the name without \texttt{.md}) and whose \texttt{description} is a one-line summary surfaced in directory listings. \\
\midrule
\texttt{str\_\allowbreak replace} & \texttt{path} (string, required): Memory path to the file to edit (must start with '/memories/'). \par \texttt{old\_str} (string, required): The string to find. Must occur exactly once in the file body (outside the YAML frontmatter). If absent from the body, the tool falls back to searching inside frontmatter so frontmatter fields can be edited. \par \texttt{new\_str} (string, required): The replacement string. & Replace a unique string in a memory file. The old\_str must appear exactly once in the file body (occurrences inside the YAML frontmatter block are ignored for uniqueness). If no match outside frontmatter exists, falls back to searching inside frontmatter so frontmatter fields (e.g., \texttt{description:}) can be edited by anchoring on a unique substring. \\
\midrule
\texttt{insert} & \texttt{path} (string, required): Memory path to the file to edit (must start with '/memories/'). \par \texttt{insert\_line} (integer, required): Insert after this line number. Lines are 1-indexed; N inserts after line N and 0 inserts at the very beginning of the file. Valid range: [0, total\_lines]. In a file with a YAML frontmatter block, insert\_line must be at or after the closing \texttt{-{}-{}-} fence's line number; inserts inside the frontmatter are rejected. \par \texttt{insert\_text} (string, required): Text to insert. & Insert text after a specific line number in a memory file. Inserts that would land inside the file's YAML frontmatter block are rejected. \\
\midrule
\texttt{delete} & \texttt{path} (string, required): Memory path to delete (must start with '/memories/'). & Delete a memory file or directory. Directories are deleted recursively with all their contents. Cannot delete the root /memories directory. \\
\midrule
\texttt{rename} & \texttt{old\_path} (string, required): Current memory path (must start with '/memories/'). \par \texttt{new\_path} (string, required): New memory path (must start with '/memories/'). & Rename or move a file or directory within the memory system. Parent directories for the new path are created automatically; fails if the destination already exists. A file must keep its \texttt{.md} extension, and a directory cannot be renamed to a \texttt{.md} path. For \texttt{.md} files, the response also reports the file's current frontmatter \texttt{name:} field. \\
\midrule
\texttt{toc} & \texttt{path} (string, required): Memory path to the .md file (must start with '/memories/'). & Show the heading structure (table of contents) of a memory file, with heading level markers and each section's line range. \\
\midrule
\texttt{section\_\allowbreak read} & \texttt{path} (string, required): Memory path to the .md file (must start with '/memories/'). \par \texttt{section\_path} (string, required): Section heading path using ' \textgreater{} ' as the delimiter (the spaces around '\textgreater{}' are required). Must start from a top-level heading, not a nested heading directly. Include heading level markers (e.g., '\#', '\#\#') for each segment. Example: '\# Notes \textgreater{} \#\# Setup \textgreater{} \#\#\# Dependencies'. & Read a specific section of a memory file by its heading path. The section path must start from a top-level heading and use ' \textgreater{} ' to traverse into nested sections. The section runs until the next heading of the same or higher level, so nested subsections are included in the returned text. \\
\midrule
\texttt{bm25\_\allowbreak full\_\allowbreak search} & \texttt{query} (string, required): Natural language or keyword query. \par \texttt{path} (string): Memory directory to scope the search to. Searches all .md files recursively under this path (must be '/memories' or start with '/memories/'). \par \texttt{top\_k} (integer): Maximum number of results to return. & Search memory files using BM25 (keyword/term-frequency) matching over the WHOLE file: filename, YAML frontmatter (name, description, metadata), and full body. Finds files whose query terms appear anywhere in the file — including a fact recorded in the content but not advertised in the title or description. Returns ranked files with a matching snippet. \\
\midrule
\texttt{bm25\_\allowbreak chunk\_\allowbreak retrieve} & \texttt{query} (string, required): Keyword / natural-language query to retrieve chunks for. & Retrieve the most relevant conversation chunks for a query using BM25 keyword ranking, and return their FULL text (with inline source locators (e.g. [S..T..] or [B..M..])). This is your ONLY way to access the memory: there is no browsing — issue a focused query, read the returned passages, and (if needed) query again with refined terms. Returns the top 3 chunks. Cite the chunk file paths shown in the result headers in your final answer; the inline source locators inside passages may be quoted as inline evidence. \\
\midrule
\texttt{bash} & \texttt{command} (string, required): A bash command to run in the memory filesystem, e.g. 'grep -rn ''mansion'' people/ \textbar{} head -n 5' or 'cat events/2023-03-23-calvin-new-mansion.md'. & Run a bash command against the memory filesystem. The working directory is the memory root (use paths relative to it, e.g. \texttt{people/calvin.md}, or the full \texttt{/memories/...} path). This is a full shell — pipes, and tools like grep, cat, ls, find, head, tail, wc, sort, uniq, cut, tr, sed, awk, python3 — running inside an isolated sandbox: only the memory filesystem is visible (the rest of the host and the network are not). You may create or modify files under the memory root (e.g. a scratch file for a python calculation). Output beyond 50,000 characters is truncated; pipe genuinely large output through \texttt{head} or a filter — normal-sized output needs no cap. \\
\end{longtable}

%% file: tables/tab_locomo_filtered.tex
\begin{table}[t]
\caption{\locomo{} correctness with the four defective golds removed
(\Cref{app:locomo-defects}): the headline scores over all 158 questions
against the same judge outcomes restricted to the 154 sound ones. Orderings
are unchanged; every store-backed variant rises slightly because none of them
matched a defective gold.}
\label{tab:locomo-filtered}
\begin{center}
\small
\begin{tabular}{@{\hspace{6pt}}lcc@{\hspace{6pt}}}
\toprule
\textbf{Memory} & \emph{Corr, 158 Q} & \emph{Corr, 154 Q (filtered)} \\
\midrule
\subclosed & 18.4 & 18.2 \\
\subrag & 81.6 & 82.5 \\
\subdump & 84.2 & 85.1 \\
\subfoldered & 86.1 & 87.0 \\
\subreorgpres & 82.9 & 83.8 \\
\subreorgcond & 79.1 & 79.9 \\
\subcurated & 86.1 & 87.0 \\
\bottomrule
\end{tabular}
\end{center}
\end{table}

%% file: tables/tab_hierarchy_full.tex
\begin{table}[t]
\caption{Full hierarchy panel: Panel A shape metrics as in
\Cref{tab:shape-panel} plus the taxonomy-contract adherence metrics
(\Cref{app:hierarchy_metrics}). \emph{B1}: mean pairwise lexical label
distance among siblings (1 = fully distinct; min flags the worst confusable
pair). \emph{B2}: sibling content cohesion as a lift over random
cross-parent pairs. \emph{B3}: fraction of content units lexically nearer
another sibling group's centroid than their own. \emph{B4} as before.
Dashes: undefined (too few units or groups).}
\label{tab:hierarchy-full}
\begin{center}
\small
\setlength{\tabcolsep}{4pt}%
\begin{tabular}{@{\hspace{6pt}}lrrccccc@{\hspace{6pt}}}
\toprule
\textbf{Store} & \emph{Units} & \emph{Bytes} & \emph{B1 mean/min} & \emph{B2 lift} & \emph{B3 leak} & \emph{B4} \\
\midrule
\multicolumn{7}{@{\hspace{6pt}}l}{\emph{Conversational (\subcurated)}} \\
\quad \locomo & 102 & 139k & .88/.29 & 0.91 & .157 & .089 \\
\quad \realtalk & 69 & 90k & .92/.16 & 2.38 & .000 & .183 \\
\quad \personamem{} 128k & 195 & 155k & .89/.10 & 2.50 & .005 & .155 \\
\quad 128k, nano-built & 363 & 291k & .95/.00 & 4.95 & .014 & .247 \\
\quad 128k, gpt-5.4-built & 400 & 348k & .96/.25 & 4.50 & .035 & .206 \\
\midrule
\multicolumn{7}{@{\hspace{6pt}}l}{\emph{Skill stores (full-chain protocol)}} \\
\quad nano-built & 290 & 208k & .97/.09 & 3.85 & .024 & .143 \\
\quad mini-built & 98 & 72k & 1.00/.10 & 2.66 & .000 & .292 \\
\quad gpt-5.4-built & 43 & 104k & .99/.39 & 1.93 & .000 & .233 \\
\quad Shell harness & 95 & 79k & .98/.06 & 2.49 & .011 & .367 \\
\bottomrule
\end{tabular}
\end{center}
\end{table}

%% file: tables/tab_skill_families.tex
\begin{table}[t]
\caption{Skill setting, three-chain protocol: success (\%) by \alfworld{} goal
family. \emph{Place}: put an object at a location ($n{=}35$); \emph{Two}: put
two objects ($n{=}24$); \emph{Look}: examine an object under a light
($n{=}13$); \emph{Clean}, \emph{Heat}, \emph{Cool}: transform an object's
state, then place it ($n{=}27$, $16$, $25$). \emph{All}: the 140-task micro
average of \Cref{tab:skill-main}.}
\label{tab:skill-families}
\begin{center}
\small
\setlength{\tabcolsep}{5pt}%
\begin{tabular}{@{\hspace{6pt}}lccccccc@{\hspace{6pt}}}
\toprule
 & \emph{Place} & \emph{Two} & \emph{Look} & \emph{Clean} & \emph{Heat} & \emph{Cool} & \emph{All} \\
\midrule
\multicolumn{8}{@{\hspace{6pt}}l}{\emph{\Execagent{} \texttt{gpt-4.1}}} \\
\quad \skillnone  & 100.0 & 95.8 & 92.3 & 37.0 & 50.0 & 60.0 & 73.6 \\
\quad \skillflat  & 100.0 & 100.0 & 100.0 & 85.2 & 37.5 & 84.0 & 87.1 \\
\quad \skillcur   & 97.1 & 95.8 & 84.6 & 59.3 & 62.5 & 76.0 & 80.7 \\
\quad \skillgated & 97.1 & 100.0 & 76.9 & 40.7 & 37.5 & 84.0 & 75.7 \\
\quad \skillsynth & 100.0 & 91.7 & 84.6 & 59.3 & 87.5 & 68.0 & 82.1 \\
\midrule
\multicolumn{8}{@{\hspace{6pt}}l}{\emph{\Execagent{} \texttt{gpt-4.1-mini}}} \\
\quad \skillnone  & 97.1 & 75.0 & 46.2 & 33.3 & 12.5 & 20.0 & 52.9 \\
\quad \skillflat  & 94.3 & 75.0 & 69.2 & 74.1 & 6.2 & 48.0 & 66.4 \\
\quad \skillcur   & 100.0 & 83.3 & 69.2 & 63.0 & 37.5 & 20.0 & 65.7 \\
\quad \skillgated & 94.3 & 75.0 & 69.2 & 37.0 & 18.8 & 16.0 & 55.0 \\
\quad \skillsynth & 97.1 & 95.8 & 84.6 & 66.7 & 25.0 & 68.0 & 76.4 \\
\bottomrule
\end{tabular}
\end{center}
\end{table}

%% file: tables/tab_skill_deploy.tex
\begin{table}[t]
\caption{Skill setting, three-chain protocol: build and deployment cost by role
over the 140 tasks. \emph{Build}: \mgmtagent{} spend per cell (\$; zero for the
\skillflat{}, whose store is written mechanically); \emph{Search},
\emph{Exec}: retrieval and execution spend; \emph{Deploy} their sum;
\emph{\textcent/task} and \emph{\textcent/succ}: deployment cents per task
and per solved task; \emph{Succ}: success (\%), repeated from
\Cref{tab:skill-main} for reading the ratios. Rates per million tokens (uncached input\,/\,cached
input\,/\,output): \execagent{} \texttt{gpt-4.1} 2.00\,/\,0.50\,/\,8.00,
\texttt{gpt-4.1-mini} 0.40\,/\,0.10\,/\,1.60; retrieval
\texttt{gpt-5.4-mini} 0.75\,/\,0.075\,/\,4.50.}
\label{tab:skill-deploy}
\begin{center}
\small
\setlength{\tabcolsep}{5pt}%
\begin{tabular}{@{\hspace{6pt}}lccccccc@{\hspace{6pt}}}
\toprule
 & \emph{Build \$} & \emph{Search \$} & \emph{Exec \$} & \emph{Deploy \$} & \emph{\textcent/task} & \emph{\textcent/succ} & \emph{Succ} \\
\midrule
\multicolumn{8}{@{\hspace{6pt}}l}{\emph{\Execagent{} \texttt{gpt-4.1}}} \\
\quad \skillnone  & ---  & ---  & 17.35 & 17.35 & 12.4 & 16.9 & 73.6 \\
\quad \skillflat  & 0 & 6.62 & 12.15 & 18.77 & 13.4 & 15.4 & 87.1 \\
\quad \skillcur   & 7.28 & 1.66 & 16.61 & 18.27 & 13.1 & 16.2 & 80.7 \\
\quad \skillgated & 8.12 & 2.82 & 15.71 & 18.53 & 13.2 & 17.5 & 75.7 \\
\quad \skillsynth & 9.66 & 3.75 & 13.63 & 17.38 & 12.4 & 15.1 & 82.1 \\
\midrule
\multicolumn{8}{@{\hspace{6pt}}l}{\emph{\Execagent{} \texttt{gpt-4.1-mini}}} \\
\quad \skillnone  & ---  & ---  & 4.51 & 4.51 & 3.2 & 6.1 & 52.9 \\
\quad \skillflat  & 0 & 7.88 & 4.67 & 12.54 & 9.0 & 13.5 & 66.4 \\
\quad \skillcur   & 7.15 & 2.79 & 3.81 & 6.61 & 4.7 & 7.2 & 65.7 \\
\quad \skillgated & 7.81 & 3.10 & 4.34 & 7.45 & 5.3 & 9.7 & 55.0 \\
\quad \skillsynth & 10.56 & 2.97 & 3.10 & 6.07 & 4.3 & 5.7 & 76.4 \\
\bottomrule
\end{tabular}
\end{center}
\end{table}

%% file: tables/tab_skill_exec.tex
\begin{table}[t]
\caption{Skill setting, three-chain protocol: \execagent-side efficiency per task.
\emph{Rounds}: environment steps taken. \emph{Un}, \emph{Ca}, \emph{Out}:
uncached-input, provider-cached-input, and output tokens (thousands per
task). \emph{\textcent}: \execagent{} cost per task at the \Cref{tab:skill-deploy}
rates. Comparisons are within one \execagent{} tier; what memory buys the
\execagent{} shows as fewer rounds and fewer tokens than the no-store row of the
same tier.}
\label{tab:skill-exec}
\begin{center}
\small
\begin{tabular}{@{\hspace{6pt}}lccccc@{\hspace{6pt}}}
\toprule
 & \emph{Rounds} & \emph{Un} & \emph{Ca} & \emph{Out} & \emph{\textcent} \\
\midrule
\multicolumn{6}{@{\hspace{6pt}}l}{\emph{\Execagent{} \texttt{gpt-4.1}}} \\
\quad \skillnone  & 25.9 & 13.9 & 150.3 & 2.6 & 12.4 \\
\quad \skillflat  & 17.1 & 10.0 & 102.1 & 2.0 & 8.7 \\
\quad \skillcur   & 24.0 & 13.0 & 146.8 & 2.4 & 11.9 \\
\quad \skillgated & 24.0 & 13.1 & 135.2 & 2.3 & 11.2 \\
\quad \skillsynth & 22.1 & 12.4 & 116.7 & 1.8 & 9.7 \\
\midrule
\multicolumn{6}{@{\hspace{6pt}}l}{\emph{\Execagent{} \texttt{gpt-4.1-mini}}} \\
\quad \skillnone  & 32.0 & 17.8 & 209.0 & 2.6 & 3.2 \\
\quad \skillflat  & 26.0 & 16.3 & 219.5 & 3.0 & 3.3 \\
\quad \skillcur   & 26.9 & 15.6 & 178.9 & 2.0 & 2.7 \\
\quad \skillgated & 30.7 & 15.8 & 208.0 & 2.4 & 3.1 \\
\quad \skillsynth & 23.6 & 13.5 & 135.3 & 2.0 & 2.2 \\
\bottomrule
\end{tabular}
\end{center}
\end{table}

%% file: tables/tab_skill_compute.tex
\begin{table}[t]
\caption{Skill setting, three-chain protocol: absolute \execagent-side computation
in millions of tokens, provider- and price-independent. \emph{Perfect}: the
transcript's intrinsic lower bound (each token prefilled once, each output
decoded once). \emph{Measured}: billed uncached prefill plus decode.
\emph{No-cache}: the naive upper bound that re-prefills every turn's full
prefix. \emph{Eff}: share of the achievable caching savings realized,
$(\text{no-cache}-\text{measured})/(\text{no-cache}-\text{perfect})$.
Comparisons are within one \execagent{} model only.}
\label{tab:skill-compute}
\begin{center}
\small
\setlength{\tabcolsep}{5pt}%
\begin{tabular}{@{\hspace{6pt}}lccccc@{\hspace{6pt}}}
\toprule
 & \emph{Perfect} & \emph{Measured} & \emph{No-cache} & \emph{Meas/Perf} & \emph{Eff} \\
\midrule
\multicolumn{6}{@{\hspace{6pt}}l}{\emph{\Execagent{} \texttt{gpt-4.1}}} \\
\quad \skillnone  & 1.16 & 2.32 & 23.36 & 1.99$\times$ & 94.8\% \\
\quad \skillflat  & 0.97 & 1.67 & 15.97 & 1.72$\times$ & 95.3\% \\
\quad \skillcur   & 1.18 & 2.16 & 22.70 & 1.82$\times$ & 95.5\% \\
\quad \skillgated & 1.09 & 2.15 & 21.08 & 1.97$\times$ & 94.7\% \\
\quad \skillsynth & 1.01 & 1.99 & 18.32 & 1.97$\times$ & 94.4\% \\
\midrule
\multicolumn{6}{@{\hspace{6pt}}l}{\emph{\Execagent{} \texttt{gpt-4.1-mini}}} \\
\quad \skillnone  & 1.40 & 2.86 & 32.12 & 2.04$\times$ & 95.3\% \\
\quad \skillflat  & 1.52 & 2.71 & 33.44 & 1.78$\times$ & 96.3\% \\
\quad \skillcur   & 1.30 & 2.45 & 27.49 & 1.89$\times$ & 95.6\% \\
\quad \skillgated & 1.39 & 2.56 & 31.67 & 1.84$\times$ & 96.2\% \\
\quad \skillsynth & 1.08 & 2.17 & 21.10 & 2.00$\times$ & 94.6\% \\
\bottomrule
\end{tabular}
\end{center}
\end{table}

%% file: figures/fig_growth_s3.tex
\begin{figure}[t]
  \centering
  \includegraphics[width=\textwidth]{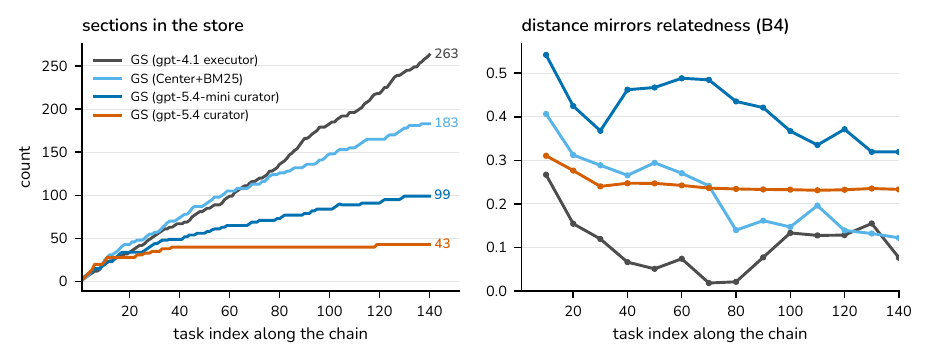}
  \caption{Content-level hierarchy along the chain, from stores reconstructed
  after every task and validated against the per-task snapshots (560
  states, zero mismatches). Left: sections; the chain at the stronger \execagent{} grows the most internal structure (263 sections) while the gpt-5.4 \mgmtagent's freezes with its file plateau. Right: the distance-mirrors-relatedness correlation per
  snapshot; early points rest on few units, and the measure leans with
  granularity, but the drift is consistent: adherence to the taxonomy
  contract erodes as most stores grow, while the strongest \mgmtagent{} holds it roughly constant across the full chain.}
  \label{fig:growth-hierarchy}
\end{figure}

%% file: figures/fig_growth_s6.tex
\begin{figure}[t]
  \centering
  \includegraphics[width=0.62\textwidth]{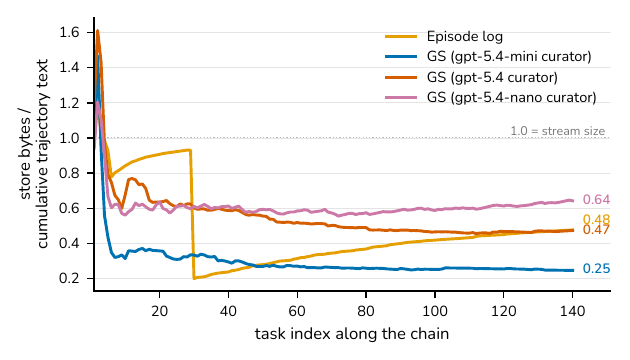}
  \caption{Store bytes over the cell's own cumulative streamed trajectory
  text (approximated by the archived trajectories' text length; denominators
  are cell-specific streams, since \execagent{} behavior sets how much text each chain generates). The \skillflat's step near task 30 is one runaway
  episode, a 300-kilobyte trajectory at 218 times the chain's median, that
  inflates its stream; the ratio then climbs back to 0.48 by chain's end.
  \skillsynth{} at mini compresses to about a quarter; the gpt-5.4-nano \mgmtagent{} chain ends highest (0.64), writing more store bytes per experienced byte than
  even the \skillflat{} retains; the gpt-5.4 \mgmtagent's 0.47 is measured against a smaller stream, since its frequent successes end episodes early.}
  \label{fig:growth-compression}
\end{figure}

%% file: figures/fig_growth_s4.tex
\begin{figure}[t]
  \centering
  \includegraphics[width=\textwidth]{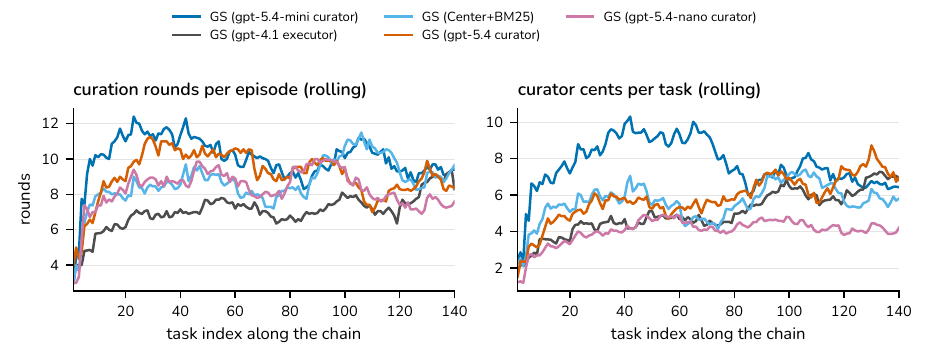}
  \caption{Curation effort per episode along the chain (20-task rolling),
  from the per-episode curation records; cents are priced at one
  shared rate card, so the panel compares effort rather than price. Effort
  does not amortize at any capability: even the gpt-5.4 \mgmtagent{}, whose store stops growing a quarter of the way in, keeps spending eight to ten
  rounds per episode on maintenance, and the nano \mgmtagent{} works the chain at a similar per-episode effort while producing the sprawling store of
  \Cref{fig:store-growth}. This matches the edit-dominated phase of
  \Cref{fig:lifecycle} and the flat per-chunk build effort of
  \Cref{fig:memory-growth}.}
  \label{fig:curation-effort}
\end{figure}

%% file: figures/fig_growth_s1pf.tex
\begin{figure}[t]
  \centering
  \includegraphics[width=\textwidth]{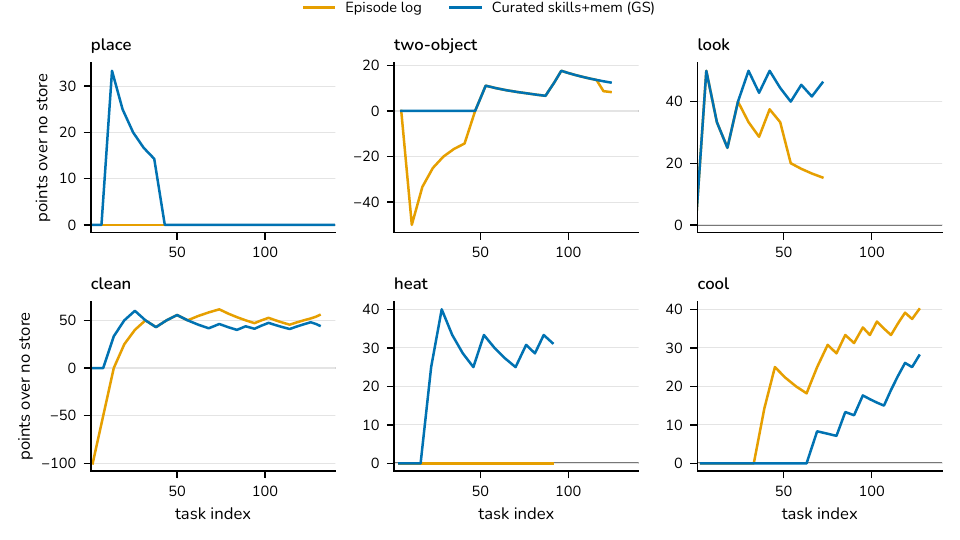}
  \caption{Per-family differenced running success at the family's interleaved
  positions (mini \execagent{}, \skillflat{} against \skillsynth{}, both
  minus the no-store chain); each curve ends at its family's last task in
  the fixed order (look's 13 tasks all fall by position 72, heat's 16 by
  91). Transfer onset is family-specific: the \skillflat's
  clean advantage builds steadily, \skillsynth{} opens its two-object and look edges early, and heat stays hard for both.}
  \label{fig:growth-families}
\end{figure}